\documentclass[10pt,twocolumn,letterpaper]{article}

\usepackage[pagenumbers]{cvpr} 
\newcounter{todocounter}
\usepackage[dvipsnames,table,xcdraw]{xcolor} 

\usepackage{multirow}

\usepackage{array}
\usepackage{booktabs}
\usepackage[dvipsnames]{xcolor}
\usepackage{tikz}

\newcommand{\+}[1]{\ensuremath{\boldsymbol{#1}}}

\graphicspath{ {./assets/images/} }

\usepackage{etoolbox,lipsum}
\usepackage{rotating}
\usepackage{soul}  
\usepackage{graphicx} 
\usepackage{caption} 

\definecolor{cvprblue}{rgb}{0.21,0.49,0.74}
\usepackage[pagebackref,breaklinks,colorlinks,citecolor=cvprblue]{hyperref}

\def\paperID{49} 
\def\confName{CVPR}
\def\confYear{2024}

\title{InteractDiffusion: Interaction Control in Text-to-Image Diffusion Models}

\author{Jiun Tian Hoe$^{1}$ \quad Xudong Jiang$^{1}$ \quad Chee Seng Chan$^{2}$ \quad Yap-Peng Tan$^{1}$ \quad Weipeng Hu$^1$ \vspace{0.3em} \\
{\normalsize $^1$Nanyang Technological University, Singapore} \quad
{\normalsize $^2$Universiti Malaya, Malaysia} \quad \\
{\tt\small jiuntian001@e.ntu.edu.sg} \quad {\tt\small \{exdjiang,eyptan,weipeng.hu\}@ntu.edu.sg} \quad {\tt\small cs.chan@um.edu.my}
}

\begin{document}

\twocolumn[{
\renewcommand\twocolumn[1][]{#1}
\maketitle
\begin{center}
    \captionsetup{type=figure}
    \includegraphics[width=0.98\linewidth]{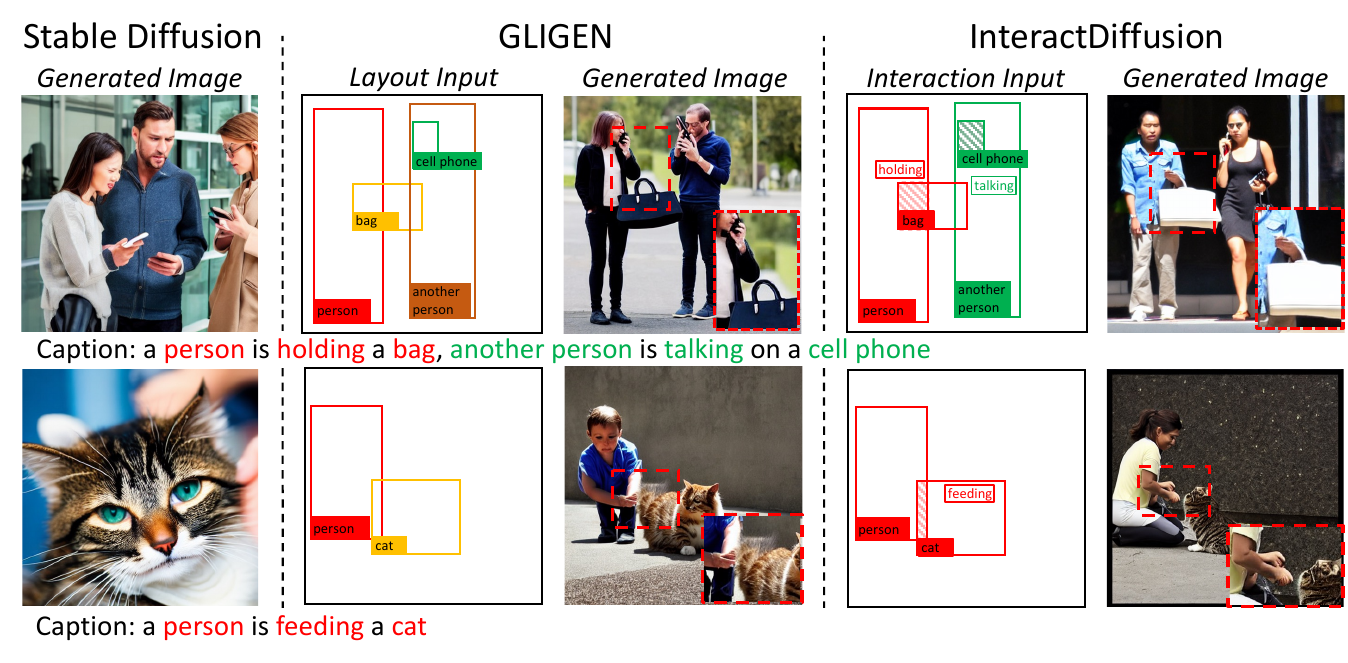}
    \captionof{figure}{Generated samples of size 512x512. Stable Diffusion conditions on text caption only, while GLIGEN conditions on extra layout input. Our proposed InteractDiffusion conditions on extra interaction label and its location shown by the shaded area.}
    \label{fig:teaser}
\end{center}
}]  
\begin{abstract}
Large-scale text-to-image (T2I) diffusion models have showcased incredible capabilities in generating coherent images based on textual descriptions, enabling vast applications in content generation. While recent advancements have introduced control over factors such as object localization, posture, and image contours, a crucial gap remains in our ability to control the interactions between objects in the generated content. Well-controlling interactions in generated images could yield meaningful applications, such as creating realistic scenes with interacting characters. In this work, we study the problems of conditioning T2I diffusion models with Human-Object Interaction (HOI) information, consisting of a triplet label (person, action, object) and corresponding bounding boxes. We propose a pluggable interaction control model, called InteractDiffusion that extends existing pre-trained T2I diffusion models to enable them being better conditioned on interactions. Specifically, we tokenize the HOI information and learn their relationships via interaction embeddings. A conditioning self-attention layer is trained to map HOI tokens to visual tokens, thereby conditioning the visual tokens better in existing T2I diffusion models. Our model attains the ability to control the interaction and location on existing T2I diffusion models, which outperforms existing baselines by a large margin in HOI detection score, as well as fidelity in FID and KID. Project page: \url{https://jiuntian.github.io/interactdiffusion}.
\end{abstract}    
\section{Introduction}\label{sec:intro}

The advent of diffusion generative models recently opens up new creative task opportunities. While diffusion models could generate diverse high quality images that reconstruct the original data distributions, it is important to control the content generated. Numerous literatures have since extensively studied how to control the image generation of the diffusion models via \eg class \cite{diffusionbeatsgan2021,ed-dpm2022}, text \cite{glide2021,stablediffusion2021,dalle2-2022,imagen2022}, image (including edge, line, scribble and skeleton) \cite{controlnet2023,universalguidance2023,composer2023} and layout \cite{reco2023,gligen2023,layoutdiffusion2023,universalguidance2023,layoutguidance2023}. However, these are insufficient to effectively express the nuanced intentions and desired outcomes, especially the interactions between objects. Our work introduces another important control in image generation: {\it interaction}.

Interaction refers to a reciprocal action between two entities or individuals. Without a doubt, interaction is an integral part of describing our daily activities. However, we find that existing diffusion models work well on static images such as paintings or scenic photos but face great challenges in generating images involving interactions. For instance, GLIGEN \cite{gligen2023} adds layout as a condition to help specify the location of objects, but controlling the relationship or interaction between the objects remains an open difficult problem, as shown in \cref{fig:teaser}. Control at the interaction level in text-to-image (T2I) diffusion models has countless applications, \eg~e-commerce, gaming, interactive storytelling etc.

This paper studies the problem of interaction-conditioned image generation, \ie how to specify the interaction in the image generation process. It faces three main challenges:
\begin{enumerate}[label=\alph*)]
\item {\bf Interaction representation}: How to represent interaction information in a meaningful token representation.
\item {\bf Intricate interaction relationship}: The relationship among objects with interaction is complex, and generating coherent images remains a great challenge.
\item {\bf Integrating conditions into existing models}: Current T2I diffusion models excel in image generation quality but lack interaction control. A pluggable module that can be seamlessly integrated into them is imperative.
\end{enumerate}

\begin{figure*}[t]
\centering
\includegraphics[width=0.95\linewidth]{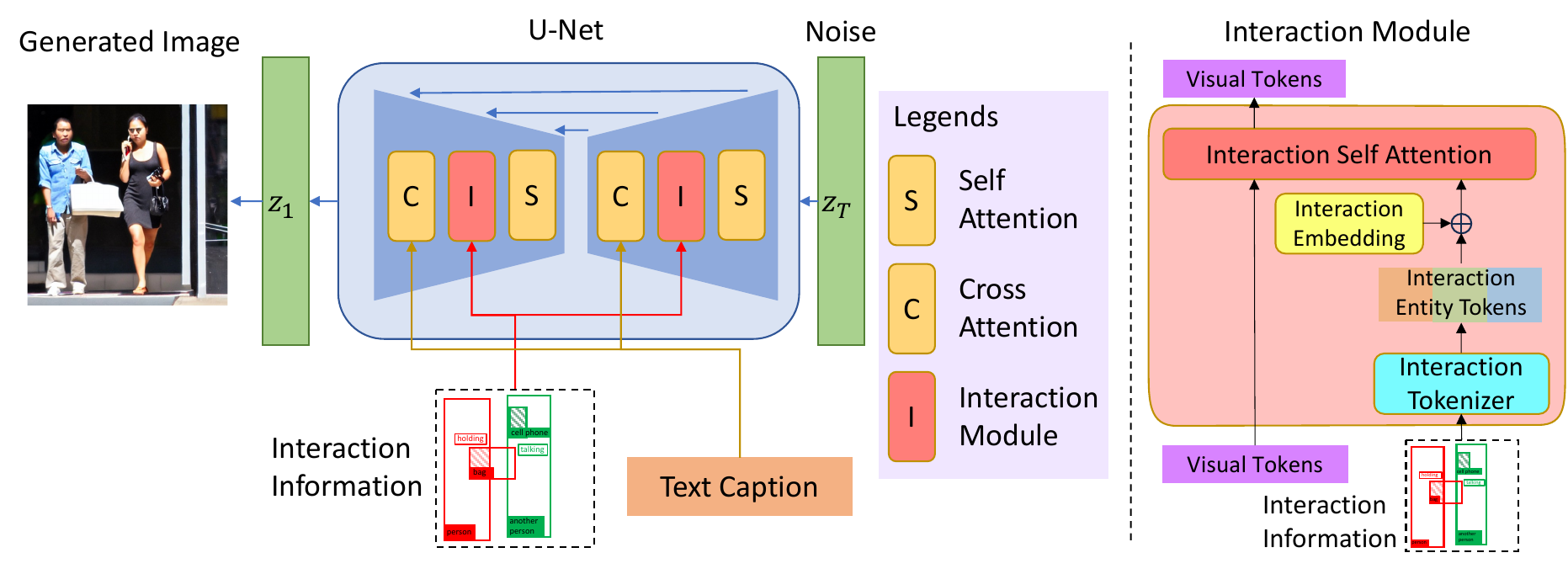}
\vspace{-5pt}
\caption{The overall framework of InteractDiffusion.
Our proposed pluggable Interaction Module $I$ seamlessly incorporates interaction information into an existing T2I diffusion model (left). The proposed module $I$ (right) consists of Interaction Tokenizer (\cref{subsec:interaction_tokenizer}) that transforms interaction information into meaningful tokens, Interaction Embedding (\cref{subsec:interaction_embed}) that incorporates intricate interaction relationship, and Interaction Self-Attention (\cref{subsec:interaction_transformer}) that integrates interaction control information into Visual Tokens of the existing T2I diffusion model.}
\label{fig:arch}
 \vspace{-10pt}
\end{figure*}

To address the aforementioned issues, we propose an interaction control model called \textbf{InteractDiffusion} as a pluggable module to existing T2I diffusion model as illustrated in \cref{fig:arch}, aiming to impose interaction control. First, to provide conditioning information to the diffusion model, we treat each interacting pair as a HOI triplet and transform its information into a meaningful token representation that contains information about position, size, and category label. Particularly, we generate three different tokens for each HOI triplet, \ie~{\it subject}, {\it action}, and {\it object} tokens. While both {\it subject} and {\it object} tokens contain information about location, size, and object category, the {\it action} token includes the location of the interaction and its category label.

Secondly, the challenge of representing intricate interaction lies in encoding the relationship between the tokens of multiple interactions where tokens are from different interaction instances and have different role within an interaction instance.
To address this challenge, we propose instance embedding and role embedding to group the tokens of the same interaction and embed their role semantically.

Thirdly, as the existing transformer block consists of a self-attention and a cross-attention layer \cite{stablediffusion2021}, we add a new Interaction Self-Attention layer in between them to incorporate interaction tokens into the existing T2I model. This helps to preserve the original model during training, while simultaneously incorporating additional interaction conditioning information.

Our main contributions are summarized as follows:

\begin{enumerate}[label=(\roman*)]
\item We address the interaction-mismatch problem in existing T2I models and raise a new challenge: controlling interaction in T2I diffusion models. We propose a new framework named \textbf{InteractDiffusion} that is pluggable to existing T2I model. It incorporates interaction information as additional conditions for training an interaction-controllable T2I diffusion model, enhancing the precision of interactions in generated images.
\item To effectively capture intricate interaction relationships, we introduce a novel method where we tokenize the localization and category information of \textlangle subject, action, object\textrangle~into three distinct tokens. These tokens are then grouped together and specified in their roles of interaction through an embedding framework. This innovative approach enhances the representation of complex interactions.
\item InteractDiffusion significantly outperforms the baseline methods in HOI Detection Scores and maintains generation quality with slight improvements in both FID and KID metrics. To the best of our knowledge, this work is the first attempt to introduce {\it interaction control} to diffusion models.
\end{enumerate}

\section{Related Work}\label{sec:related_work}
\noindent \textbf{Human-Object Interactions}
Recent advancements in Human-Object Interactions (HOI) have focused on detecting HOIs in images. It aims to locate interacting human and object pairs via bounding boxes and categorize these objects and their interactions in a triplet form, such as (person, feeding, cat). Recent works on HOI detection \cite{hotr2021, qahoi2021, fgahoi2023, rlip2022, thid2022} were DETR-based and have shown promising results. However, they still suffer from data scarcity, which hinders detection performance for rare interactions. Meanwhile, HOI image synthesis, an inverse task of HOI detection, is relatively underexplored. InteractGAN \cite{interactgan2020} proposed HOI image generation via human pose and reference images of humans and objects. However, this approach is complicated as it requires a pose-template pool and reference images of humans and objects. A more closely related work is the layout-proposal-based method \cite{relationship-generation2021}, which focuses on scene layout proposals according to HOI triplets to synthesize images. However, it is only able to generate "object placement" proposals based on inputs.  
Our work focuses on a new problem, namely, controlling the interaction in existing T2I diffusion models using simple bounding box and interaction relations in an end-to-end manner, without the need for human pose information and reference images. This approach efficiently addresses the need for more data for HOI detection tasks and opens a wide range of applications.

\noindent \textbf{Diffusion Models}
The diffusion probabilistic model was first proposed in \cite{diffusion2015}, and further improved in training and sampling methods by \cite{ddpm2020,ddim2020}. Training and evaluating diffusion models in pixel space could be costly and slow, and training on high-resolution images always requires calculating expensive gradients.
Latent Diffusion Model (LDM) \cite{stablediffusion2021} compresses the image into a latent representation of lower dimensionality \cite{tamingtransformer2021} and carries out the diffusion process in latent space to reduce the computation which was further extended to Stable Diffusion. Our work adds interaction control to the Stable Diffusion Model.

\noindent \textbf{Controlling Image Generation}
T2I diffusion models \cite{glide2021, stablediffusion2021, imagen2022, dalle2-2022} often utilize a pretrained language model like CLIP \cite{clip2021} to guide the image diffusion process. This allows the generated image's content to be controlled by a provided text caption. However, a text caption alone often provides insufficient control over the generated content, particularly when aiming to create specific content such as object location and layout, scene depth maps, human poses, boundary lines, and interactions. To address this issue, several models have proposed different methods for controlling the generated content, including object layout \cite{gligen2023, layoutdiffusion2023} and images \cite{controlnet2023}. Although controlling image generation via object layout and images can generally yield better results, one essential aspect of image has been largely ignored, namely, the interaction between objects. Our work extends the capabilities of the current T2I model by strengthening the control of interactions in the generated content.
\section{Method}\label{sec:method}
We first formulate the problem and then detail our InteractDiffusion model, as illustrated in \cref{fig:arch}. It comprises four parts: (a) \textit{interaction tokenizer} that transforms interaction conditions into tokens, (b) \textit{interaction embedding} that links the relationship between tokens of interacting triplets, (c) \textit{interaction transformer} that constructs attention between image patches and interaction information, and (d) \textit{interaction-conditional diffusion model} that generates images with interaction conditions.

\subsection{Preliminary}
We study the problem of incorporating interaction conditions $\mathbf{d}$ into existing T2I diffusion model alongside with text caption condition $\mathbf{c}$. Our aim is to train a diffusion model $f_\theta(\mathbf{z},\mathbf{c},\mathbf{d})$ to generate images conditioned on interaction $\mathbf{d}$ and text caption $\mathbf{c}$, where $\mathbf{z}$ is the initial noise.

Stable Diffusion, one of the best models, is a scale-up of the Latent Diffusion Model (LDM) \cite{stablediffusion2021} with a larger model and data size. Unlike other diffusion models, LDM splits into two stages to reduce computational complexity. It first learns a bi-directional projection to project image $\mathbf{x}$ from pixel space to a latent space as latent representation $\mathbf{z}$ and then trains a diffusion model $f_\theta(\mathbf{z},\mathbf{c})$ in the latent space with latent $\mathbf{z}$. Our work focuses on the second stage as we are only interested in conditioning the diffusion model with interaction.
\looseness=-1

LDM learns a reverse process of a fixed Markov Chain of length $T$. 
It can be interpreted as an equally weighted sequence of denoising autoencoders $\epsilon_\theta(\+z_t,t); t=1,\cdots,T$, which are trained to predict a denoised version of their input $\+z_t$, where $\+z_t$ is a noisy version of the input $\+z$. 

The unconditional objective can be viewed as
\begin{align}
\min_\mathbf{\theta} \mathcal{L}_{\text{LDM}} = \mathbb{E}_{\+z,\mathbf{\epsilon}\sim\mathcal{N}(\mathbf{0},\mathbf{I}),t} \left[ \| \mathbf{\epsilon} - \+\epsilon_\theta(\+z_t,t) \|^2_2 \right],
\end{align}
with $t$ uniformly sampled from $\{1,\cdots,T\}$.
The model iteratively produces less noisy samples from noise $\+z_T$ to $\+z_{T-1},\+z_{T-2},\cdots,\+z_{0}$, where the model $\+\epsilon_\theta(\+z_t,t)$ is realized by a UNet \cite{unet2015}. The final image is obtained by projecting $\+z_{0}$ in latent space back into image space in a single pass through the decoder trained in the first stage.

\noindent\textbf{Conditioning} In LDM, to condition the diffusion model with various modalities like text captions, a cross-attention mechanism was added on top of the UNet backbone. The conditional input of various modalities is denoted as $y$ and a domain specific encoder $\tau_\theta(\cdot)$ is used to project $y$ to an intermediate token representation $\tau_\theta(y)$.

In StableDiffusion, text captions represented by $y$ are used to condition the model. It uses a CLIP encoder denoted as $\tau_\theta(\cdot)$ to project the text caption $y$ into 77 text embeddings, \ie  $\tau_\theta(y)$. In particular, the conditioned objective for StableDiffusion can be viewed as
\begin{align}\label{eq:ldm_conditional}
\min_\mathbf{\theta} \mathcal{L}_{\text{LDM}} = \mathbb{E}_{\+z,\mathbf{\epsilon}\sim\mathcal{N}(\mathbf{0},\mathbf{I}),t} \left[ \| \mathbf{\epsilon} - \+\epsilon_\theta(\+z_t,t,\tau_\theta(y)) \|^2_2 \right],
\end{align}
where $\tau_\theta(\cdot)$ represents the CLIP text encoder and $y$ represents the text caption.
\begin{figure}[t]
\centering
\includegraphics[width=0.95\linewidth]{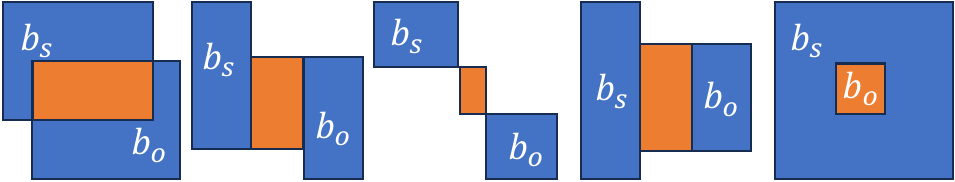}
\caption{``Between'' operation obtains the action focus area (highlighted in orange) between subject and object bounding boxes.}
\label{fig:between}
\vspace{-10pt}
\end{figure}

\subsection{Interaction Tokenizer (InToken)}\label{subsec:interaction_tokenizer} 
We define interaction $\+d$ as a triplet label consisting of \textlangle subject $s$, action $a$, and object $o$ \textrangle, as well as their corresponding bounding boxes denoted as \textlangle $\+b_s$, $\+b_a$, and $\+b_o$\textrangle, respectively. We use the subject and object bounding boxes to describe their location and sizes, and introduce an action bounding box to specify the spatial location of the action. For example, a subject (\eg women, boy) performing a specific action (\eg carrying, kicking) toward a particular object (\eg handbag, ball).

To obtain the action bounding box, we define a ``between'' operation, applied to the subject and object bounding boxes. Suppose $\+b_s$ and $\+b_o$ be specified by their corner coordinates $[\alpha_i,\beta_i], i=1,2,3,4$, the ``between'' operation on $\+b_s$ and $\+b_o$ to obtain $\+b_a$ is:
\begin{align}
    \+b_{a} &= \+b_s~\text{between}~\+b_o \nonumber\\
    &= {[R_2(\alpha_i), R_2(\beta_i)], [R_3(\alpha_i), R_3(\beta_i)]},
\end{align}
where $R_k(\cdot)$ is the $k^{\text{th}}$ rank of its arguments. Some examples of the "between" operation results are shown in \cref{fig:between}.

With this, our interaction condition inputs of an image is:
\begin{align}
    \mathcal{D}= [\+d_1,\dots,\+d_N]=[&(s_1,a_1,o_1,\+b_{s_1},\+b_{a_1},\+b_{o_1}),\dots,\nonumber\\
    &(s_N,a_N,o_N,\+b_{s_N},\+b_{a_N},\+b_{o_N})],
\end{align}
where $N$ is the number of interaction instances. 

\noindent\textbf{Subject and Object tokens} 
We first pre-process the text label and the bounding box into an intermediate representation. In particular, we use the pre-trained CLIP text encoder to encode the text of subject, action and object as a representative text embedding and use Fourier embedding \cite{nerf2022} to encode their respective bounding boxes following GLIGEN \cite{gligen2023}. To generate the subject and object tokens, $h^s, h^o$, we use a multi-layer perceptron $\text{ObjectMLP}(\cdot)$ to fuse them as:
\begin{align}
    h^s = \text{ObjectMLP}([f_{\text{text}}(s), \text{Fourier}(\+b_s)])\label{eq:objectmlp_hs}\\
    h^o = \text{ObjectMLP}([f_{\text{text}}(o), \text{Fourier}(\+b_o)]).\label{eq:objectmlp_ho}
\end{align}

\noindent\textbf{Action token} For action token, we train a separate multi-layer perceptron $\text{ActionMLP}(\cdot)$ since action is semantically apart from the subject and object,
\begin{align}
    h^a = \text{ActionMLP}([f_{\text{text}}(a), \text{Fourier}(\+b_a)]).\label{eq:actionmlp_ha}
\end{align}

\begin{figure}[t]
\centering
\includegraphics[width=0.95\linewidth]{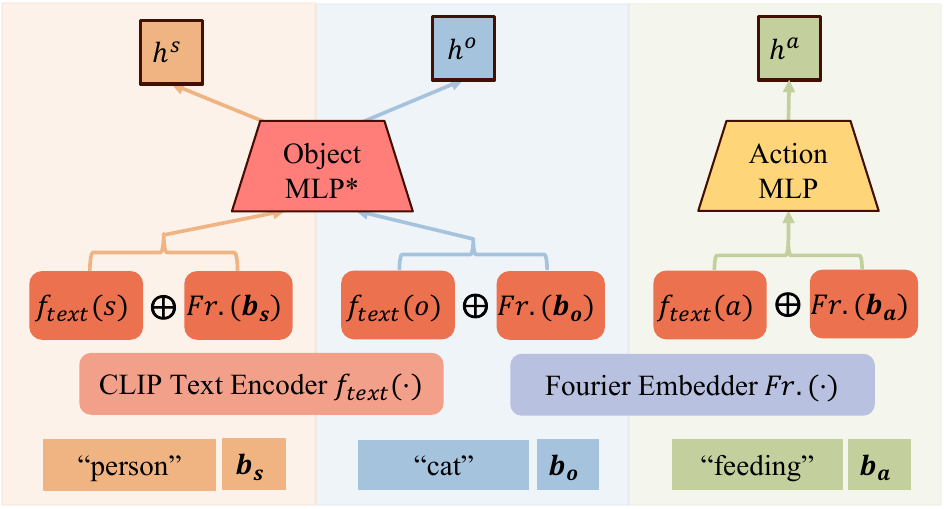}
\caption{Interaction Tokenizer. View bottom-up.}
\label{fig:interaction_tokenizer}
\vspace{-10pt}
\end{figure}

For each interaction, we transform the interaction condition input $\+d$ into a triplet of tokens $\+h$:
\begin{align}
    \+h = (h^s, h^a, h^o) = \text{InToken}(s,a,o,\+b_{s},\+b_{a},\+b_{o}),
    \label{eq:intoken}
\end{align}
where $\text{InToken}(\cdot)$ is a combination of \cref{eq:objectmlp_hs,eq:objectmlp_ho,eq:actionmlp_ha} as shown in \cref{fig:interaction_tokenizer}. 

\subsection{Interaction Embedding (InBedding)}\label{subsec:interaction_embed}
Interaction is an intricate relationship between subject, object and their action. From \cref{eq:intoken}, tokens $h^s,h^a,h^o$ are individually embedded (as shown in \cref{fig:arch}). For multiple interaction instances, all tokens $h^s_i,h^a_i,h^o_i;i=1,\cdots,N$, are individually embedded. Therefore, it is necessary to group these tokens by interaction instance and specify different role of tokens within the interaction instance.
Segment Embedding, as introduced in \cite{bert2018}, has demonstrated its effectiveness in capturing relationships between segments in a text sequence by adding a learnable embedding to tokens to group a sequence of words into segments. 
In our work, we extend this concept to group the tokens into triplets. Specifically, we add a new \textit{instance embedding} denoted as $q\in \{q_1,\dots,q_N\}$ to interaction instances $\+h \in \{\+h_1, \cdots, \+h_N\}$ as:
\begin{align}
    \+e_i &= \+h_i + q_i,
\end{align}
where all tokens in the same instance share the same instance embedding. This groups all tokens into interaction instances or triplets.

\begin{figure}[t]
    \centering
    \includegraphics[width=0.95\linewidth]{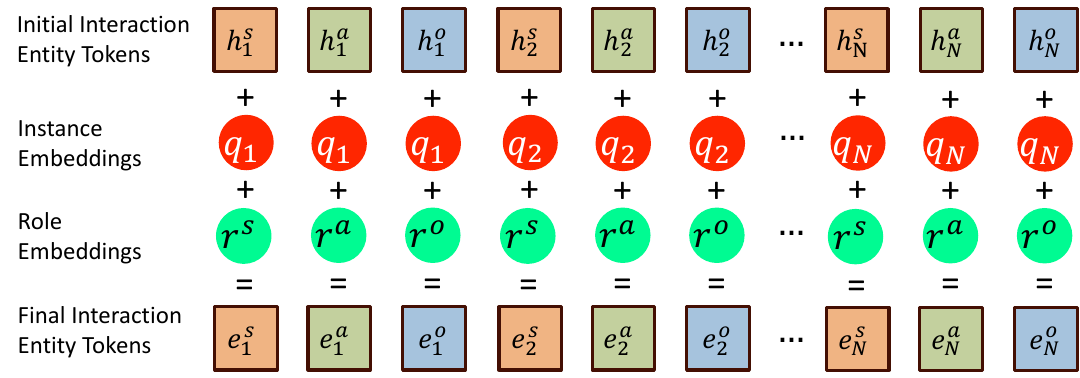}
    \caption{Interaction Embeddings. Learnable instance embedding $q$ and role embedding $r$ are added to tokens to represent intricate interaction relationships between subject $s$, action $a$ and object $o$.}
    \label{fig:embeddings}
    \vspace{-10pt}
\end{figure}

Besides, each token in the triplet has different role. So, we embed their roles with three \textit{role embeddings} $r\in \{r^s,r^a,r^o\}$ to form final entity token $\+e_i$: 
\begin{align}
    \+e_i &= \+h_i + q_i + r \nonumber\\
    &= (h^s_i + q_i + r^s,~ h^a_i + q_i + r^a,~ h^o_i + q_i + r^o)\label{eq:interaction_em2},
\end{align} 
where $r^s$, $r^a$ and $r^o$ represent the role embeddings for subject, action and object respectively. From \cref{eq:interaction_em2} we see that tokens of the same role in all instances share the same role embedding.
Adding instance and role embedding to the interaction entity token $\+h_i$ (as in \cref{fig:embeddings}) encodes the intricate interaction relationship, \ie specifies a token's role and interaction instance, which results in significantly improved image generation, especially in scenarios with multiple interaction instances.

\subsection{Interaction Transformer (InFormer)}\label{subsec:interaction_transformer}

Large-scale T2I models such as Stable Diffusion have been trained on massive-scale image-text pairs and demonstrated remarkable capabilities in generating highly realistic images, owing to the knowledge acquired during large-scale pre-training. In this paper, we aim to incorporate the interaction control into these T2I models with minimal cost. Therefore, it is crucial to preserve the valuable knowledge embedded in them.

Lets denote $\+v=[v_1,\cdots,v_M]$ as the visual feature tokens of an image, and $\+c$ as the caption tokens where $\+c=\tau_\theta(y)$. In LDM models, a Transformer block consists of two attention layers, \ie (i) self-attention layer for the visual tokens and (ii) cross-attention layers that model the attention between visual tokens and caption tokens: 
\begin{align}
    \+v &= \+v + \text{SelfAttn}(\+v) \label{eq:selfattn};~~~
    \+v = \+v + \text{CrossAttn}(\+v, \+c) 
\end{align}

\begin{figure}
    \centering
    \includegraphics[width=0.95\linewidth]{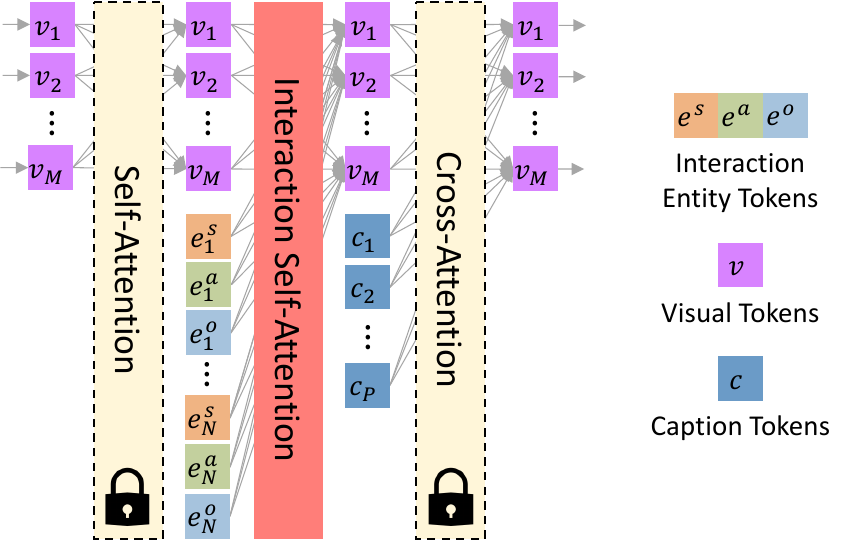}
    \caption{Interaction Transformer. An Interaction Self-Attention is added between the visual token self-attention and the visual-caption cross-attention to incorporate the interaction conditions.}
    \label{fig:transformer_block}
    \vspace{-10pt}
\end{figure}

\noindent\textbf{Interaction Self-Attention} Following GLIGEN \cite{gligen2023}, we freeze the two original attention layers and introduce a new gated self-attention layer namely \textit{Interaction Self-Attention} (see \cref{fig:transformer_block}) between them. This is to add the interaction condition onto the existing Transformer block. Different from \cite{gligen2023}, we perform self-attention over the concatenation of visual and interaction tokens $[\+v,\+e^s,\+e^a,\+e^o]$, which focuses on the relationship of interactions as:
\begin{align}
    \+v 
    &= \+v + \eta \cdot \tanh{\gamma} \cdot \text{TS}(\text{SelfAttn}([\+v,\+e^s,\+e^a,\+e^o]))\label{eq:gated_attention},
\end{align}
where $\text{TS}(\cdot)$ is a Token Slicing operation to keep only the output of visual tokens and slice off the others as shown in \cref{fig:transformer_block}, $\eta$ is a hyper-parameter for scheduled sampling that controls the activation of Interaction Self-Attention and $\gamma$ is a zero-initialized learnable scale that gradually controls the flow of the gate. 
Note that \cref{eq:gated_attention} performs in between the two parts of \cref{eq:selfattn}. As a summary, our Interaction Self-Attention layer transforms the interaction information, including the interaction, subject and object bounding boxes, into visual tokens.

\noindent \textbf{Scheduled Sampling} We set $\eta$ = $1$ in \cref{eq:gated_attention} during training and standard inference scheme as to \cite{gligen2023}. However, in some occasional situations, the newly added Interaction Self-Attention layer could cause sub-optimal effects on existing T2I models. Thus we include a control on sampling interval on the Interaction Self-Attention layer, which can balance out the level of text caption and interaction control. 
 
Technically, our scheduled sampling scheme is controlled during the inference time by a hyper-parameter $\omega\in [0,1]$. It defines the proportion of diffusion steps influenced by the interaction control as follow:
\begin{align}
    \eta = 
    \begin{cases}
        1, & t \leq \omega * T \quad \text{\# Text + Interaction} \\
        0, & t > \omega * T \quad \text{\# Text only}
    \end{cases}
\end{align}
where $T$ is total number of diffusion steps.

\subsection{Interaction-conditional Diffusion Model}
We combine InToken, InBedding and InFormer to form the pluggable Interaction Module, enabling interaction control in existing T2I diffusion models.
The LDM training objective (\cref{eq:ldm_conditional}) is adopted. Denoting the newly added parameters as $\theta'$, the diffusion model is now defined as $\+\epsilon_{\theta,\theta'}(\cdot)$ where the extra interaction information is processed by the interaction tokenizer $\tau_{\theta'}(\cdot)$. As such, the overall training objective of our model is:
\begin{align}
\min_\mathbf{\theta'}~&\mathcal{L}_{\text{InteractDiffusion}} = \\
&\mathbb{E}_{\+z,\mathbf{\epsilon}\sim\mathcal{N}(\mathbf{0},\mathbf{I}),t} \left[ \| \mathbf{\epsilon} - \+\epsilon_{\theta,\theta'}(\+z_t,t,\tau_{\theta}(y),\tau_{\theta'}(\mathcal{D})) \|^2_2 \right].\nonumber
\end{align}
\section{Experiments}\label{sec:experiments}
\begin{figure*}[ht]
\begin{minipage}[t]{0.55\textwidth}
\centering
\setlength{\tabcolsep}{3pt} 
\resizebox{!}{1.25cm}{ 
\begin{tabular}{l|ll|cccc|cccc}
\hline
\multirow{3}{*}{Model} & \multicolumn{2}{c|}{\multirow{2}{*}{Quality~$\downarrow$}}                                & \multicolumn{4}{c|}{FGAHOI Swin-Tiny (mAP)~$\uparrow$}                                                                           & \multicolumn{4}{c}{FGAHOI Swin-Large (mAP)~$\uparrow$}                                                                          \\ \cline{4-11} 
                       & \multicolumn{2}{l|}{}                                                        & \multicolumn{2}{c|}{Default}              & \multicolumn{2}{l|}{Known Object} & \multicolumn{2}{c|}{Default}              & \multicolumn{2}{l}{Known Object} \\ \cline{2-11} 
                       & \multicolumn{1}{c}{FID} & \multicolumn{1}{c|}{KID} & Full           & \multicolumn{1}{c|}{Rare}           & Full                  & Rare                 & Full           & \multicolumn{1}{c|}{Rare}           & Full                 & Rare                 \\ \hline
StableDiffusion   & 35.85                                & 0.01297                               & 0.63           & \multicolumn{1}{c|}{0.68}           & 0.66                  & 0.70                 & 0.64           & \multicolumn{1}{c|}{0.83}           & 0.65                 & 0.84                 \\
GLIGEN                 & 29.35                                & 0.01275                               & 21.73          & \multicolumn{1}{c|}{15.35}          & 23.31                 & 17.24                & 23.99          & \multicolumn{1}{c|}{19.56}          & 24.99                & 20.37                \\
GLIGEN*                & 18.82                                & 0.00694                               & 25.23          & \multicolumn{1}{c|}{17.45}          & 26.66                 & 18.78                & 26.45          & \multicolumn{1}{c|}{18.93}          & 27.32                & 19.90                \\ \hline
InteractDiffusion      & \textbf{18.69}                       & \textbf{0.00676}                      & \textbf{29.53} & \multicolumn{1}{c|}{\textbf{23.02}} & \textbf{30.99}        & \textbf{24.93}       & \textbf{31.56} & \multicolumn{1}{c|}{\textbf{26.09}} & \textbf{32.52}       & \textbf{27.04}       \\ \hline \hline
HICO-DET              & \multicolumn{1}{c}{-}                & \multicolumn{1}{c|}{-}                & 29.94          & \multicolumn{1}{c|}{22.24}          & 32.48                 & 24.16                & 37.18          & \multicolumn{1}{c|}{30.71}          & 38.93                & 31.93                \\ \hline
\end{tabular}}
\captionof{table}{Comparison between InteractDiffusion and existing baselines in terms of generated image quality scores in FID and KID and HOI detection score in mAP.  GLIGEN* is HICO-DET fine-tuned GLIGEN model.
The last row shows the Detection Score from real images.}
\label{tab:fid_map}
\end{minipage}
\hfill
\begin{minipage}[t]{0.42\textwidth}
\centering
\setlength{\tabcolsep}{3pt} 
\resizebox{!}{1.2cm}{ 
\begin{tabular}{l|lll|cc|cc|cc}
\hline
\multirow{2}{*}{Model}             & \multirow{2}{*}{Tr.} & \multirow{2}{*}{To.} & \multirow{2}{*}{Em.} & \multicolumn{2}{c|}{Quality}        & \multicolumn{2}{c|}{Default~$\uparrow$} & \multicolumn{2}{c}{Kn. Obj.~$\uparrow$} \\ \cline{5-10} 
                                   &                      &                      &                      & FID~$\downarrow$ & KID~$\downarrow$ & Full               & Rare               & Full                 & Rare                 \\ \hline
StableDiffusion               &                      &                      &                      & 35.85            & 0.01297          & 0.63               & 0.68               & 0.66                 & 0.70                 \\
GLIGEN                             & \checkmark *         &                      &                      & 29.35            & 0.01275          & 21.73              & 15.35              & 23.31                & 17.24                \\
GLIGEN*                            & \checkmark *         &                      &                      & 18.82            & 0.00694          & 25.23              & 17.45              & 26.66                & 18.78                \\ \hline
\multirow{2}{*}{InteractDiffusion} & \checkmark           & \checkmark           &                      & 18.88            & 0.00686          & 28.73              & 21.93              & 30.15                & 23.38                \\
                                   & \checkmark           & \checkmark           & \checkmark           & \textbf{18.69}   & \textbf{0.00676} & \textbf{29.53}     & \textbf{23.02}     & \textbf{30.99}       & \textbf{24.93}       \\ \hline \hline
HICO-DET                           &                      &                      &                      & -                & -                & 29.94              & 22.24              & 32.48                & 24.16                \\ \hline
\end{tabular}}
\captionof{table}{Ablation study of InteractDiffusion. Tr., To., and Em. represent Interaction Transformer, Interaction Tokenizer, and Interaction Embedding respectively. \checkmark* indicate Gated Self-Attention in GLIGEN.} 
\label{tab:ablation}
\end{minipage}
\end{figure*}

We train and evaluate models at 512x512 resolution. We initialize our model with the pre-trained GLIGEN model based on StableDiffusion v1.4. Training uses a constant learning rate of 5e-5 with Adam optimization and a linear warm-up for the initial 10k iterations. It ran for 500k iterations with a batch size of 8 ($\approx$ 106 epochs), taking around 160 hours on 2 NVIDIA GeForce RTX 4090 GPUs. We use a gradient accumulate step of 2, resulting in an effective batch size of 16. For inference, we employ diffusion sampling steps of 50 with the PLMS \cite{plms2022} sampler. More details are given in \cref{sec:implementation_detail} of supplementary.

\subsection{Datasets}
Our experiments were conducted on the widely-used HICO-DET dataset \cite{hicodet2018}, which comprises 47,776 images: 38,118 for training and 9,658 for testing. The dataset includes 151,276 HOI annotations: 117,871 in training and 33,405 in testing. 
HICO-DET includes 600 types of HOI triplets constructed from 80 object categories and 117 verb classes. We extracted the annotations in the testing set as input to generate interaction images and subsequently performed HOI detection on the generated images using FGAHOI \cite{fgahoi2023}.

Following the evaluation methodology outlined in HICO-DET \cite{hicodet2018}, we evaluated the generation results in both Default and Known Object settings. In the Default setting, the average precision (AP) is computed across all testing images for each HOI class. The Known Object setting, on the other hand, calculates the AP of an HOI class solely over the images containing the object in the corresponding HOI class (e.g., the AP of the HOI class 'riding bicycle' is calculated exclusively on the images containing the 'bicycle' object). We reported the HOI detection results in the Full and Rare subsets. The Full and Rare subsets consist of 600 and 138 HOI classes, respectively, with a rare class defined as one represented by less than 10 training samples.

\subsection{Evaluation Metrics}
We evaluate the quality and controllability of interaction in generation with three metrics.

\noindent\textbf{Fréchet Inception Distance} \cite{fid2017} measures the Fréchet distance in distribution of Inception feature between the real-images and the generated images (FID). 

\noindent\textbf{Kernel Inception Distance} \cite{kid2018} measures the squared Maximum Mean Discrepancy (MMD) between the Inception features of the real and generated images using a polynomial kernel. It relaxes the Gaussian assumption in FID and requires fewer samples.

\noindent\textbf{HOI Detection Score} is proposed as a measure of the controllability of interaction in generation models. To evaluate this, we utilize the pretrained state-of-the-art HOI detector, FGAHOI \cite{fgahoi2023}, to detect the HOI instances in generated images and compare them against the ground truth from the original annotations in HICO-DET. This process quantifies the models' controllability in interaction generation. We report the HOI Detection Score based on the FGAHOI protocol in two categories, namely {\it Default} and {\it Known} Object. Default setting is more challenging as it requires distinguishing the non-related images. FGAHOI is implemented with Swin-Tiny and Swin-Large backbones, and we evaluate with the both.

In summary, FID and KID assess generation quality, while HOI Det. Score evaluates interaction controllability.

\subsection{Qualitative results}

\begin{table*}[!ht]
    \centering
    \setlength{\tabcolsep}{1pt} 
    \renewcommand{\arraystretch}{1} 
    \begin{tabular}{c ccccccccc}
        \rotatebox[origin=c]{90}{Input}
        & \raisebox{-0.5\height}{\includegraphics[width=.104\linewidth]{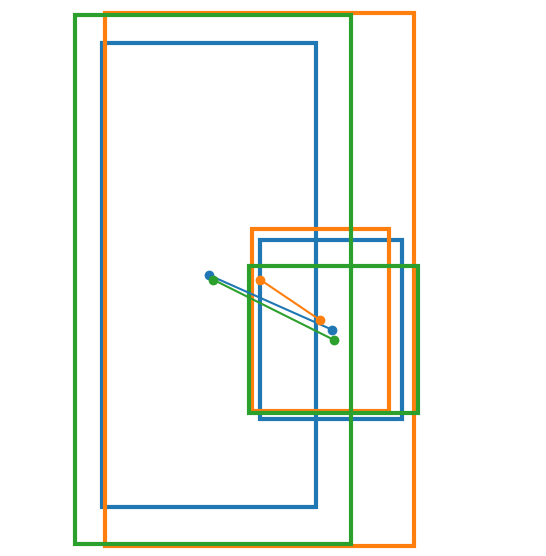}}
        & \raisebox{-0.5\height}{\includegraphics[width=.104\linewidth]{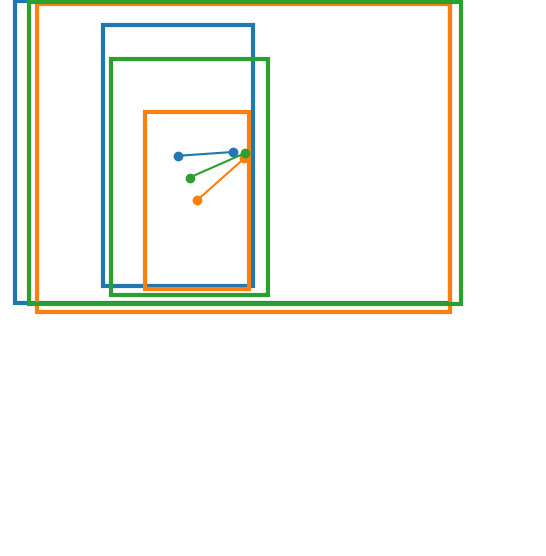}}
        & \raisebox{-0.5\height}{\includegraphics[width=.104\linewidth]{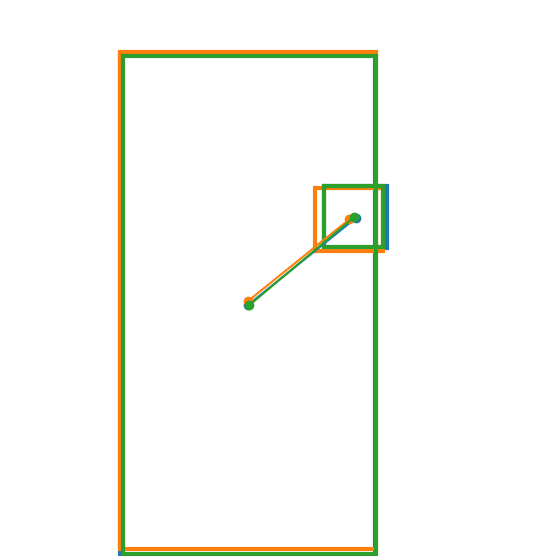}}
        & \raisebox{-0.5\height}{\includegraphics[width=.104\linewidth]{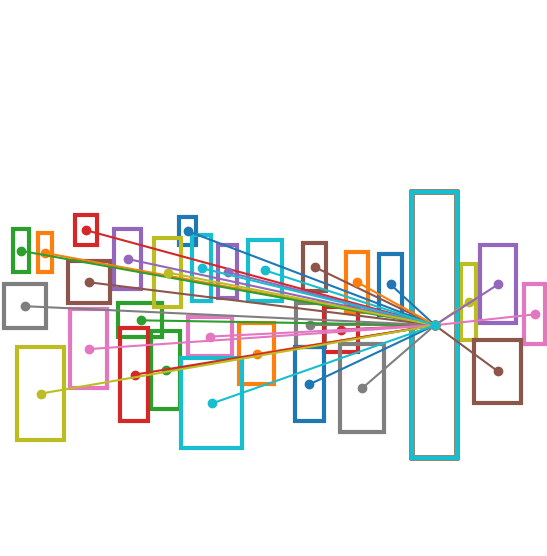}}
        & \raisebox{-0.5\height}{\includegraphics[width=.104\linewidth]{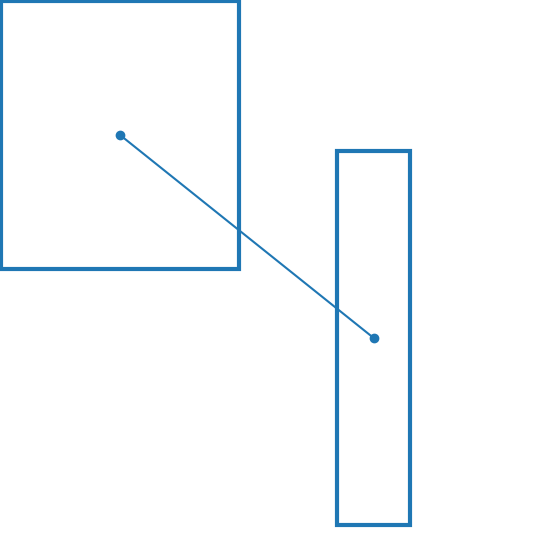}}
        & \raisebox{-0.5\height}{\includegraphics[width=.104\linewidth]{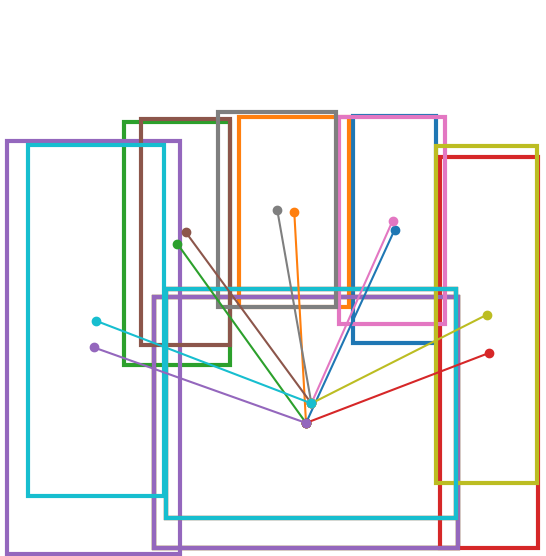}}
        & \raisebox{-0.5\height}{\includegraphics[width=.104\linewidth]{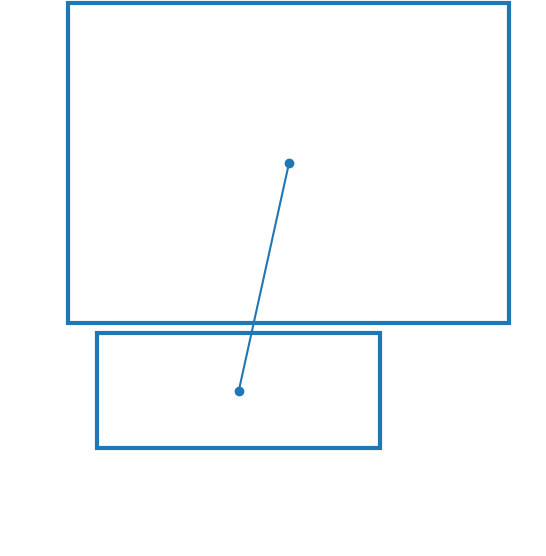}}
        & \raisebox{-0.5\height}{\includegraphics[width=.104\linewidth]{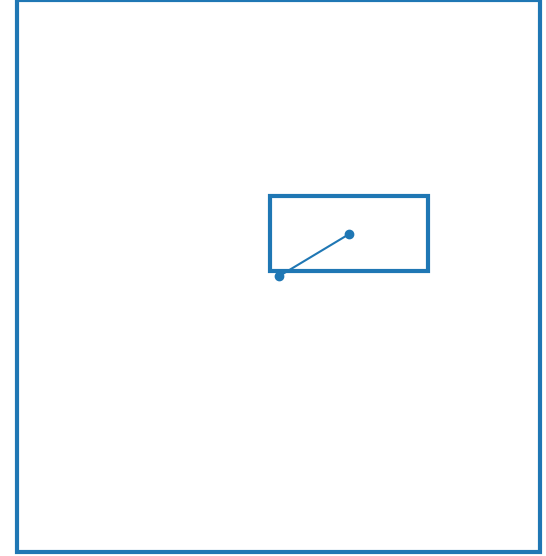}} 
        & \raisebox{-0.5\height}{\includegraphics[width=.104\linewidth]{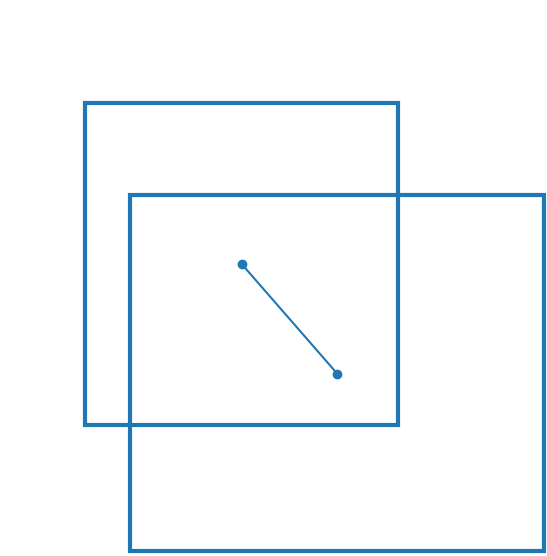}}\\
        \\[-1em]
        \rotatebox[origin=c]{90}{Caption} 
        & \raisebox{-0.5\height}{\includegraphics[width=.104\linewidth]{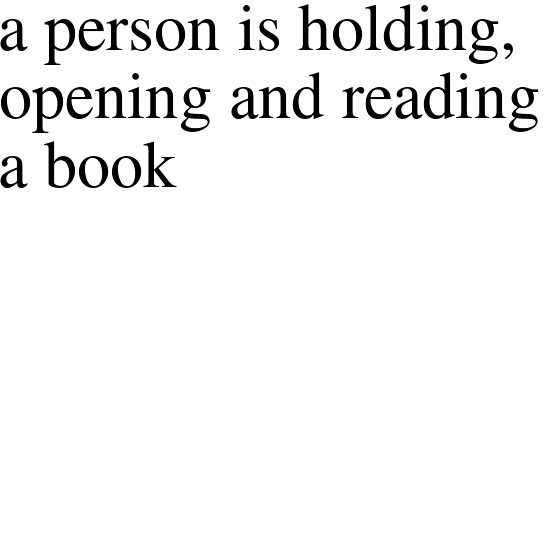}}
        & \raisebox{-0.5\height}{\includegraphics[width=.104\linewidth]{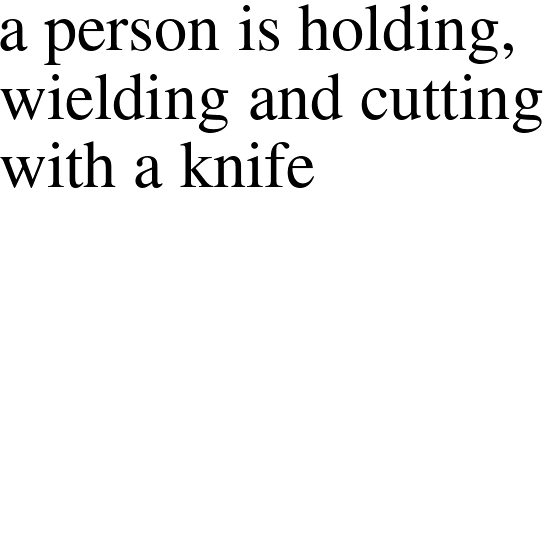}}
        & \raisebox{-0.5\height}{\includegraphics[width=.104\linewidth]{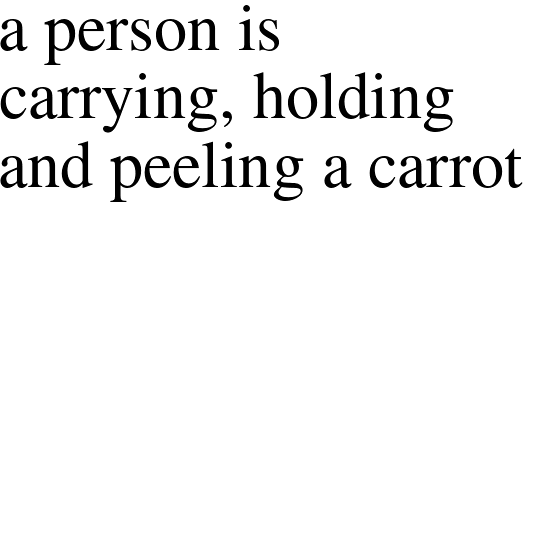}}
        & \raisebox{-0.5\height}{\includegraphics[width=.104\linewidth]{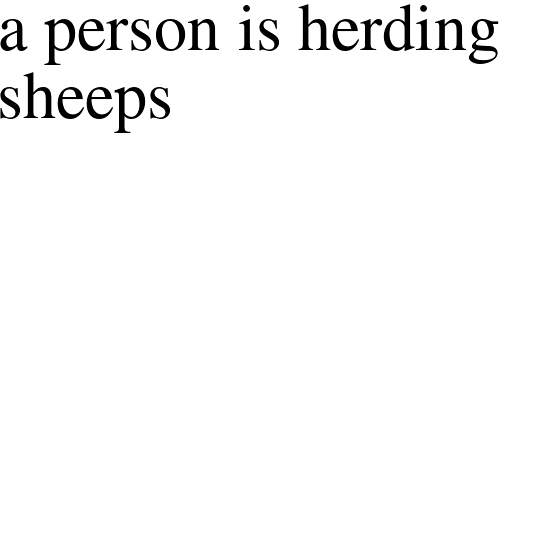}}
        & \raisebox{-0.5\height}{\includegraphics[width=.104\linewidth]{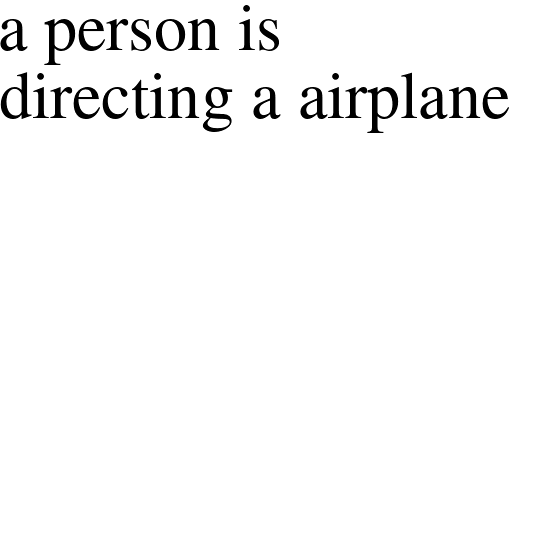}}
        & \raisebox{-0.5\height}{\includegraphics[width=.104\linewidth]{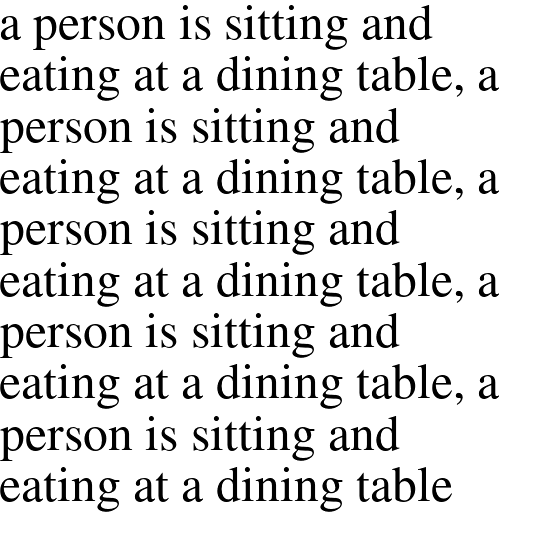}}
        & \raisebox{-0.5\height}{\includegraphics[width=.104\linewidth]{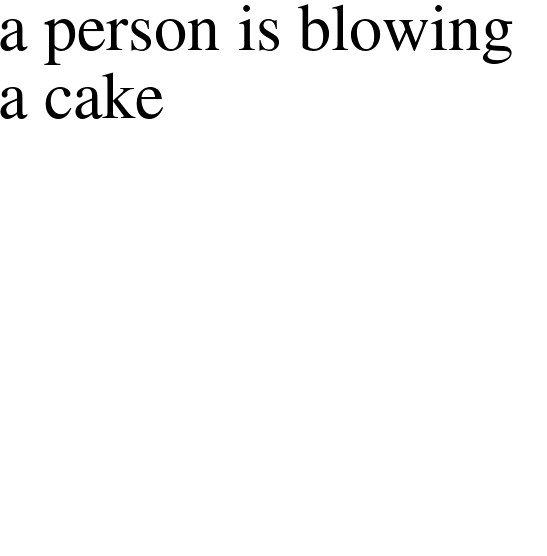}}
        & \raisebox{-0.5\height}{\includegraphics[width=.104\linewidth]{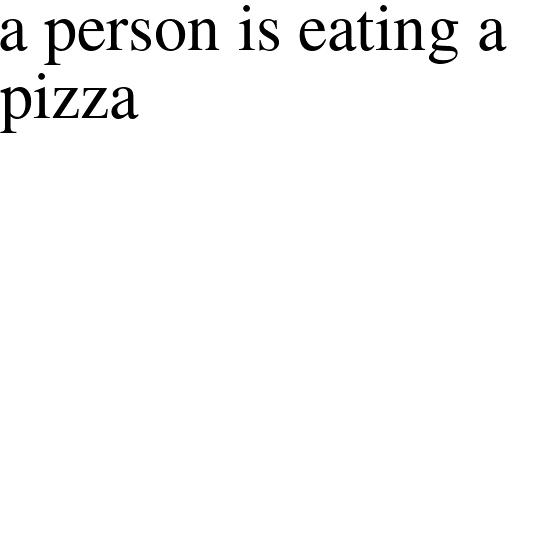}}
        & \raisebox{-0.5\height}{\includegraphics[width=.104\linewidth]{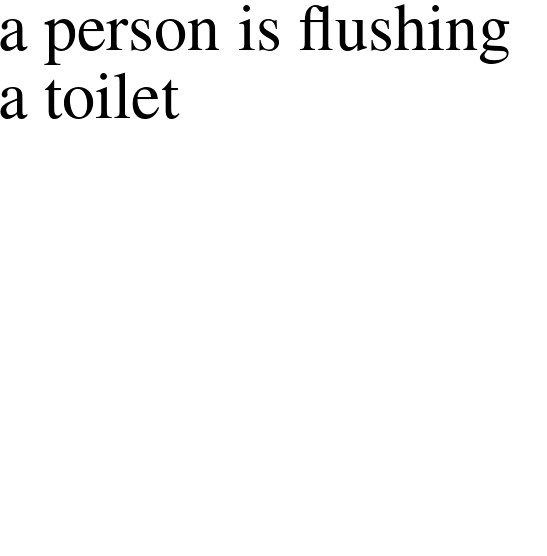}}\\
        \\[-1em]
        \rotatebox[origin=c]{90}{GT}
        & \raisebox{-0.5\height}{\includegraphics[width=.104\linewidth]{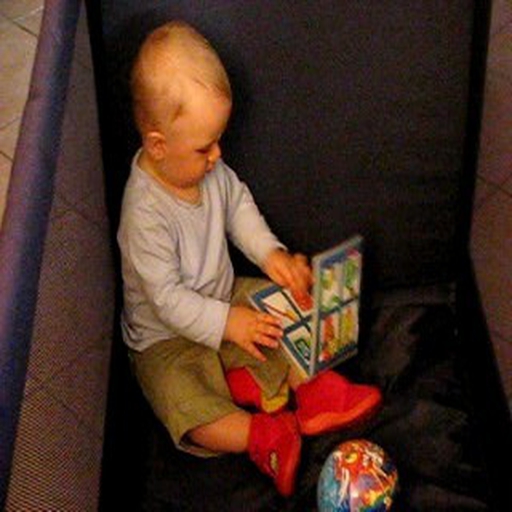}}
        & \raisebox{-0.5\height}{\includegraphics[width=.104\linewidth]{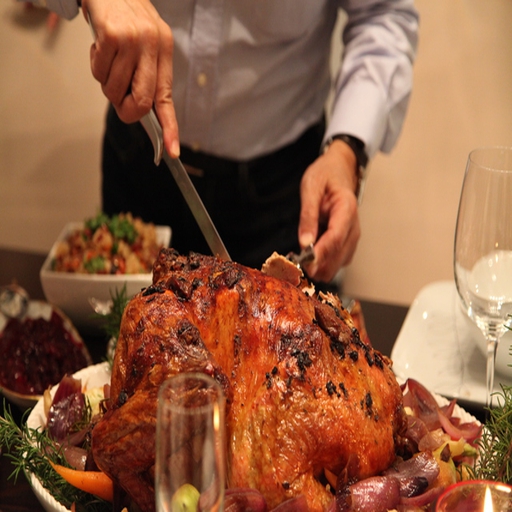}}
        & \raisebox{-0.5\height}{\includegraphics[width=.104\linewidth]{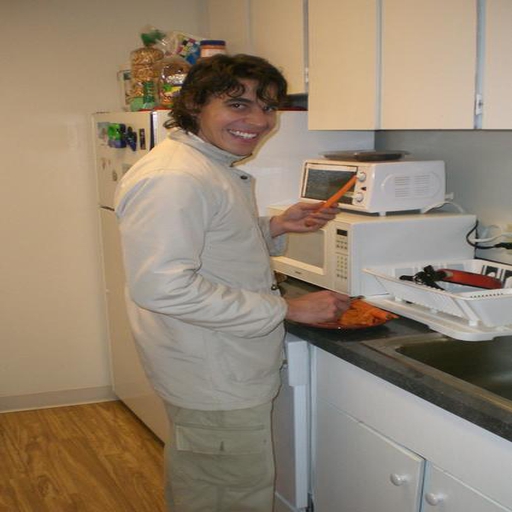}}
        & \raisebox{-0.5\height}{\includegraphics[width=.104\linewidth]{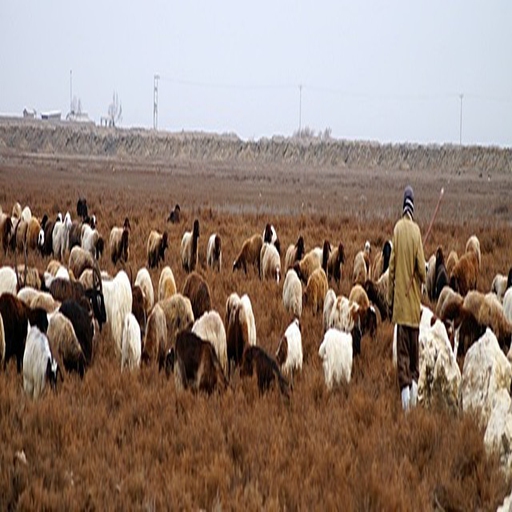}}
        & \raisebox{-0.5\height}{\includegraphics[width=.104\linewidth]{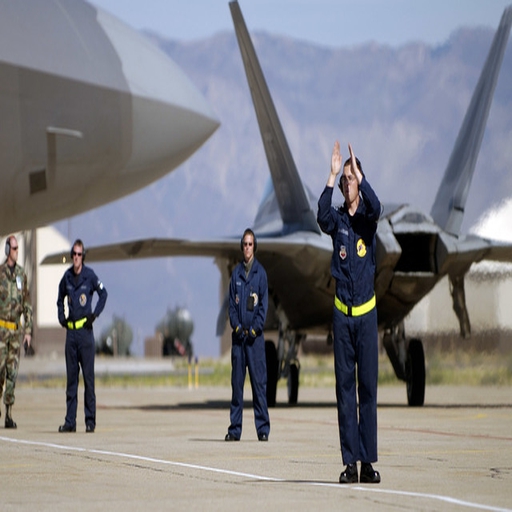}}
        & \raisebox{-0.5\height}{\includegraphics[width=.104\linewidth]{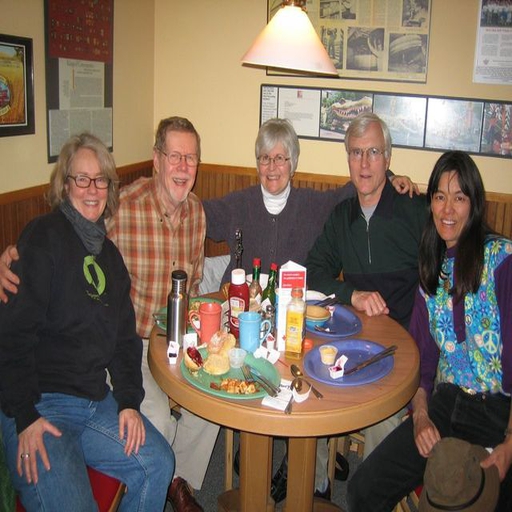}}
        & \raisebox{-0.5\height}{\includegraphics[width=.104\linewidth]{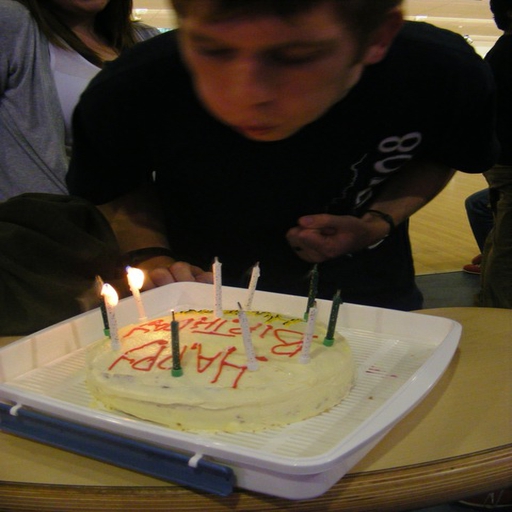}}
        & \raisebox{-0.5\height}{\includegraphics[width=.104\linewidth]{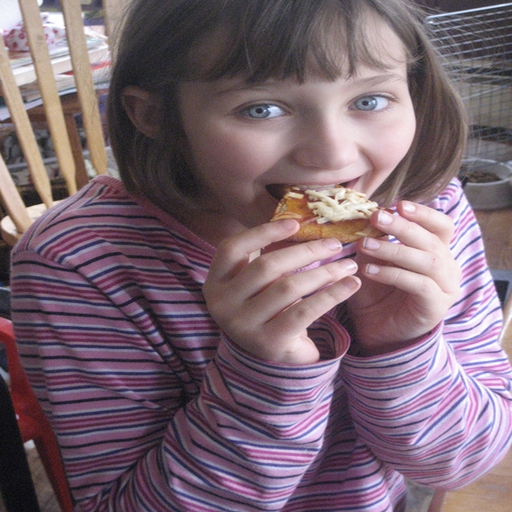}}
        & \raisebox{-0.5\height}{\includegraphics[width=.104\linewidth]{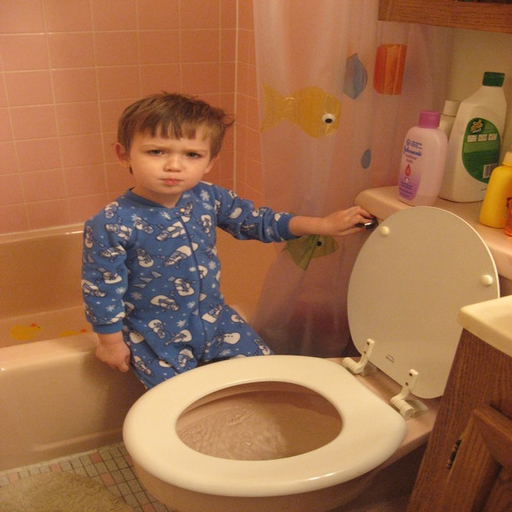}}\\
        \\[-1em]
        \rotatebox[origin=c]{90}{SD}
        & \raisebox{-0.5\height}{\includegraphics[width=.104\linewidth]{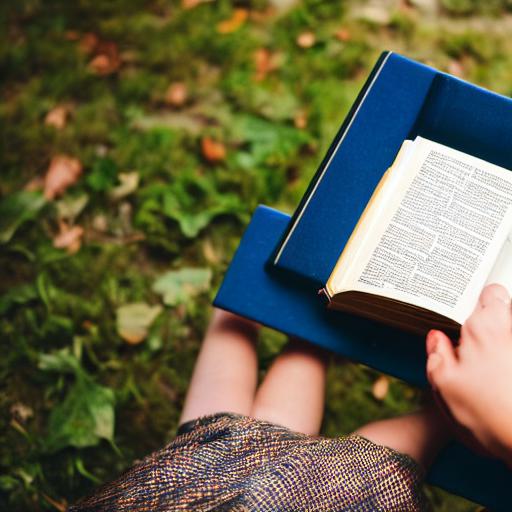}}
        & \raisebox{-0.5\height}{\includegraphics[width=.104\linewidth]{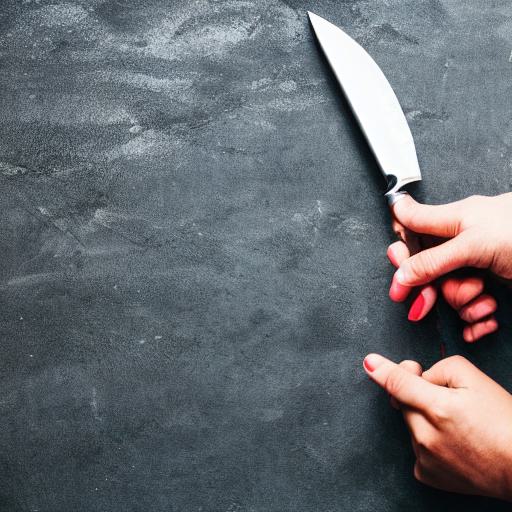}}
        & \raisebox{-0.5\height}{\includegraphics[width=.104\linewidth]{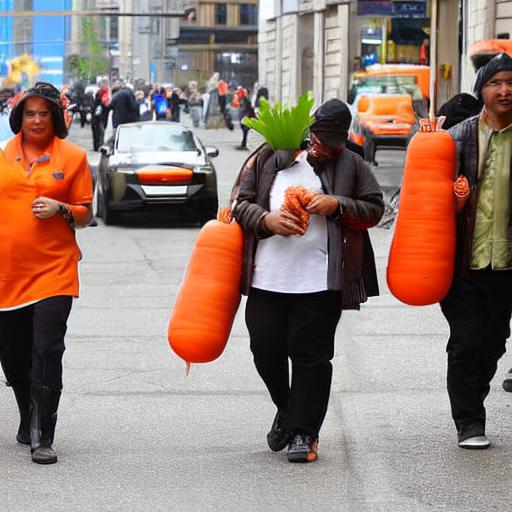}}
        & \raisebox{-0.5\height}{\includegraphics[width=.104\linewidth]{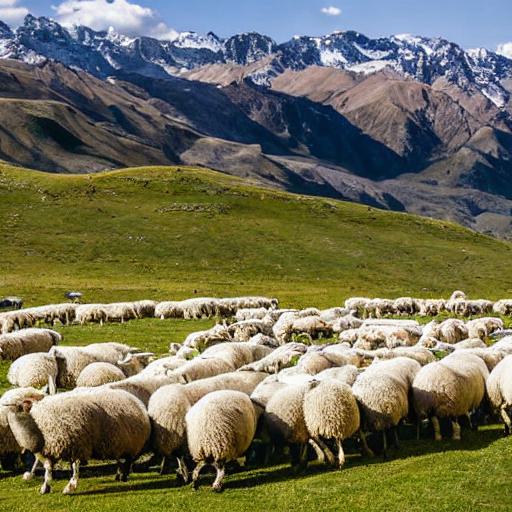}}
        & \raisebox{-0.5\height}{\includegraphics[width=.104\linewidth]{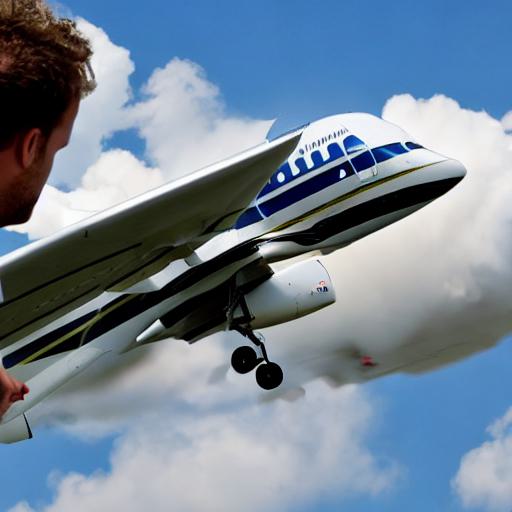}}
        & \raisebox{-0.5\height}{\includegraphics[width=.104\linewidth]{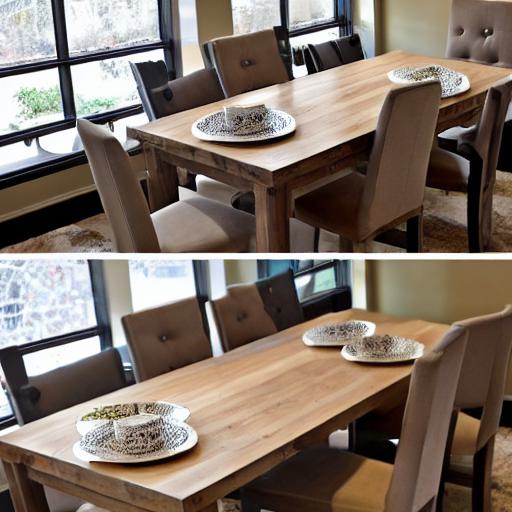}}
        & \raisebox{-0.5\height}{\includegraphics[width=.104\linewidth]{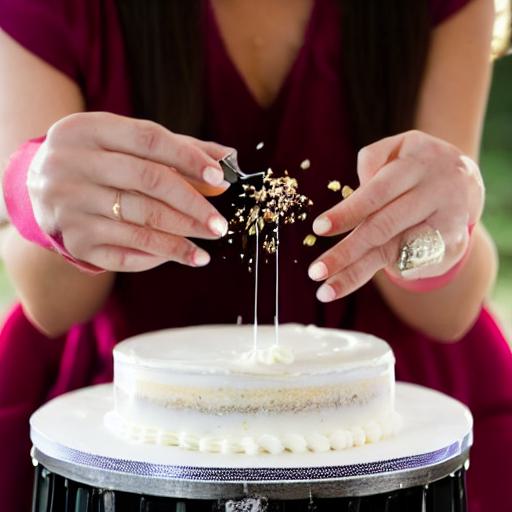}}
        & \raisebox{-0.5\height}{\includegraphics[width=.104\linewidth]{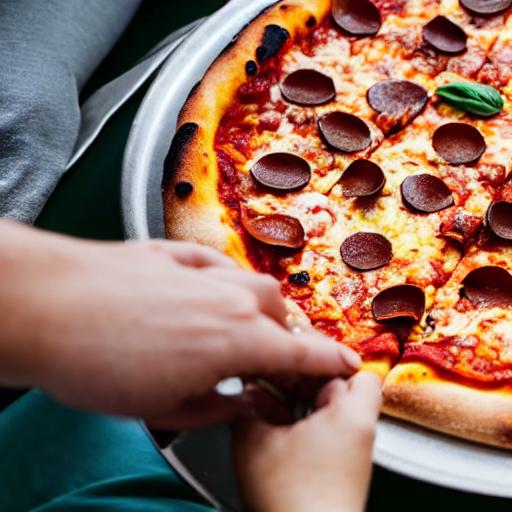}}
        & \raisebox{-0.5\height}{\includegraphics[width=.104\linewidth]{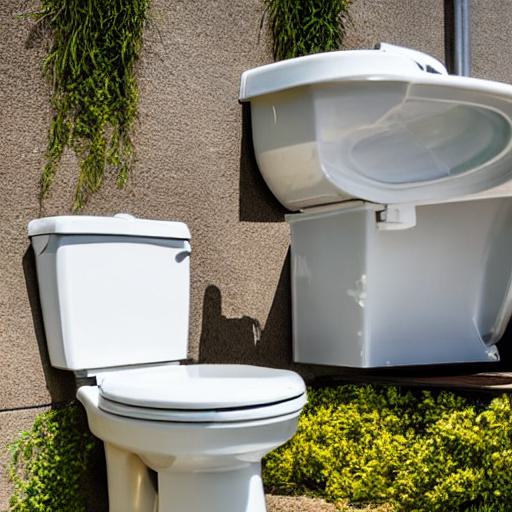}}\\
        \\[-1em]
        \rotatebox[origin=c]{90}{GLIGEN}
        & \raisebox{-0.5\height}{\includegraphics[width=.104\linewidth]{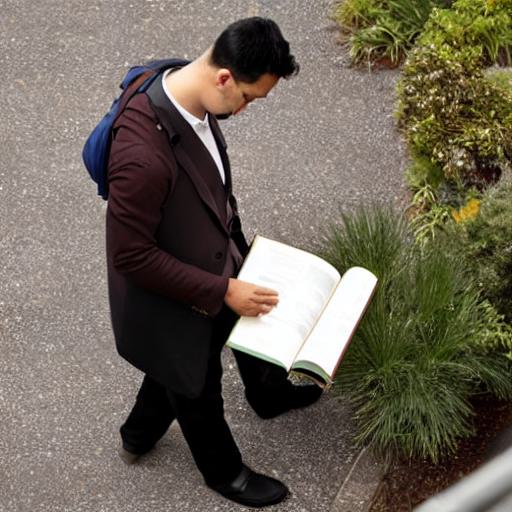}}
        & \raisebox{-0.5\height}{\includegraphics[width=.104\linewidth]{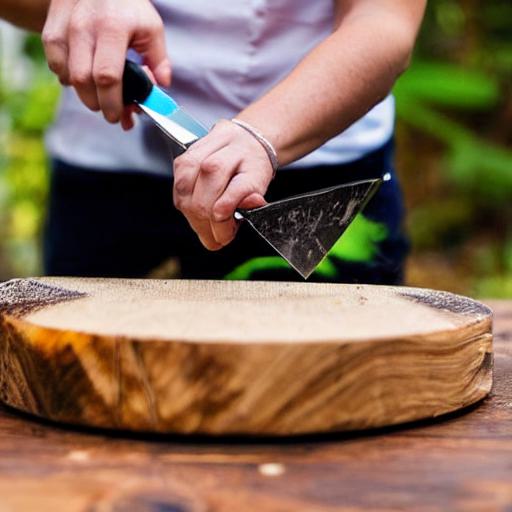}}
        & \raisebox{-0.5\height}{\includegraphics[width=.104\linewidth]{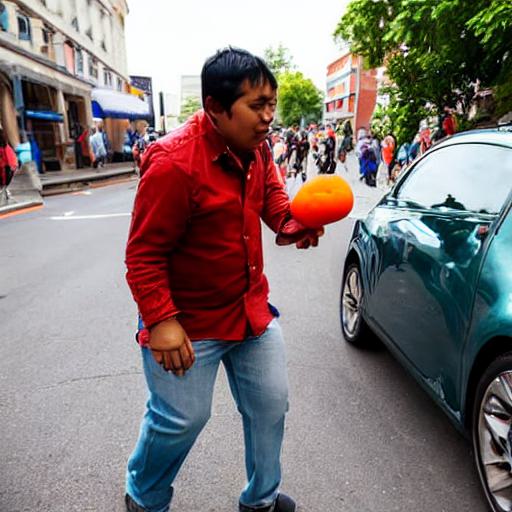}}
        & \raisebox{-0.5\height}{\includegraphics[width=.104\linewidth]{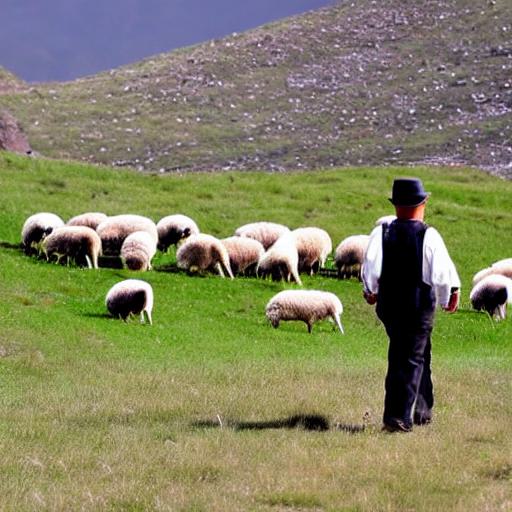}}
        & \raisebox{-0.5\height}{\includegraphics[width=.104\linewidth]{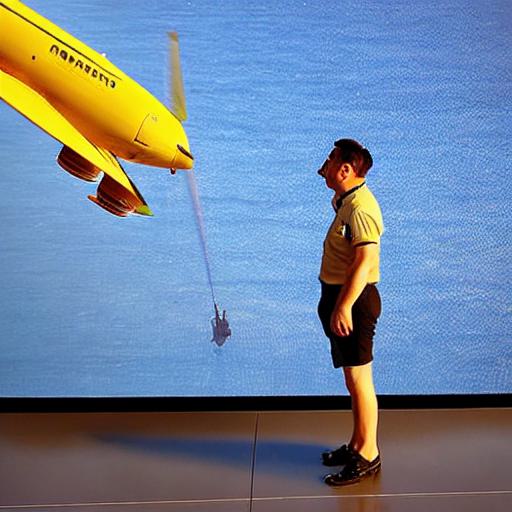}}
        & \raisebox{-0.5\height}{\includegraphics[width=.104\linewidth]{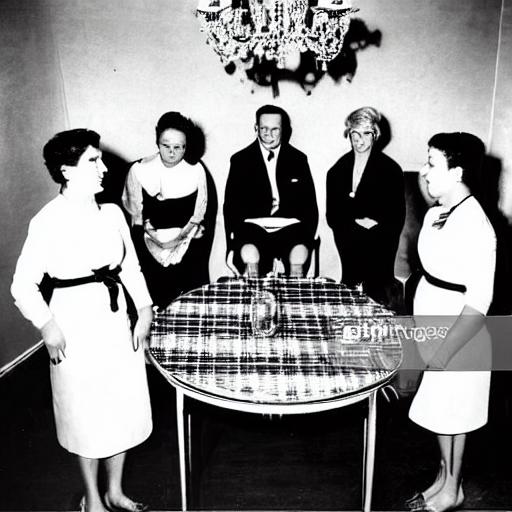}}
        & \raisebox{-0.5\height}{\includegraphics[width=.104\linewidth]{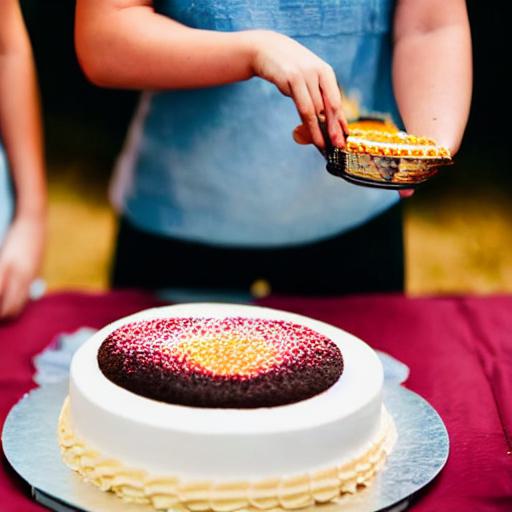}}
        & \raisebox{-0.5\height}{\includegraphics[width=.104\linewidth]{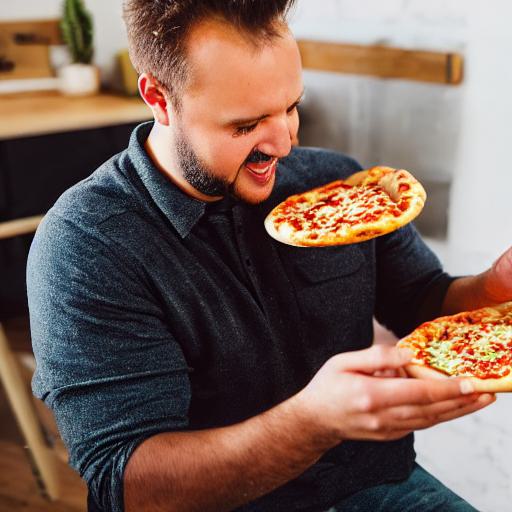}}
        & \raisebox{-0.5\height}{\includegraphics[width=.104\linewidth]{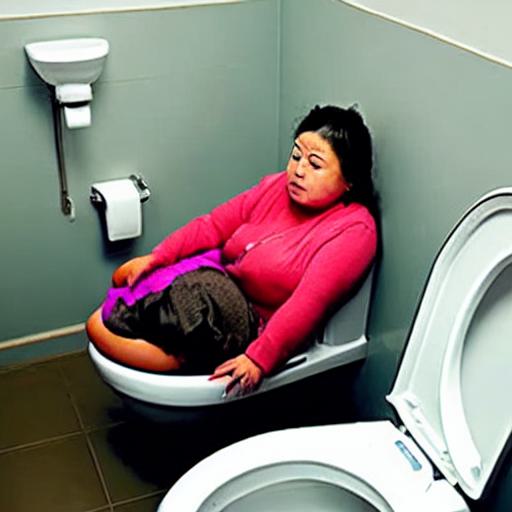}}\\
        \\[-1em]
        \rotatebox[origin=c]{90}{GLIGEN*}
        & \raisebox{-0.5\height}{\includegraphics[width=.104\linewidth]{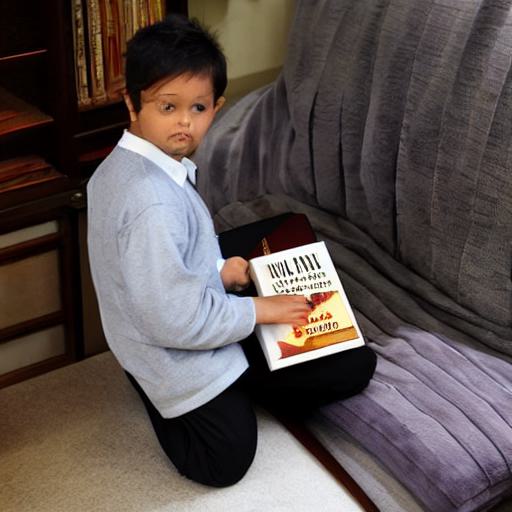}}
        & \raisebox{-0.5\height}{\includegraphics[width=.104\linewidth]{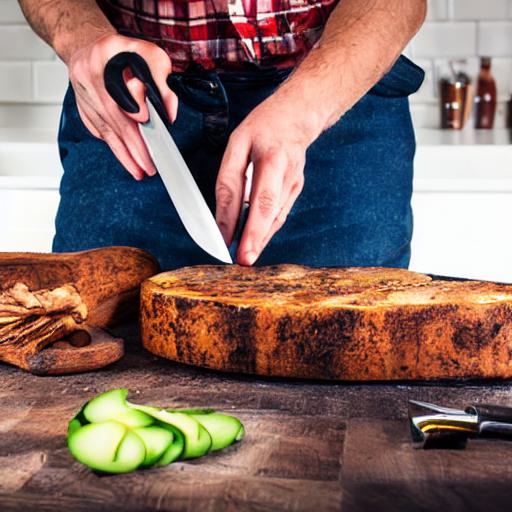}}
        & \raisebox{-0.5\height}{\includegraphics[width=.104\linewidth]{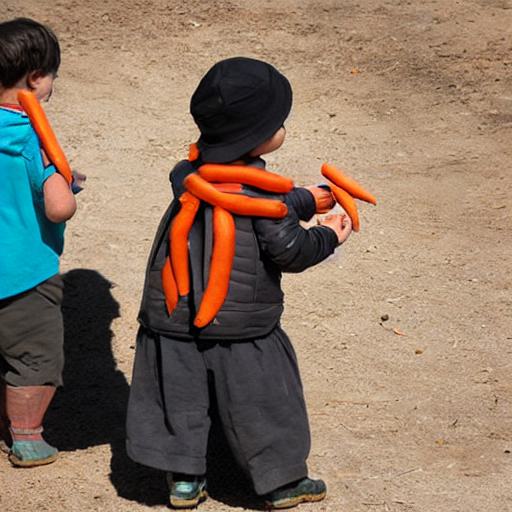}}
        & \raisebox{-0.5\height}{\includegraphics[width=.104\linewidth]{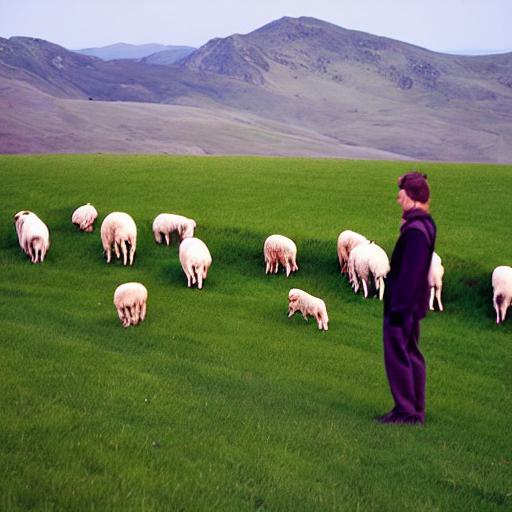}}
        & \raisebox{-0.5\height}{\includegraphics[width=.104\linewidth]{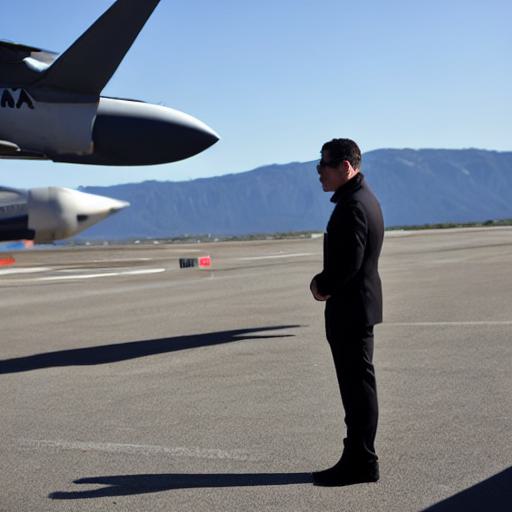}}
        & \raisebox{-0.5\height}{\includegraphics[width=.104\linewidth]{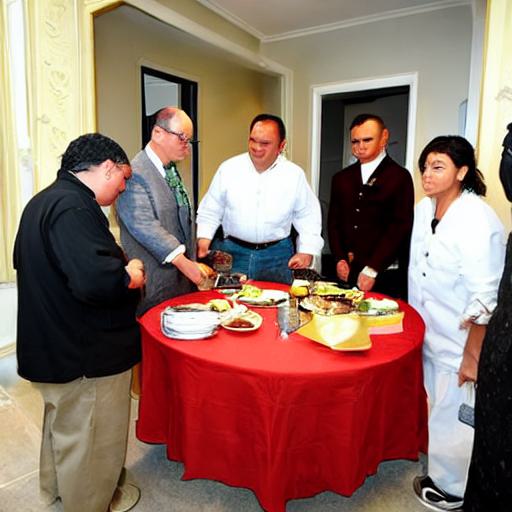}}
        & \raisebox{-0.5\height}{\includegraphics[width=.104\linewidth]{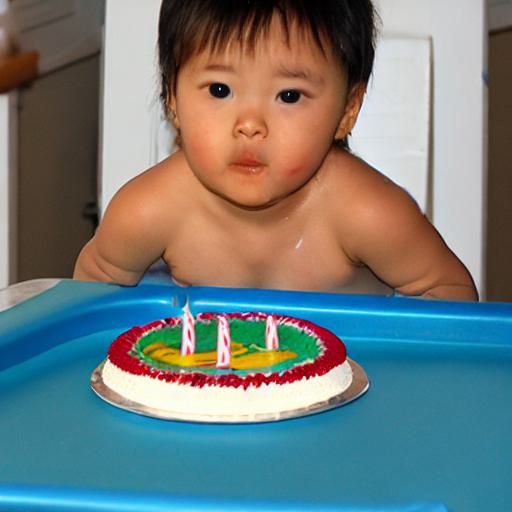}}
        & \raisebox{-0.5\height}{\includegraphics[width=.104\linewidth]{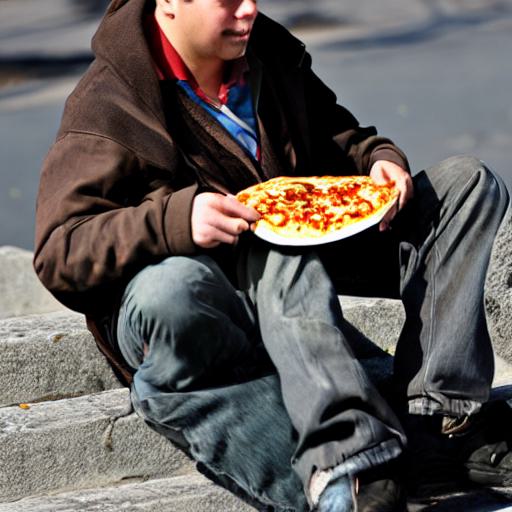}}
        & \raisebox{-0.5\height}{\includegraphics[width=.104\linewidth]{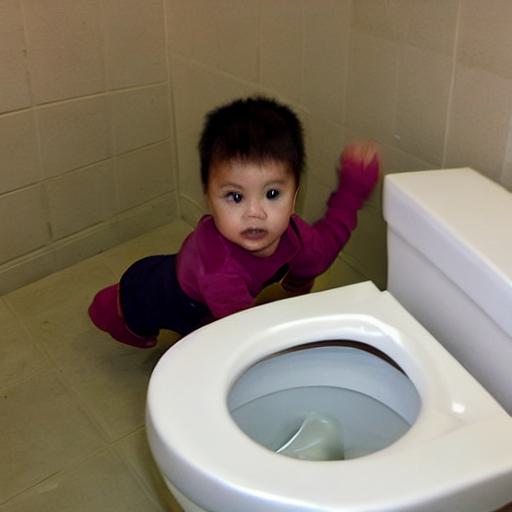}}\\
        \\[-1em]
        \rotatebox[origin=c]{90}{Ours}
        & \raisebox{-0.5\height}{\includegraphics[width=.104\linewidth]{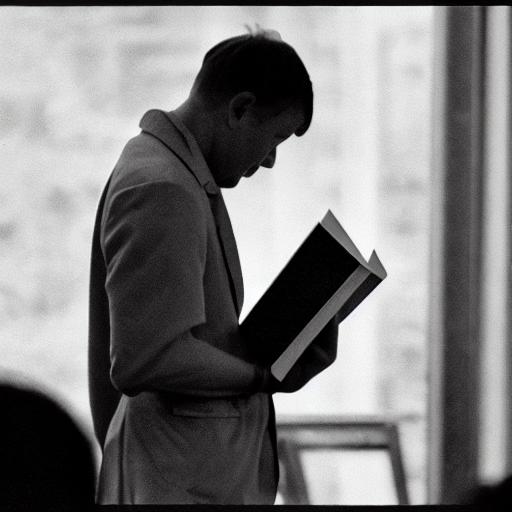}}
        & \raisebox{-0.5\height}{\includegraphics[width=.104\linewidth]{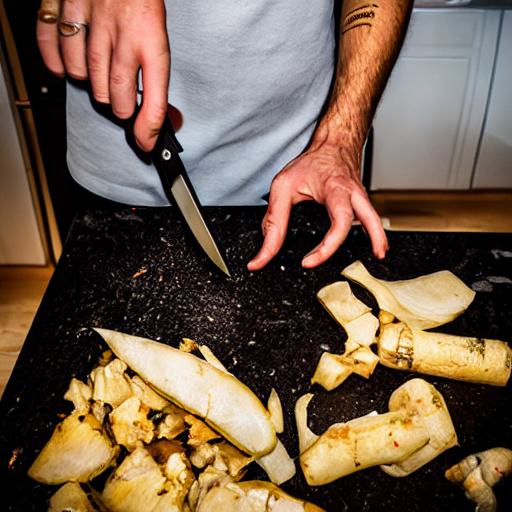}}
        & \raisebox{-0.5\height}{\includegraphics[width=.104\linewidth]{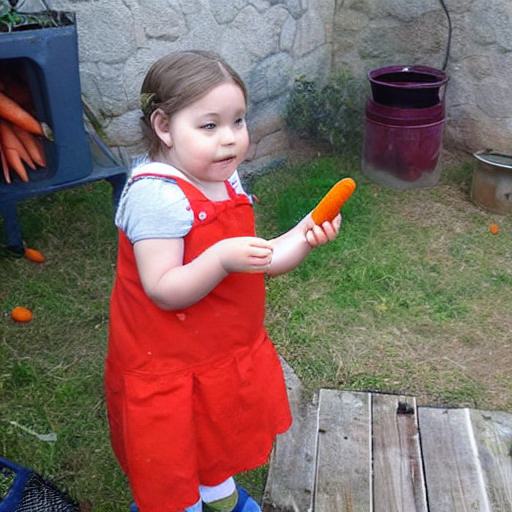}}
        & \raisebox{-0.5\height}{\includegraphics[width=.104\linewidth]{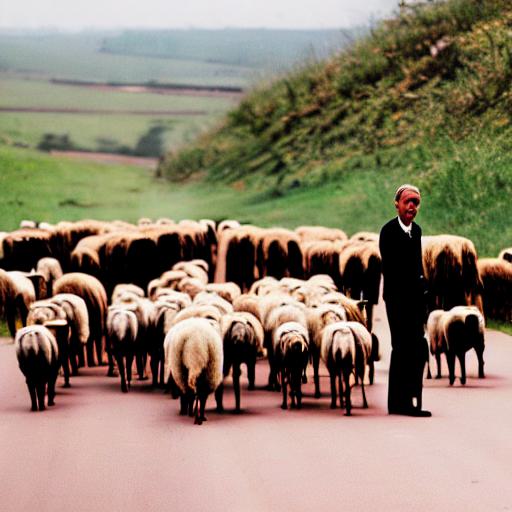}}
        & \raisebox{-0.5\height}{\includegraphics[width=.104\linewidth]{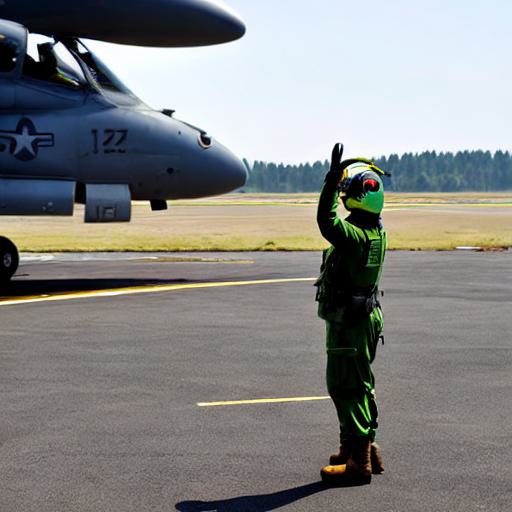}}
        & \raisebox{-0.5\height}{\includegraphics[width=.104\linewidth]{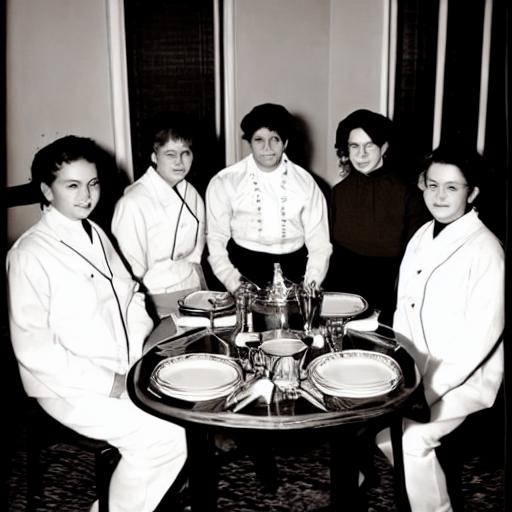}}
        & \raisebox{-0.5\height}{\includegraphics[width=.104\linewidth]{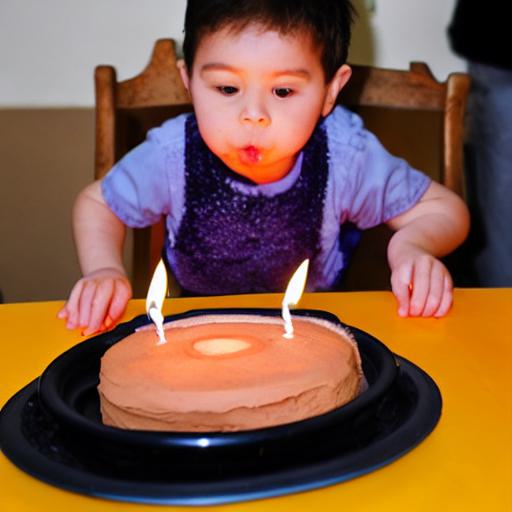}}
        & \raisebox{-0.5\height}{\includegraphics[width=.104\linewidth]{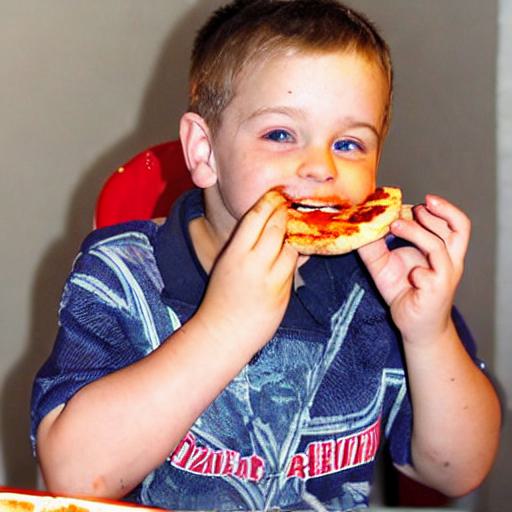}}
        & \raisebox{-0.5\height}{\includegraphics[width=.104\linewidth]{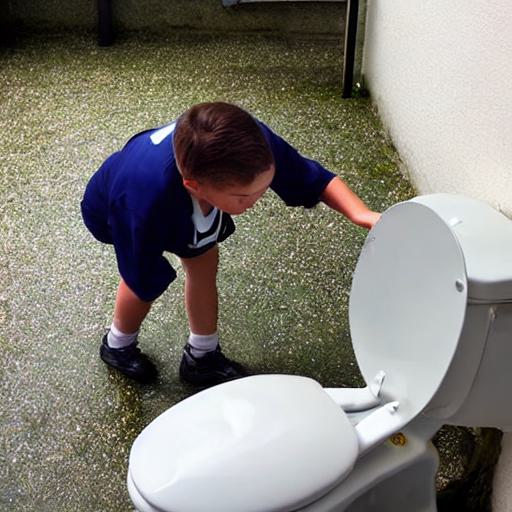}}\\
        & (a) & (b) & (c) & (d) & (e) & (f) & (g) & (h) & (i)\\
    \end{tabular}
    \vspace{-5pt}
    \captionsetup{type=figure}
    \caption{Visual comparison with existing baselines. In all methods, we use the text caption format of "a person {\it \{action\}} a {\it \{object\}}". Input and Caption rows represent the interaction conditions, each interaction pair shown by a line link them and is colored differently. GT represents the ground truth images. Ours gains better control to interaction, and renders images matching the text instructions better.}
    \label{fig:qualitative}
    \vspace{-10pt}
\end{table*}
\cref{fig:qualitative} presents a qualitative comparison with existing methods. The results demonstrate that our model renders the interaction relationship between objects better than others, aligning better with the provided interaction instructions. Other models often exhibit either mismatched actions or inaccurate interactions. For instance, while GLIGEN incorporates layout control to precisely position objects within an image, it fails to capture their intricate interactions. Especially, when multiple interaction instances occur within an image, GLIGEN's rendering of interaction relationships is often mismatched. This challenge persists even in the case of GLIGEN* which is fine-tuned on HICO-DET.

While the individual placement (location) of objects is accurate, the interactions between objects appear perplexing. Our proposed facilitates improved control over object interaction in image generation. For instance, in \cref{fig:qualitative}(a)-(c), although the interaction appears to be correct in existing works, the interaction details are inaccurate. Our proposed approach better renders these details. Moreover, when multiple interacting pairs are involved, as shown in \cref{fig:qualitative}(d), only our proposed is capable of correctly rendering all pairs of interactions. In \cref{fig:qualitative}(e)-(i), while the interactions (\eg directing airplane, sitting at the dining table, blowing cake, eating pizza, flushing the toilet) in images were inaccurately generated in existing works, our InteractDiffusion well renders these interactions. Our model's capability stems from two key components: the InToken for translating interaction conditions into meaningful tokens, and the InBedding for modeling complex interaction relationships. 

\cref{fig:diff_action2} shows how InteractDiffusion renders different actions with the same object, in comparison to StableDiffusion and GLIGEN*. This shows that our model can generate various combinations of interactions that maintain the coherence and naturalness of interactions between people and objects. More qualitative results are shown in \cref{sec:more_qualitative,fig:diff_object,fig:diff_action} of the supplementary.

\begin{table}[t]
    \centering
    \setlength{\tabcolsep}{0.5pt} 
    \renewcommand{\arraystretch}{1} 
    \begin{tabular}{ccccccc}
        & \raisebox{-0.35\height}{\includegraphics[width=.23\linewidth]{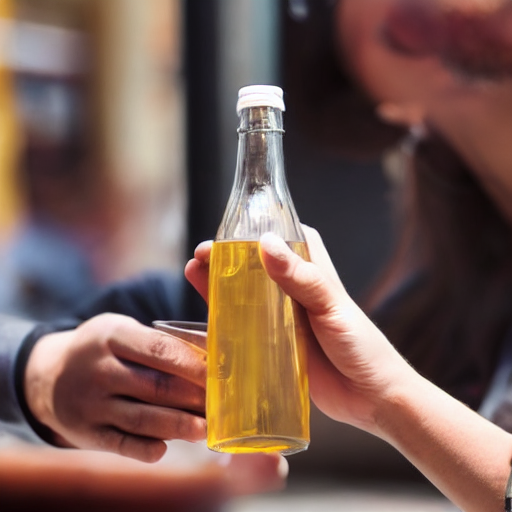}}
        & \raisebox{-0.35\height}{\includegraphics[width=.23\linewidth]{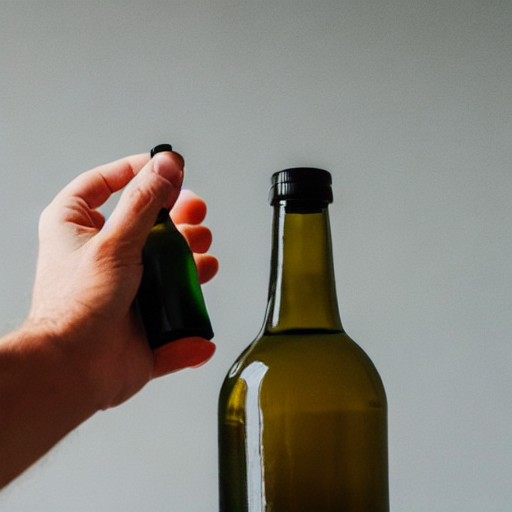}}
        & \raisebox{-0.35\height}{\includegraphics[width=.23\linewidth]{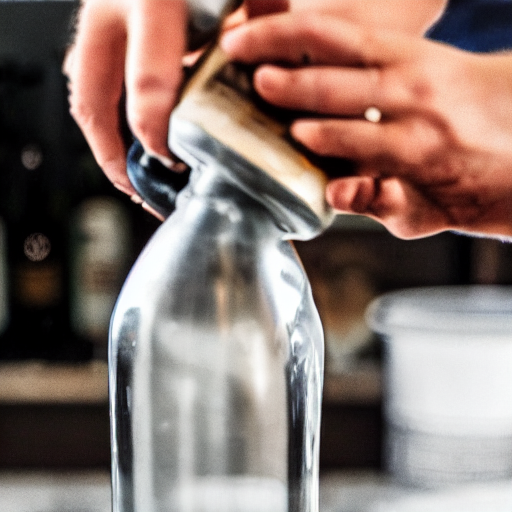}} 
        \\
        \raisebox{-0.35\height}{\frame{\includegraphics[width=.23\linewidth]{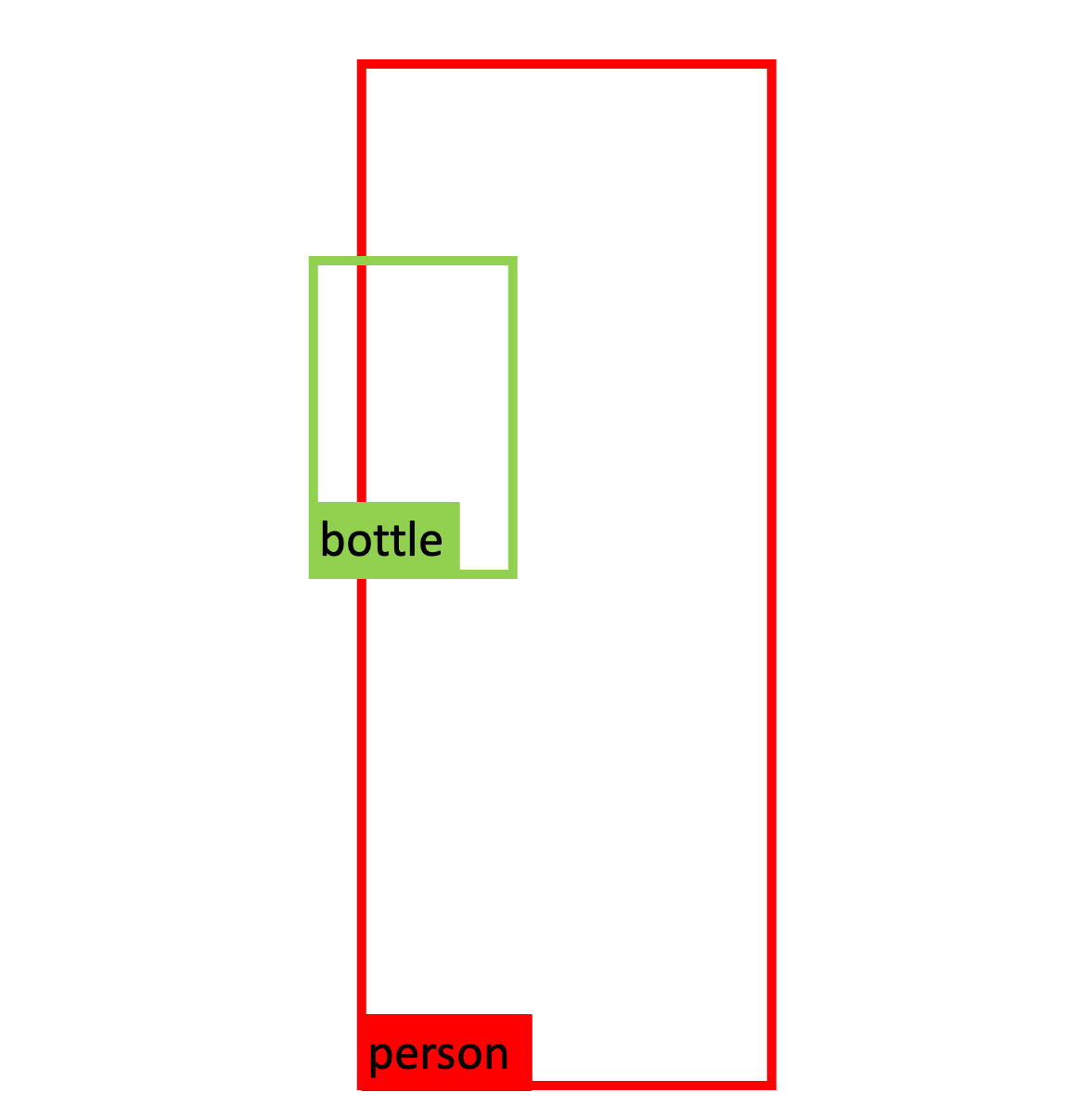}}}
        & \raisebox{-0.35\height}{\includegraphics[width=.23\linewidth]{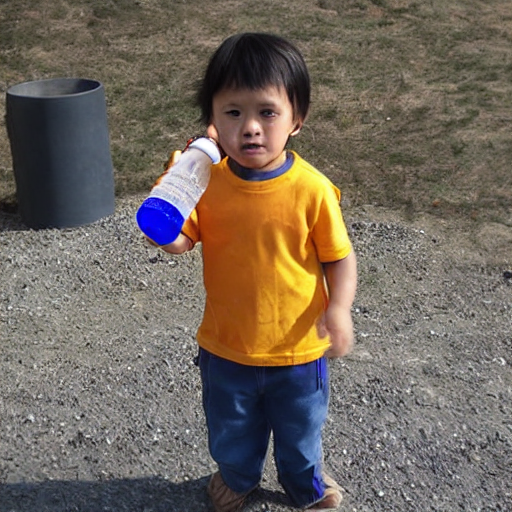}} 
        & \raisebox{-0.35\height}{\includegraphics[width=.23\linewidth]{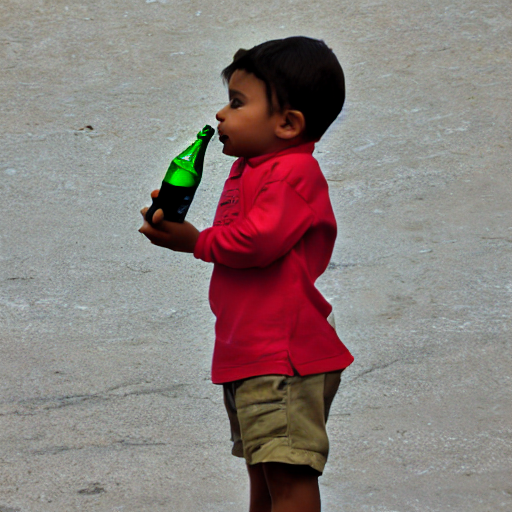}}
        & \raisebox{-0.35\height}{\includegraphics[width=.23\linewidth]{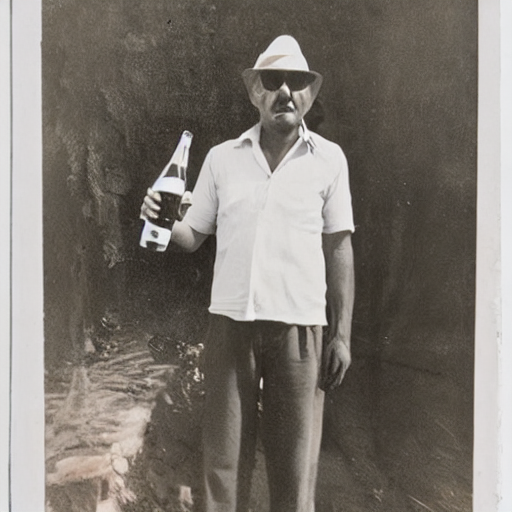}} 
        \\
        \raisebox{-0.35\height}{\frame{\includegraphics[width=.23\linewidth]{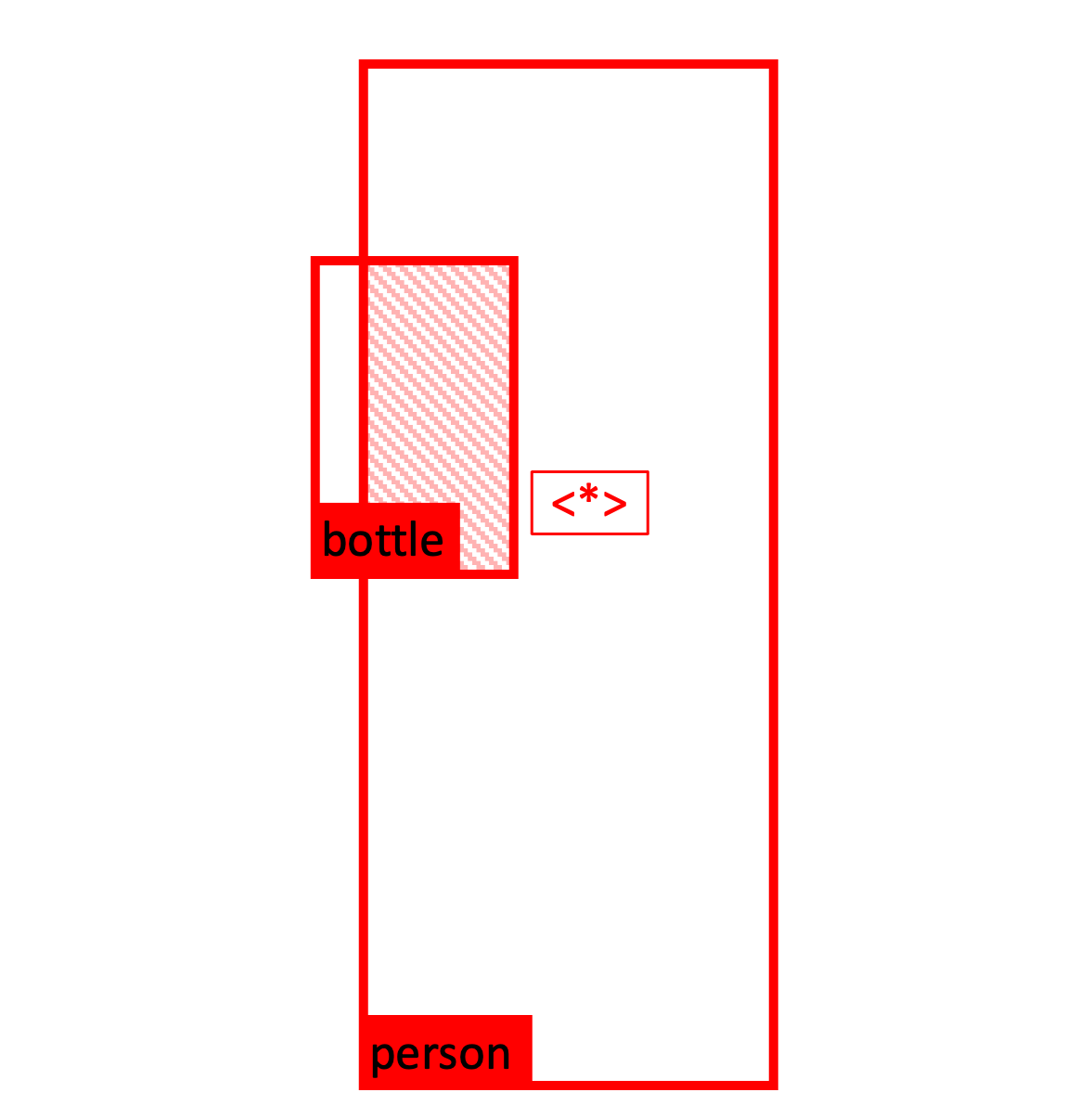}}}
        & \raisebox{-0.35\height}{\includegraphics[width=.23\linewidth]{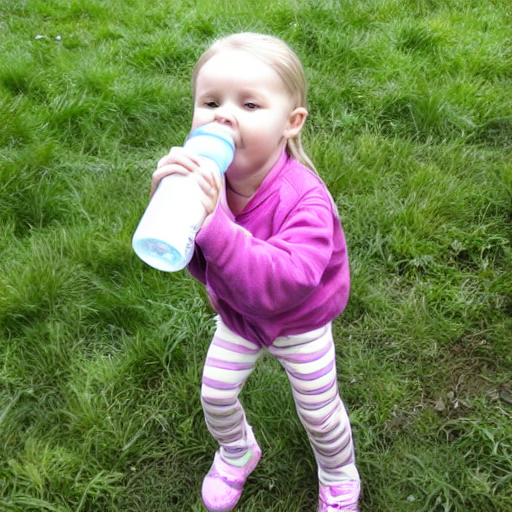}} 
        & \raisebox{-0.35\height}{\includegraphics[width=.23\linewidth]{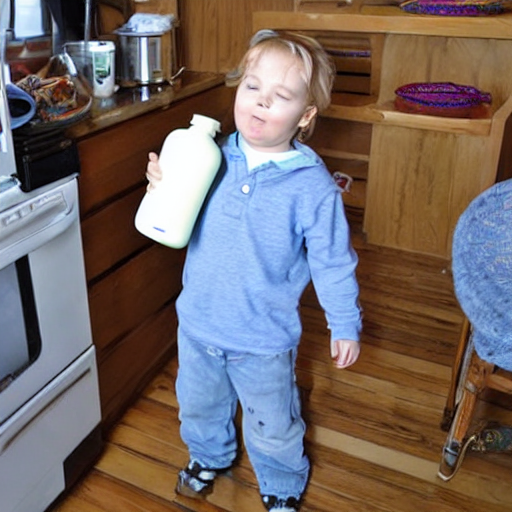}}
        & \raisebox{-0.35\height}{\includegraphics[width=.23\linewidth]{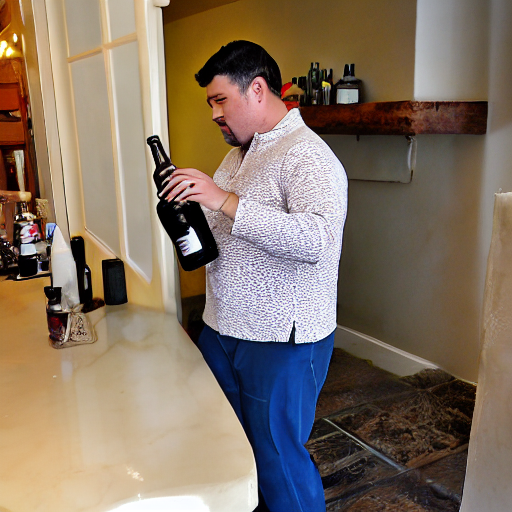}}
        \\
        \footnotesize a person is \textlangle*\textrangle a bottle
        & drinking
        & holding
        & pouring
    \end{tabular}
    \captionsetup{type=figure}
    \caption{Visualization comparison between StableDiffusion (top), GLIGEN* (middle), and InteractDiffusion (bottom)  demonstrates the generation of {\it different actions} for the same object.}
    \label{fig:diff_action2}
    \vspace{-10pt}
\end{table}

\subsection{Quantitative results}
\cref{tab:fid_map} compares our proposed with existing baselines in terms of the quality and interaction controllability, specifically FID, KID, and HOI Detection Score. Compared to the existing baselines, our proposed achieves the best result. 

For image generation quality, our proposed produces slightly higher quality than the baselines. It shows that despite additional parameters incorporated into the original model to control interactions, the image generation quality remains unaffected. It is even improved marginally. GLIGEN* exhibits higher image generation quality than StableDiffusion and GLIGEN because we fine-tuned it on the HICO-DET dataset in the same way as InteractDiffusion.

In terms of the HOI Detection Score, StableDiffusion performs poorly in this benchmark because it does not consider the object's location and size. Comparing GLIGEN and GLIGEN* that only consider the object's location and size, our method encodes the interaction control information along with localization information, leading to a significant performance gain.

Using the Tiny backbone for detection, the slight disparity in mAP between the generated images by our method and the real image dataset demonstrates that our approach can generate realistic interactions nearly indistinguishable from real-world interactions by a detection algorithm, such as FGAHOI with a Swin-Tiny backbone. Yet, we have observed that the gap between the real dataset and the generated samples widens when a detector of a large model is used. This indicates that although our generation process outperforms existing baselines, it still has room for further improvement in rendering finer details.

Empirically, the results demonstrate that our proposed enhances interaction controllability while maintaining high-quality image generation capability, thereby significantly outperforming the existing methods in all metrics. This superior performance can be attributed to the proposed components within InteractDiffusion, which include the \textit{InToken} that incorporates new interaction conditions, \textit{InBedding} that encode intricate interaction relationships, and the \textit{InFormer} that injects interaction control into the existing transformer blocks. Collectively, these components constitute a pluggable Interaction Module seamlessly integrated into existing T2I diffusion models.

\subsection{Ablation studies}
There are three key components that constitute the proposed InteractDiffusion, namely, InToken, InBedding, and InFormer. We conducted an ablation study on these components and tabulated the results in \cref{tab:ablation}. GLIGEN introduced a gated self-attention layer into the transformer block of the Stable Diffusion model to incorporate additional layout conditions, resulting in a significant performance improvement from 0.63 to 21.73 in mAP. Upon further fine-tuning on HICO-DET, it achieved an mAP of 25.23. 

In InteractDiffusion, we include interaction conditions, alongside layout conditions, to enable the interaction control. With InToken, we convert the interaction conditions (consisting of bounding boxes, object labels, action labels, and relationships) into meaningful interaction entity tokens. Compared to GLIGEN, the incorporation of additional action tokens introduces new information that enhances interaction generation and provides greater interaction control. The inclusion of InToken as a key component further improved the detection score from 25.23 to 28.73, thereby demonstrating its effectiveness. Lastly, we include InBedding to encode the complex interactions relationship, which further improved detection score from 28.73 to 29.53. More ablation studies are shown in \cref{sec:more-ablation} of the supplementary.
\section{Conclusion}
This paper proposes an interaction-conditioned T2I diffusion model, namely InteractDiffusion, which addresses problems of conditioning generated images beyond the text caption. In existing T2I diffusion models, although several controls (\eg text, images, layout, etc) have been imposed, controlling the interaction in the generated image remains a formidable challenge. Our contributions can be unified as a pluggable interaction module being seamlessly integrated into existing T2I models. The quantitative and qualitative evaluations demonstrate the effectiveness of our method in controlling the interaction of generated content, which significantly outperforms the state-of-the-art approaches.

{
    \small
    \bibliographystyle{ieeenat_fullname}
    \bibliography{main}
}

\clearpage
\setcounter{page}{1}
\maketitlesupplementary

\section{Implementation Details.}\label{sec:implementation_detail}
\noindent\textbf{Negative Prompt} We use the following negative prompt for all generation: ``longbody, lowres, bad anatomy, bad hands, missing fingers, extra digit, fewer digits, cropped, worst quality, low quality''.

\noindent\textbf{Model Complexity.} \cref{tab:complexity} shows the number of parameters in InteractDiffusion model in comparison with other diffusion-based baselines. The number of trainable parameters of InteractDiffusion is about 210 millions, only 1 millions more than GLIGEN, while introducing new interaction controllability.  Note that these parameters counts do not include the text encoder and the VAE, which are same for all methods.
\begin{table}[ht]
\centering
\begin{tabular}{l|c|c}
\hline
Method            & $N_\text{params}$ & $N_\text{trainable}$ \\ \hline
StableDiffusion   & 860M              & 860M                 \\
GLIGEN            & 1069M             & 209M                 \\
InteractDiffusion & 1070M             & 210M                 \\ \hline
\end{tabular}
\caption{Number of parameters for InteractDiffusion in comparison with other diffusion-based baselines.}
\label{tab:complexity}
\end{table}

\noindent\textbf{Network Architecture.} In all experiments, Stable Diffusion V1.4 is used as base model for all methods. We maintain the network architecture except the transformer block in U-Net was adapted to include our Interaction Module.

\section{Additional Ablation Studies}\label{sec:more-ablation}

\begin{table*}[!ht]
    \centering
    \setlength{\tabcolsep}{1pt} 
    \renewcommand{\arraystretch}{3.6} 
    \begin{tabular}{ccccccccc}
        $\omega=$ 0.0 & 0.1 & 0.2 & 0.3 & 0.4 & 0.6 & 0.8 & 1.0 & HICO-DET \\
        \raisebox{-0.5\height}{\includegraphics[width=.105\linewidth]{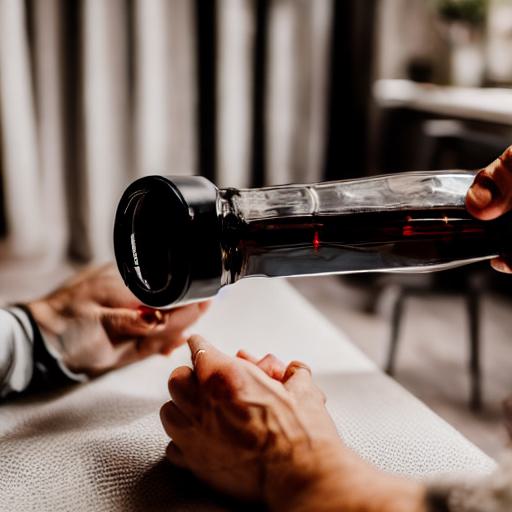}}
        & \raisebox{-0.5\height}{\includegraphics[width=.105\linewidth]{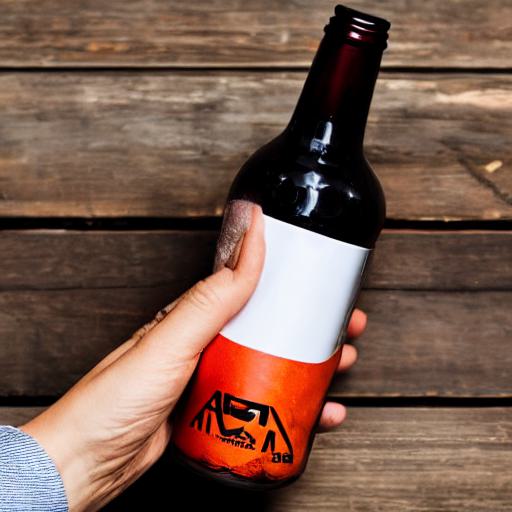}}
        & \raisebox{-0.5\height}{\includegraphics[width=.105\linewidth]{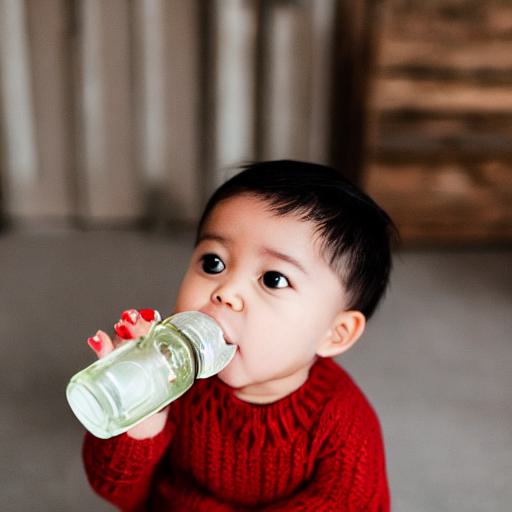}}
        & \raisebox{-0.5\height}{\includegraphics[width=.105\linewidth]{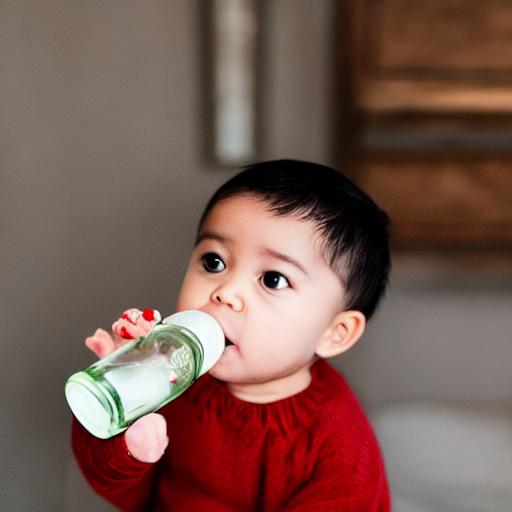}}
        & \raisebox{-0.5\height}{\includegraphics[width=.105\linewidth]{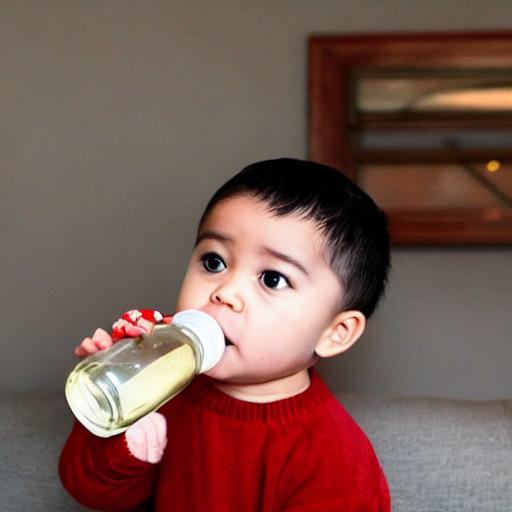}}
        & \raisebox{-0.5\height}{\includegraphics[width=.105\linewidth]{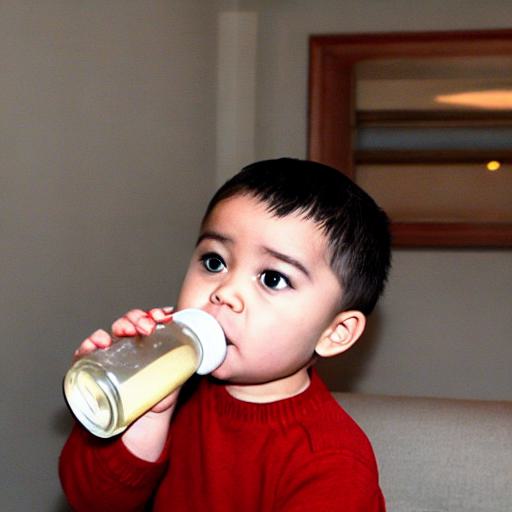}}
        & \raisebox{-0.5\height}{\includegraphics[width=.105\linewidth]{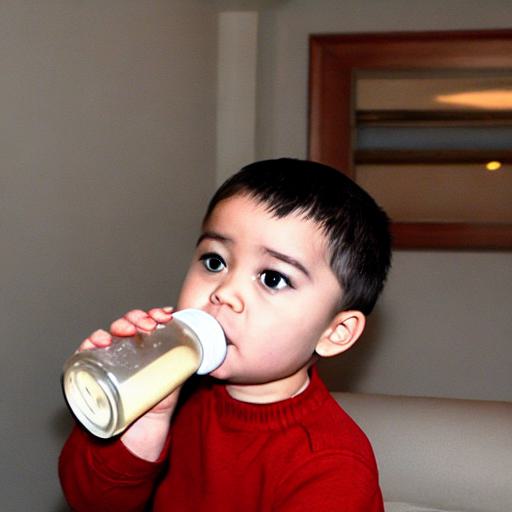}}
        & \raisebox{-0.5\height}{\includegraphics[width=.105\linewidth]{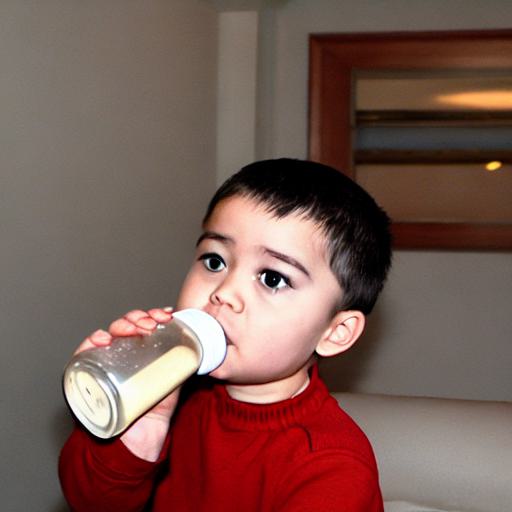}}
        & \raisebox{-0.5\height}{\includegraphics[width=.105\linewidth]{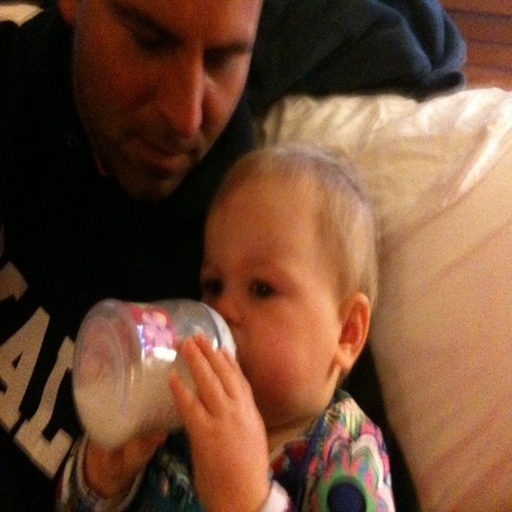}}\\
        \raisebox{-0.5\height}{\includegraphics[width=.105\linewidth]{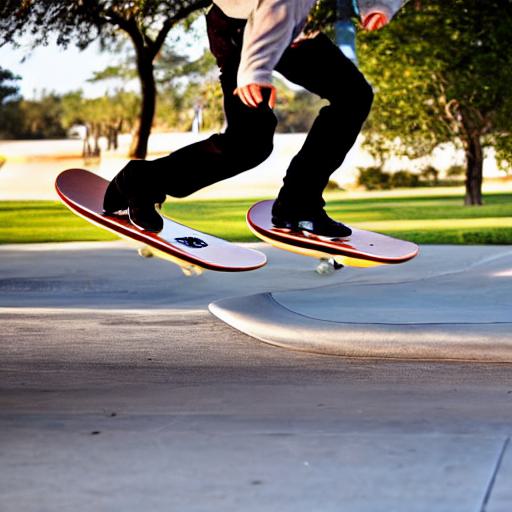}}
        & \raisebox{-0.5\height}{\includegraphics[width=.105\linewidth]{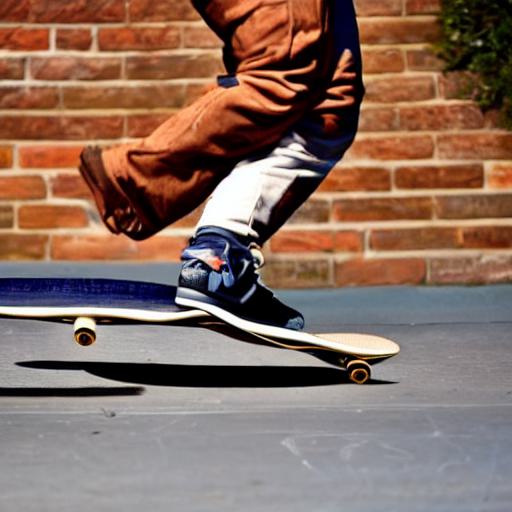}}
        & \raisebox{-0.5\height}{\includegraphics[width=.105\linewidth]{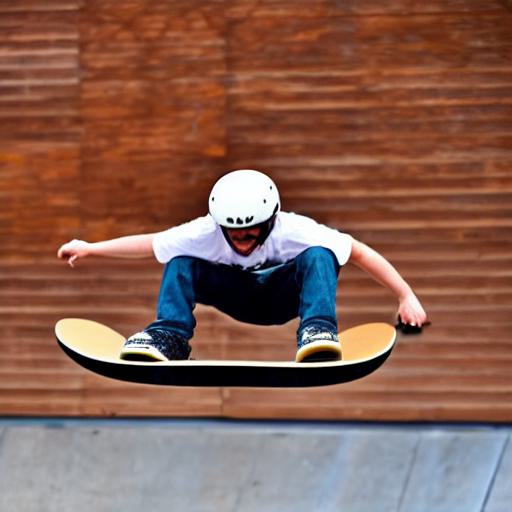}}
        & \raisebox{-0.5\height}{\includegraphics[width=.105\linewidth]{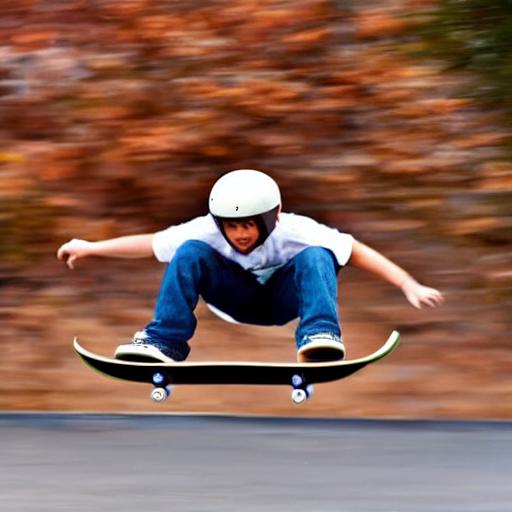}}
        & \raisebox{-0.5\height}{\includegraphics[width=.105\linewidth]{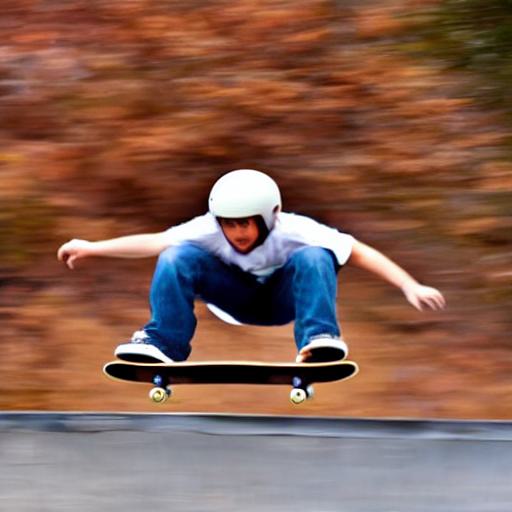}}
        & \raisebox{-0.5\height}{\includegraphics[width=.105\linewidth]{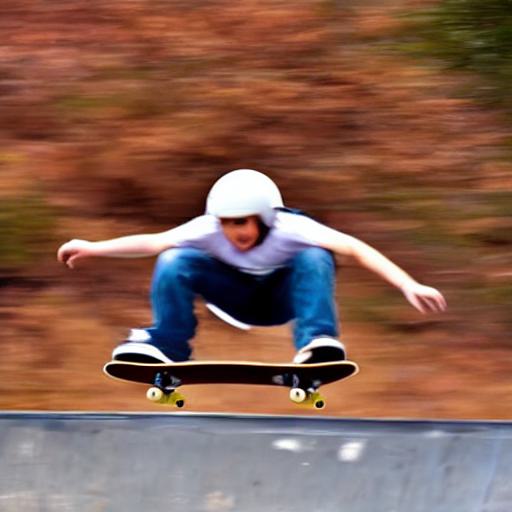}}
        & \raisebox{-0.5\height}{\includegraphics[width=.105\linewidth]{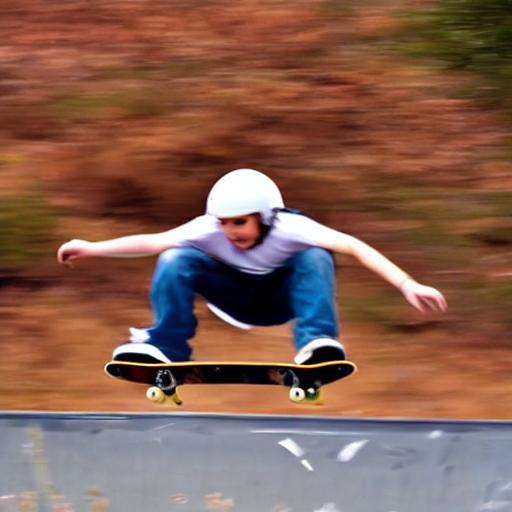}}
        & \raisebox{-0.5\height}{\includegraphics[width=.105\linewidth]{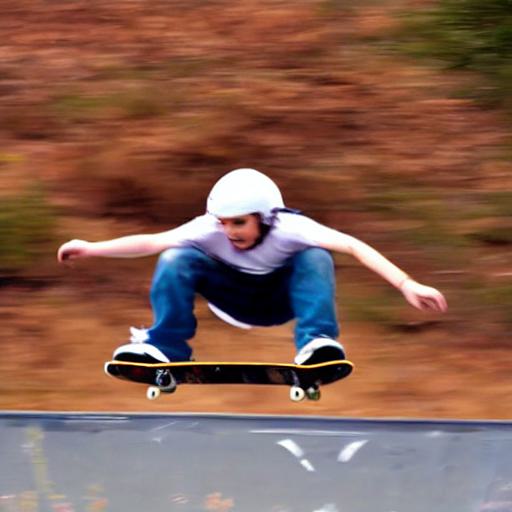}}
        & \raisebox{-0.5\height}{\includegraphics[width=.105\linewidth]{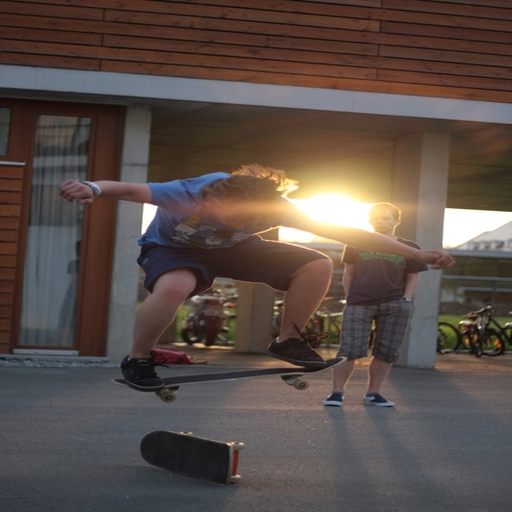}}\\
        \raisebox{-0.5\height}{\includegraphics[width=.105\linewidth]{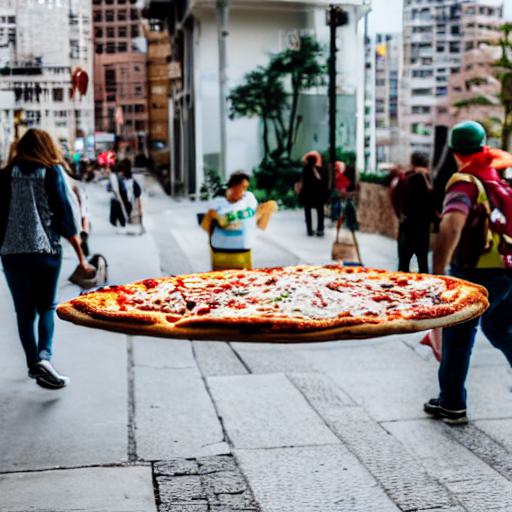}}
        & \raisebox{-0.5\height}{\includegraphics[width=.105\linewidth]{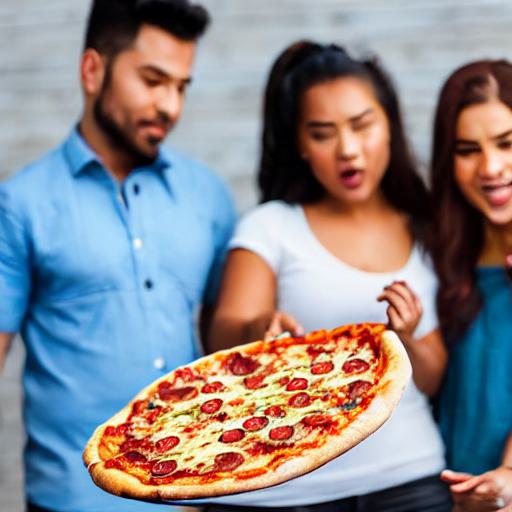}}
        & \raisebox{-0.5\height}{\includegraphics[width=.105\linewidth]{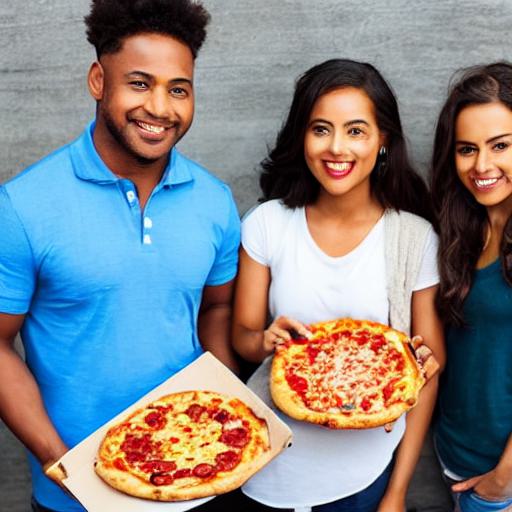}}
        & \raisebox{-0.5\height}{\includegraphics[width=.105\linewidth]{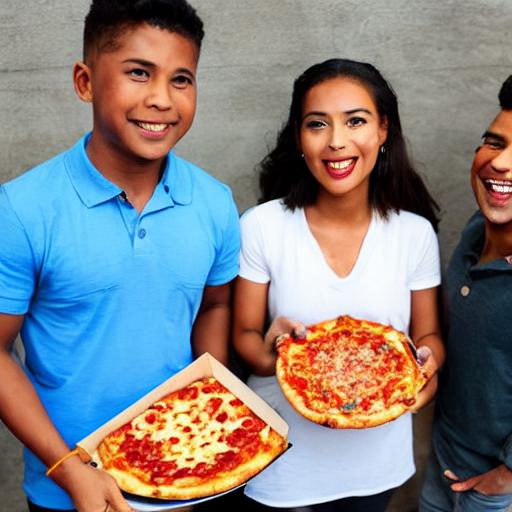}}
        & \raisebox{-0.5\height}{\includegraphics[width=.105\linewidth]{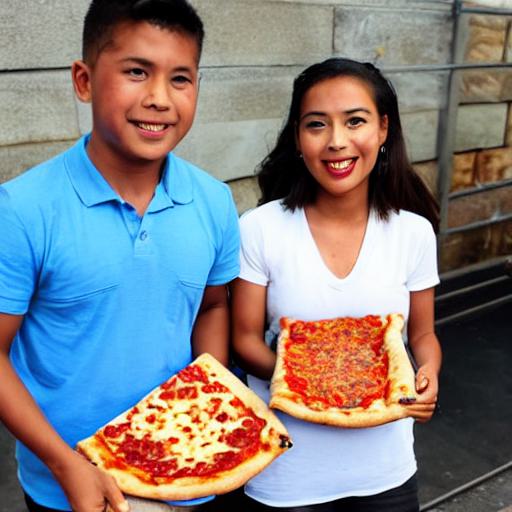}}
        & \raisebox{-0.5\height}{\includegraphics[width=.105\linewidth]{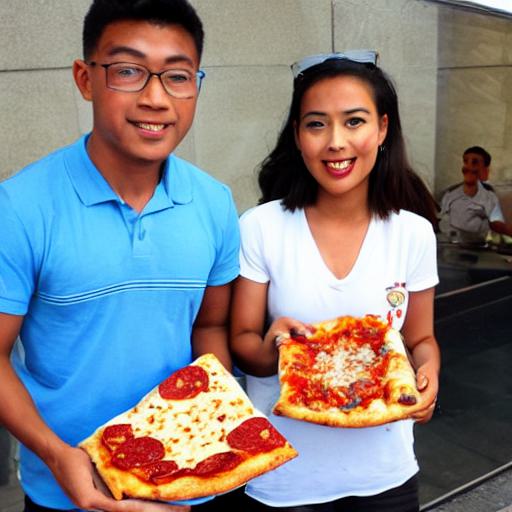}}
        & \raisebox{-0.5\height}{\includegraphics[width=.105\linewidth]{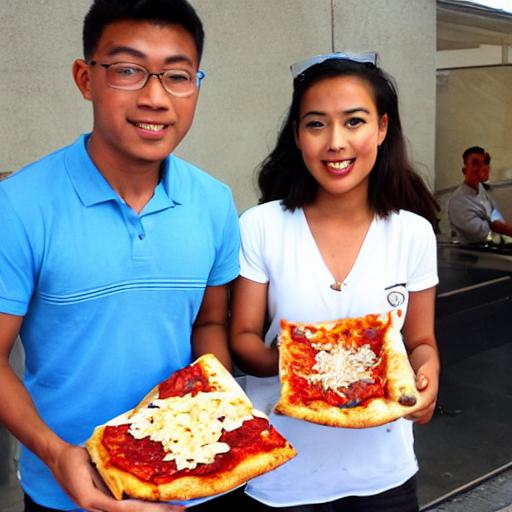}}
        & \raisebox{-0.5\height}{\includegraphics[width=.105\linewidth]{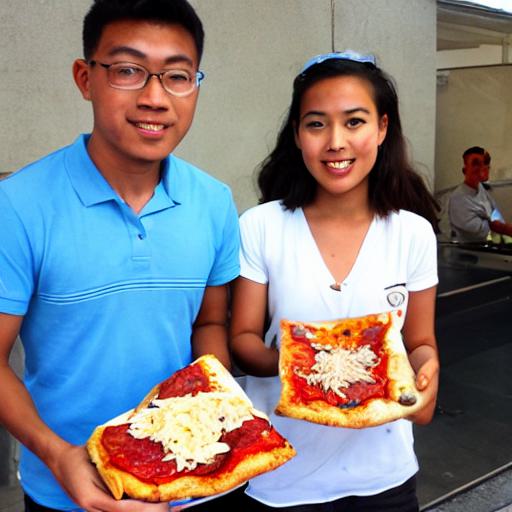}}
        & \raisebox{-0.5\height}{\includegraphics[width=.105\linewidth]{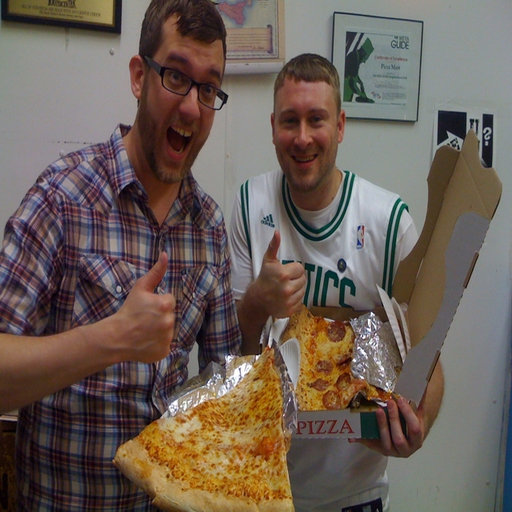}}\\
        \raisebox{-0.5\height}{\includegraphics[width=.105\linewidth]{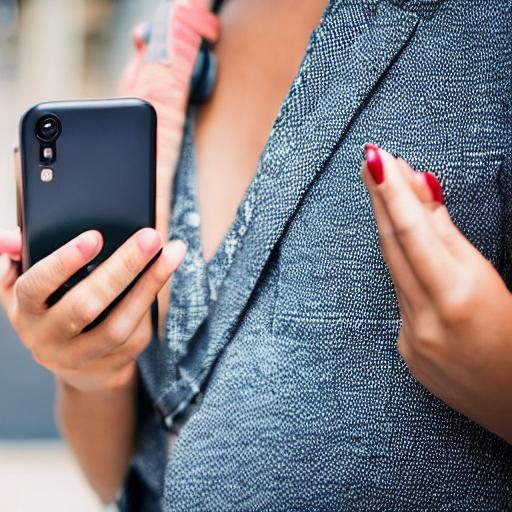}}
        & \raisebox{-0.5\height}{\includegraphics[width=.105\linewidth]{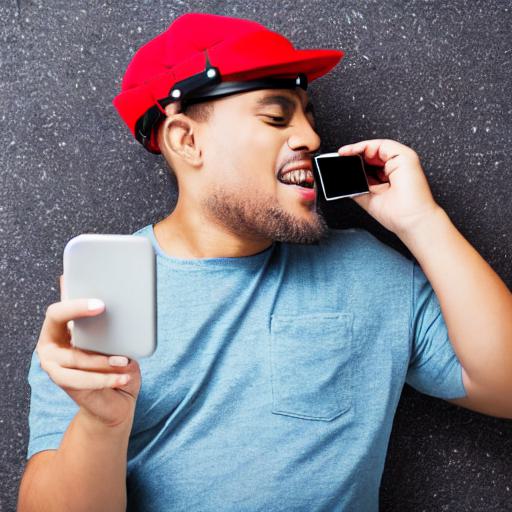}}
        & \raisebox{-0.5\height}{\includegraphics[width=.105\linewidth]{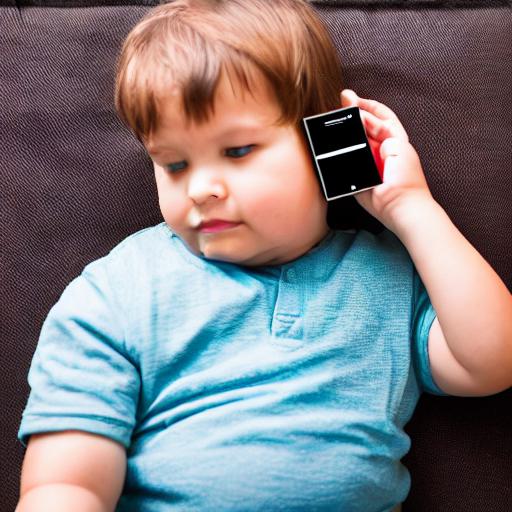}}
        & \raisebox{-0.5\height}{\includegraphics[width=.105\linewidth]{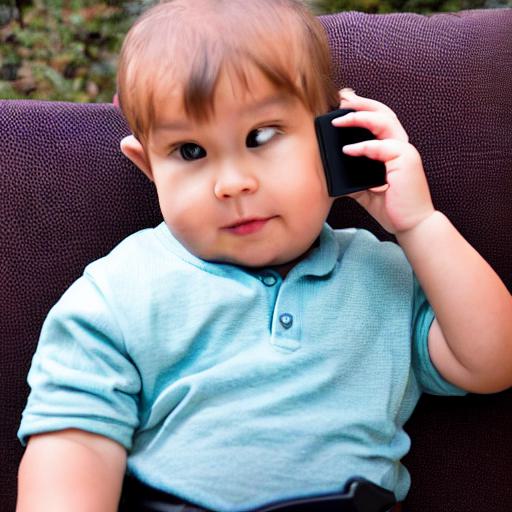}}
        & \raisebox{-0.5\height}{\includegraphics[width=.105\linewidth]{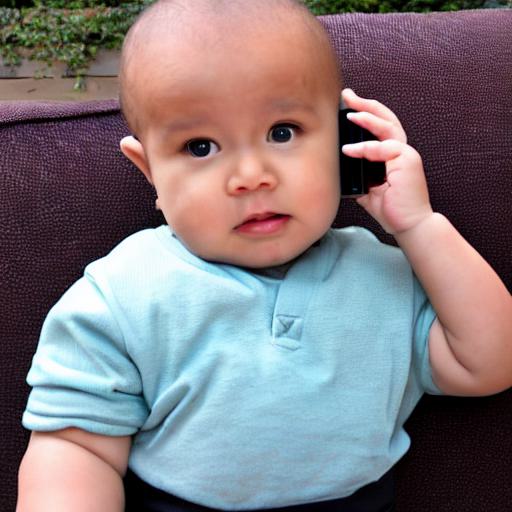}}
        & \raisebox{-0.5\height}{\includegraphics[width=.105\linewidth]{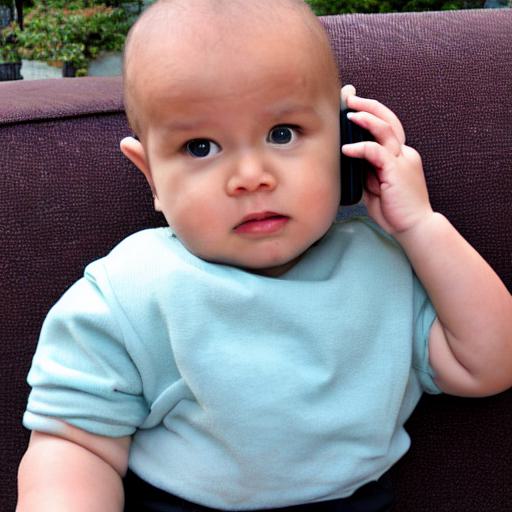}}
        & \raisebox{-0.5\height}{\includegraphics[width=.105\linewidth]{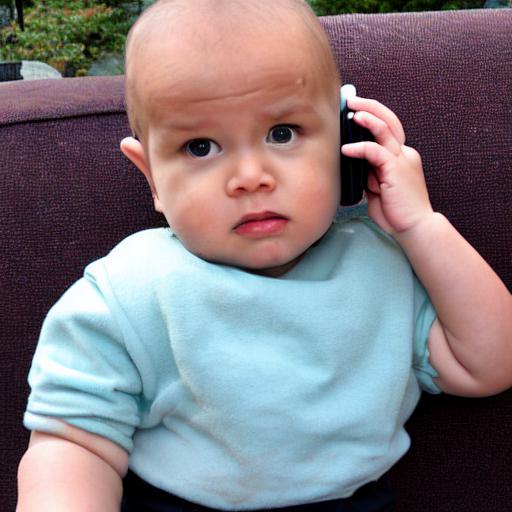}}
        & \raisebox{-0.5\height}{\includegraphics[width=.105\linewidth]{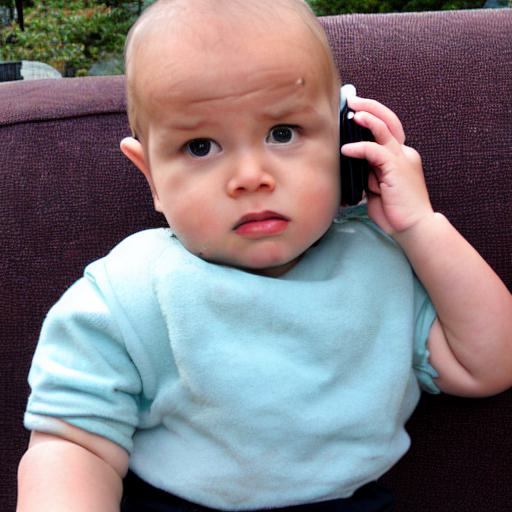}}
        & \raisebox{-0.5\height}{\includegraphics[width=.105\linewidth]{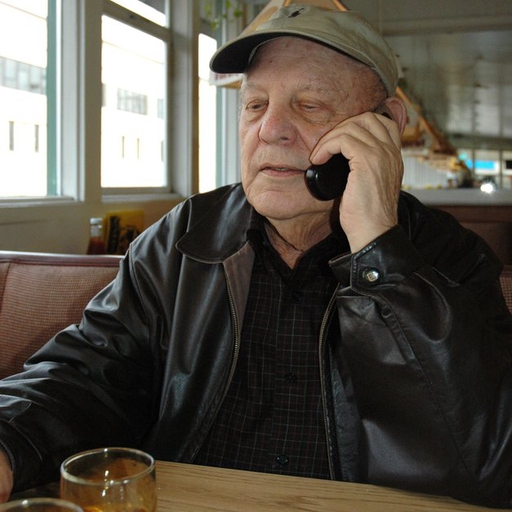}}\\
        \raisebox{-0.5\height}{\includegraphics[width=.105\linewidth]{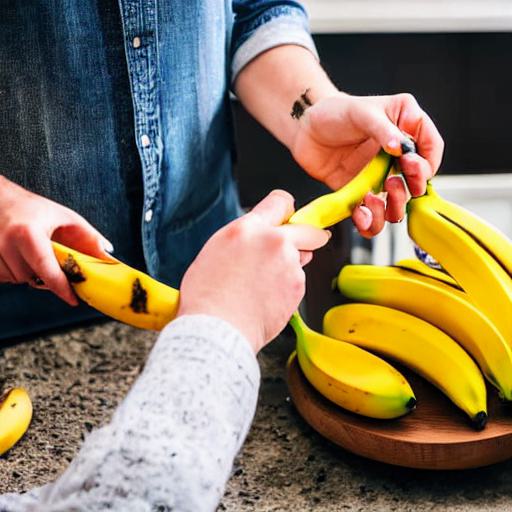}}
        & \raisebox{-0.5\height}{\includegraphics[width=.105\linewidth]{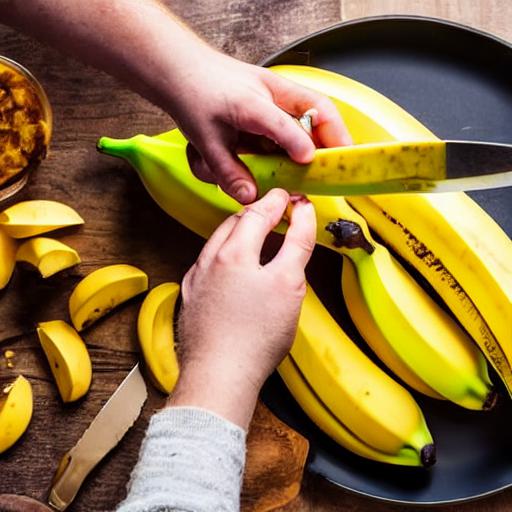}}
        & \raisebox{-0.5\height}{\includegraphics[width=.105\linewidth]{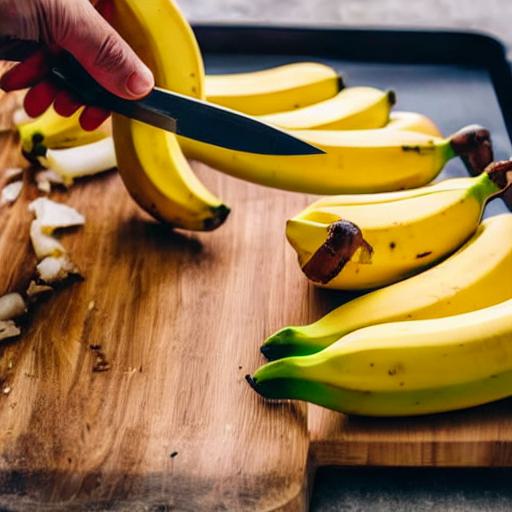}}
        & \raisebox{-0.5\height}{\includegraphics[width=.105\linewidth]{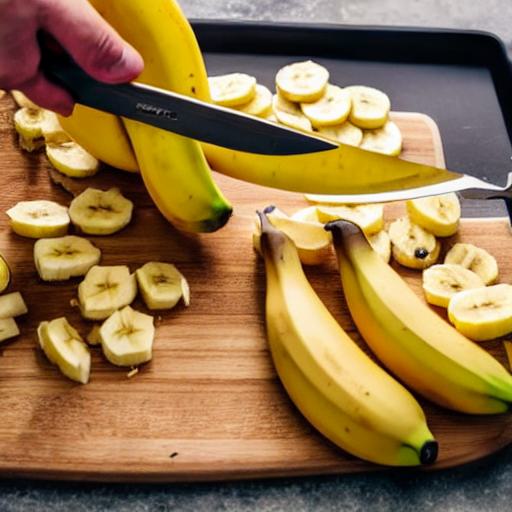}}
        & \raisebox{-0.5\height}{\includegraphics[width=.105\linewidth]{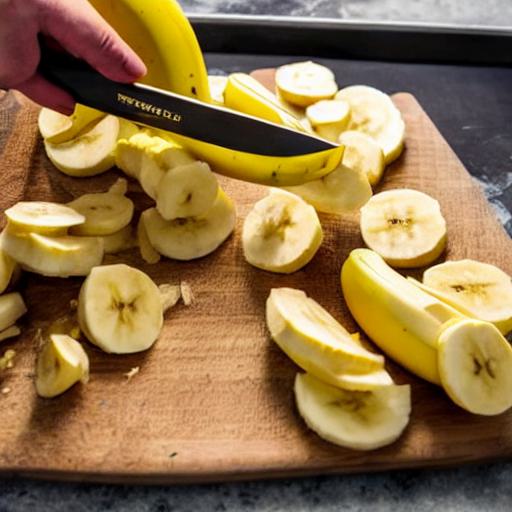}}
        & \raisebox{-0.5\height}{\includegraphics[width=.105\linewidth]{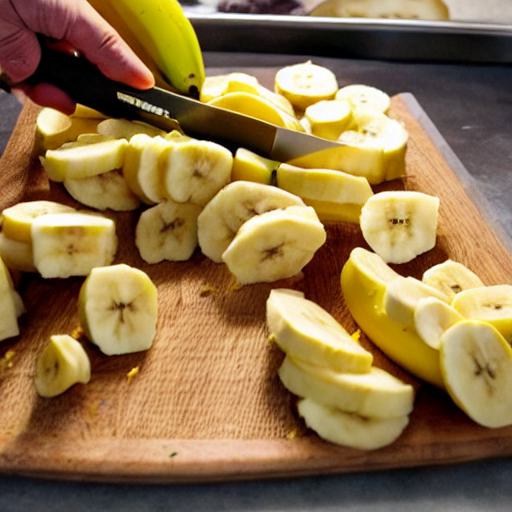}}
        & \raisebox{-0.5\height}{\includegraphics[width=.105\linewidth]{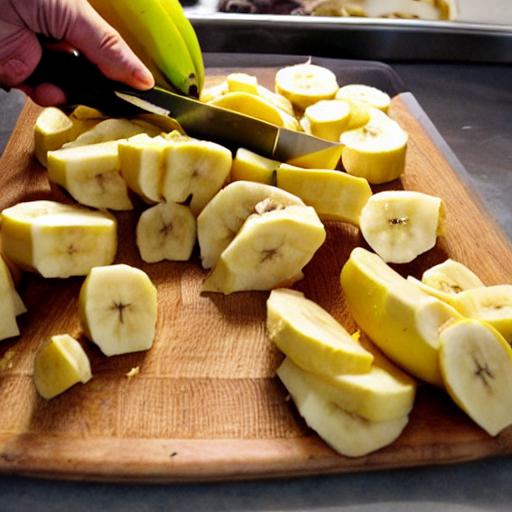}}
        & \raisebox{-0.5\height}{\includegraphics[width=.105\linewidth]{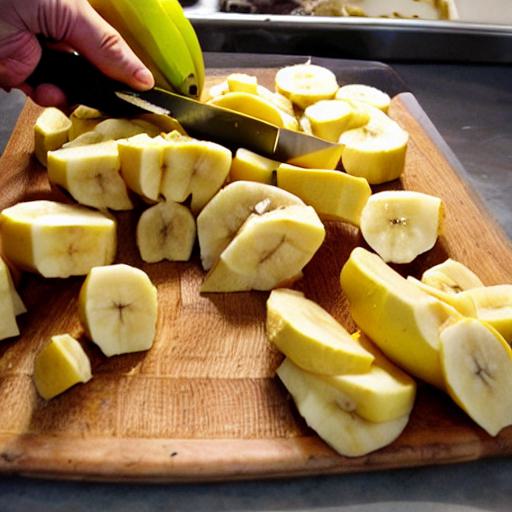}}
        & \raisebox{-0.5\height}{\includegraphics[width=.105\linewidth]{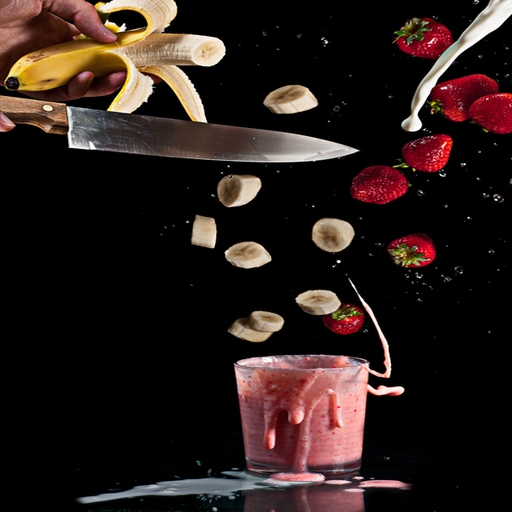}}\\
        \raisebox{-0.5\height}{\includegraphics[width=.105\linewidth]{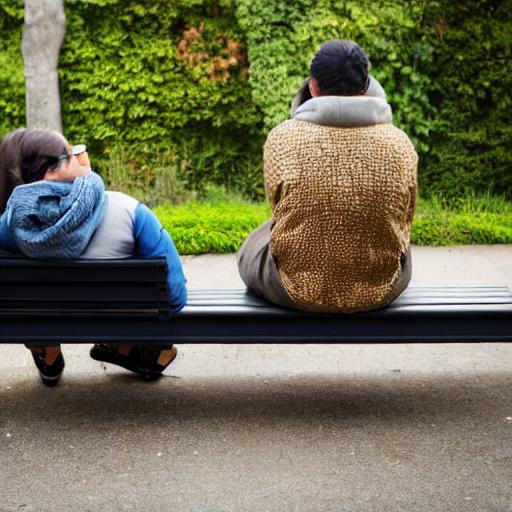}}
        & \raisebox{-0.5\height}{\includegraphics[width=.105\linewidth]{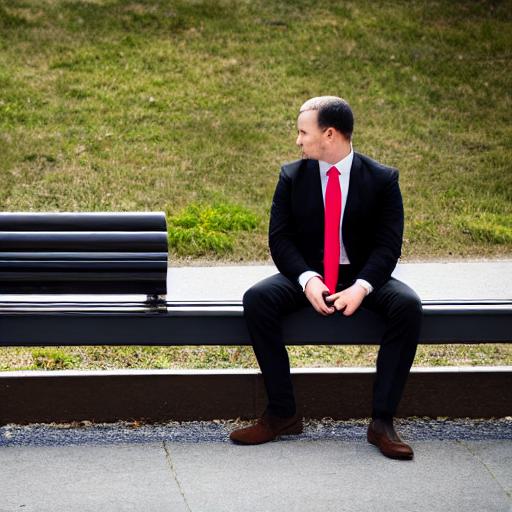}}
        & \raisebox{-0.5\height}{\includegraphics[width=.105\linewidth]{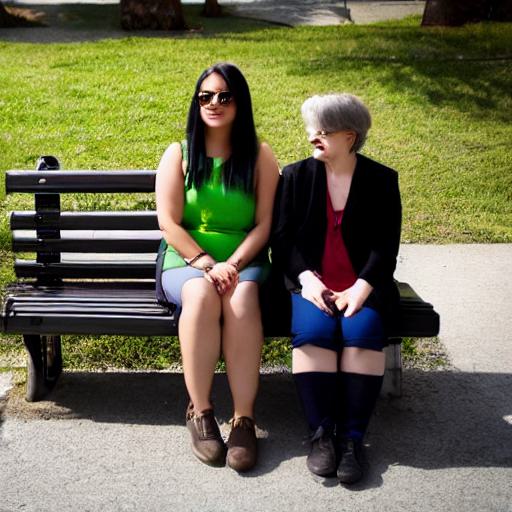}}
        & \raisebox{-0.5\height}{\includegraphics[width=.105\linewidth]{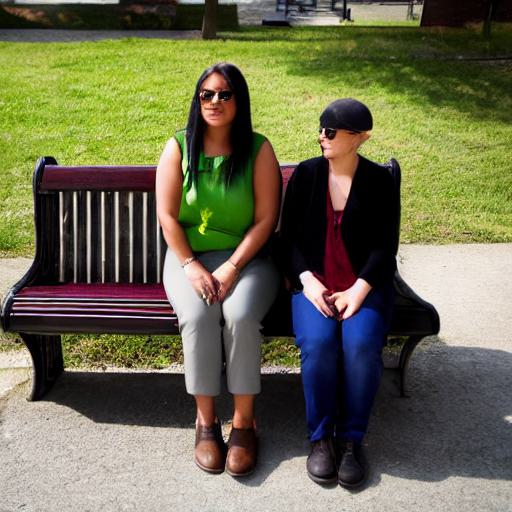}}
        & \raisebox{-0.5\height}{\includegraphics[width=.105\linewidth]{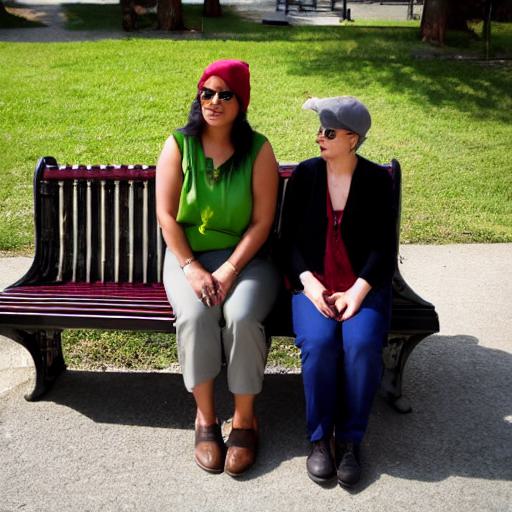}}
        & \raisebox{-0.5\height}{\includegraphics[width=.105\linewidth]{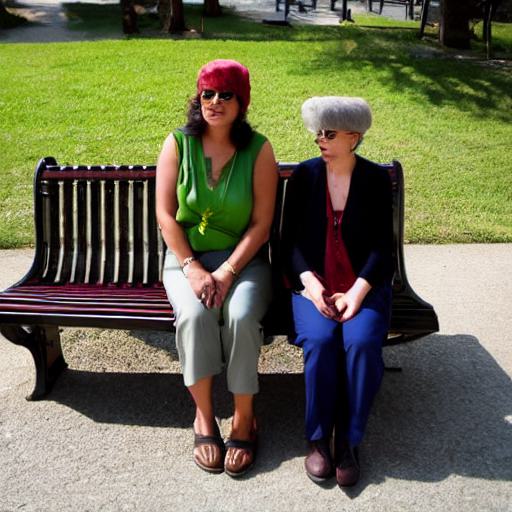}}
        & \raisebox{-0.5\height}{\includegraphics[width=.105\linewidth]{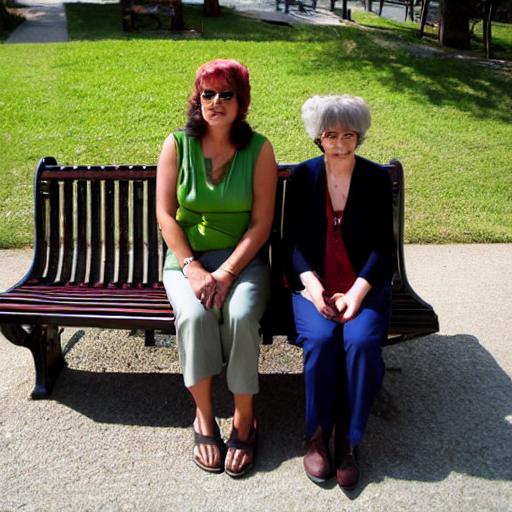}}
        & \raisebox{-0.5\height}{\includegraphics[width=.105\linewidth]{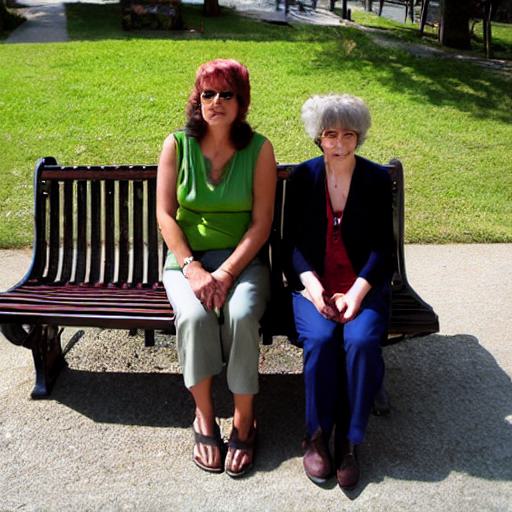}}
        & \raisebox{-0.5\height}{\includegraphics[width=.105\linewidth]{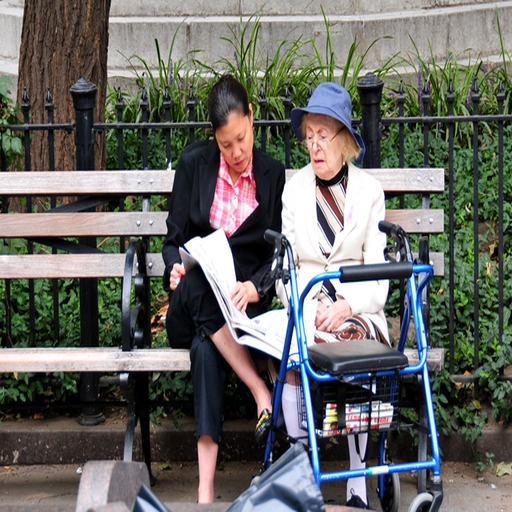}}\\
        \raisebox{-0.5\height}{\includegraphics[width=.105\linewidth]{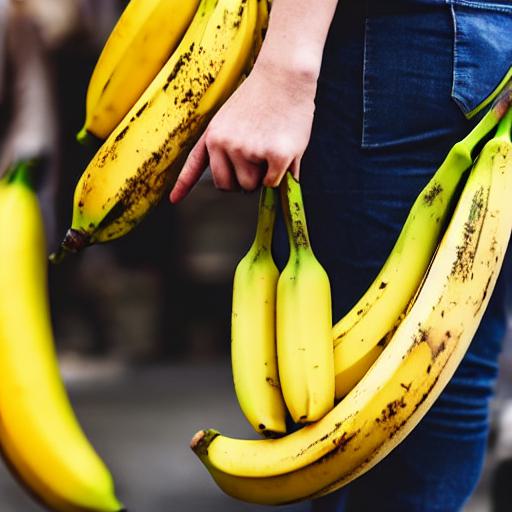}}
        & \raisebox{-0.5\height}{\includegraphics[width=.105\linewidth]{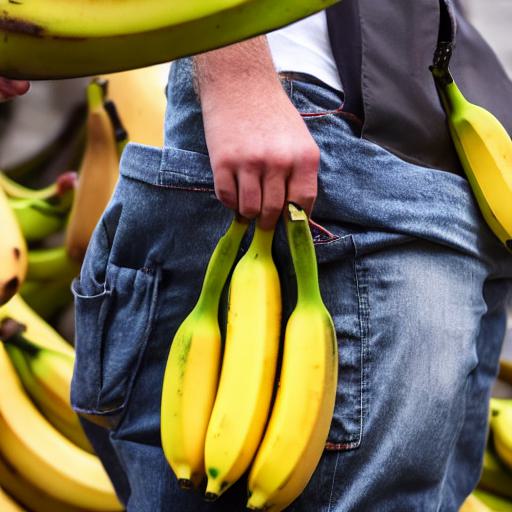}}
        & \raisebox{-0.5\height}{\includegraphics[width=.105\linewidth]{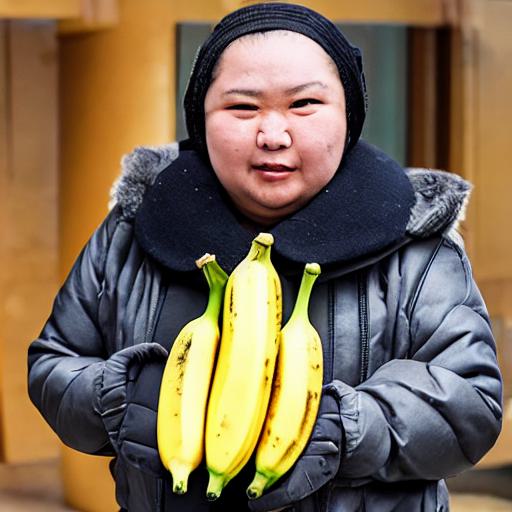}}
        & \raisebox{-0.5\height}{\includegraphics[width=.105\linewidth]{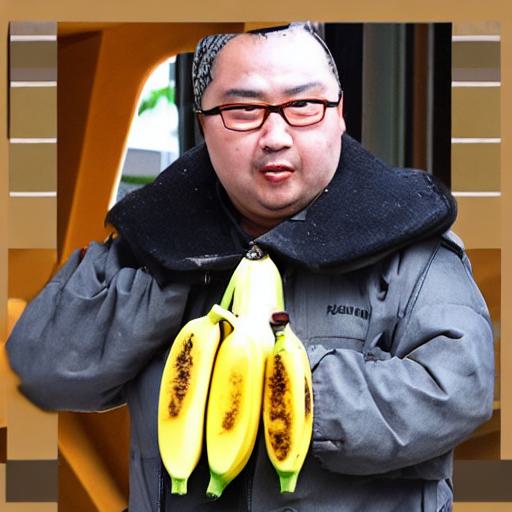}}
        & \raisebox{-0.5\height}{\includegraphics[width=.105\linewidth]{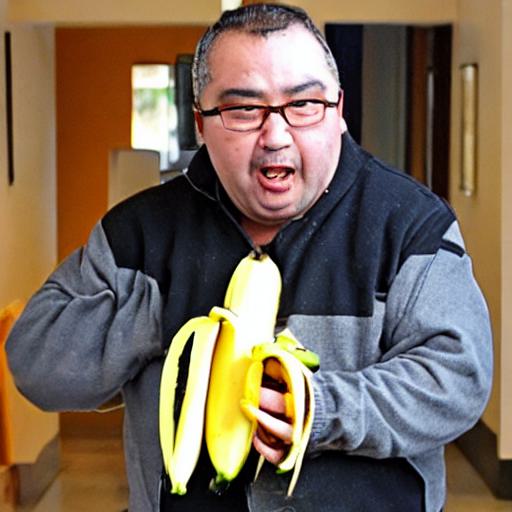}}
        & \raisebox{-0.5\height}{\includegraphics[width=.105\linewidth]{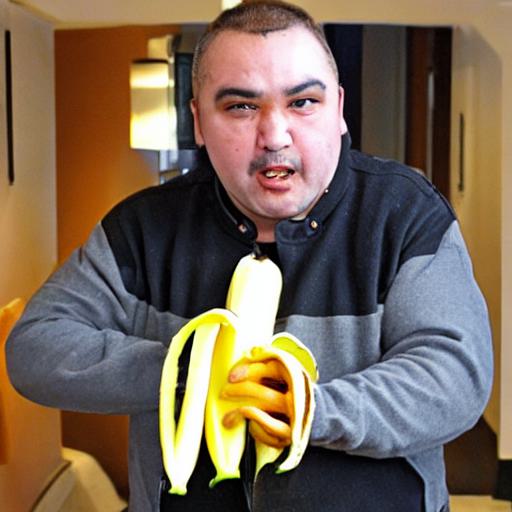}}
        & \raisebox{-0.5\height}{\includegraphics[width=.105\linewidth]{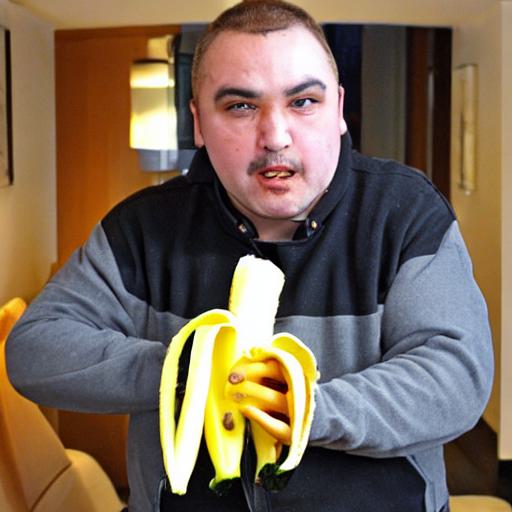}}
        & \raisebox{-0.5\height}{\includegraphics[width=.105\linewidth]{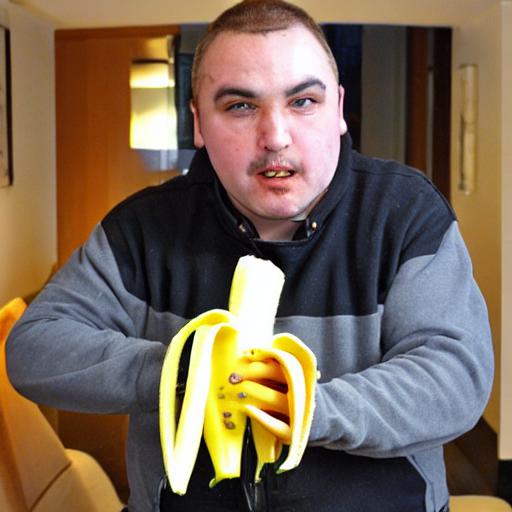}}
        & \raisebox{-0.5\height}{\includegraphics[width=.105\linewidth]{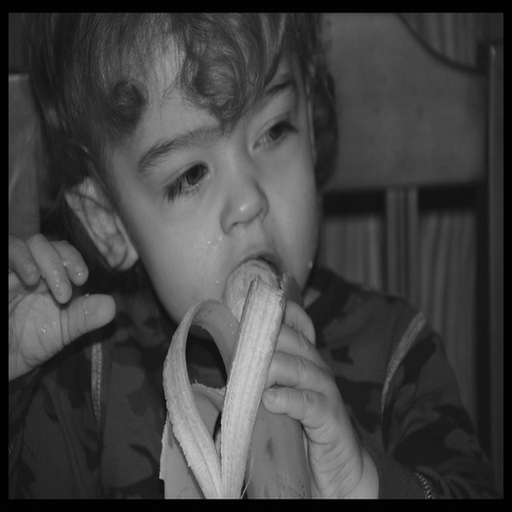}}\\
        \raisebox{-0.5\height}{\includegraphics[width=.105\linewidth]{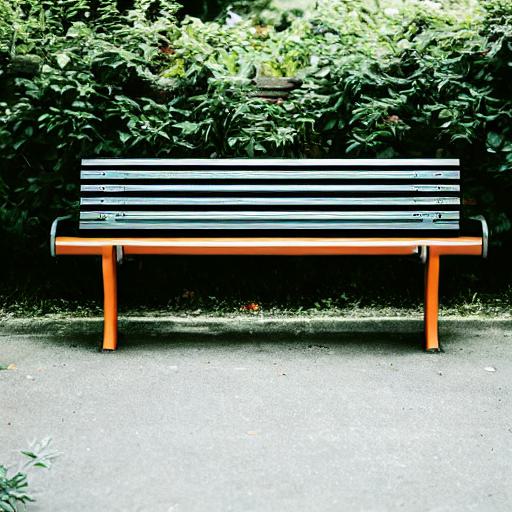}}
        & \raisebox{-0.5\height}{\includegraphics[width=.105\linewidth]{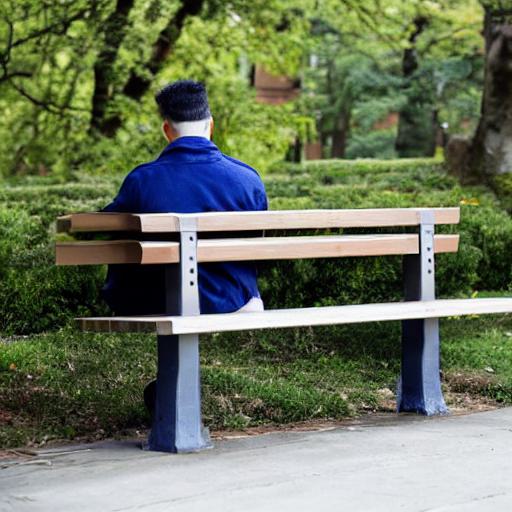}}
        & \raisebox{-0.5\height}{\includegraphics[width=.105\linewidth]{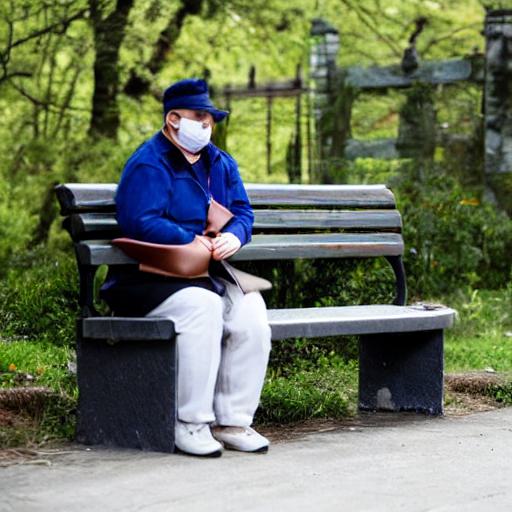}}
        & \raisebox{-0.5\height}{\includegraphics[width=.105\linewidth]{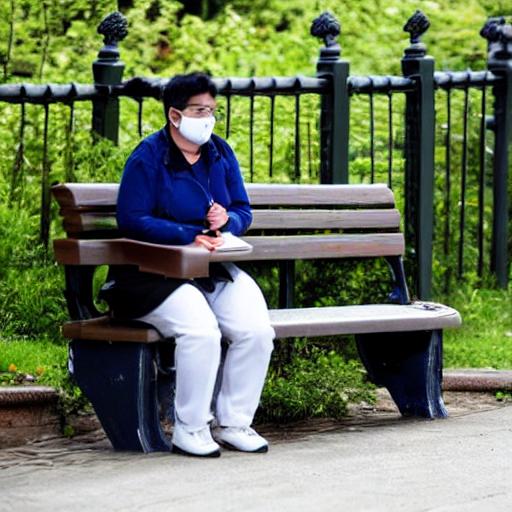}}
        & \raisebox{-0.5\height}{\includegraphics[width=.105\linewidth]{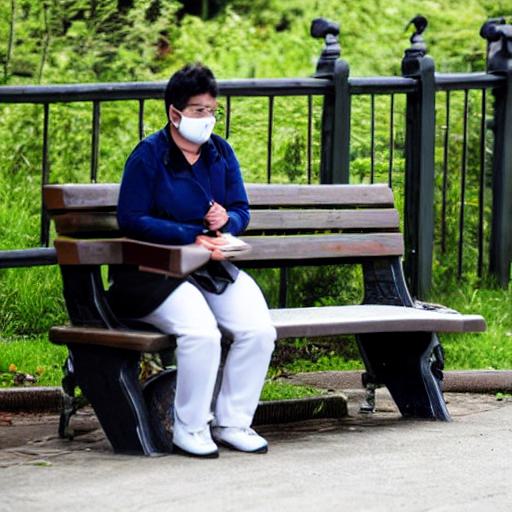}}
        & \raisebox{-0.5\height}{\includegraphics[width=.105\linewidth]{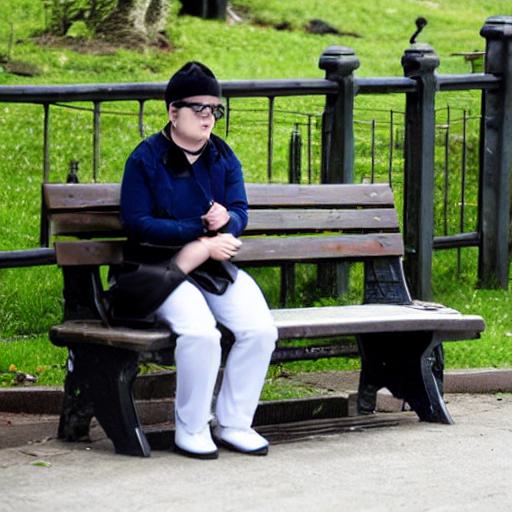}}
        & \raisebox{-0.5\height}{\includegraphics[width=.105\linewidth]{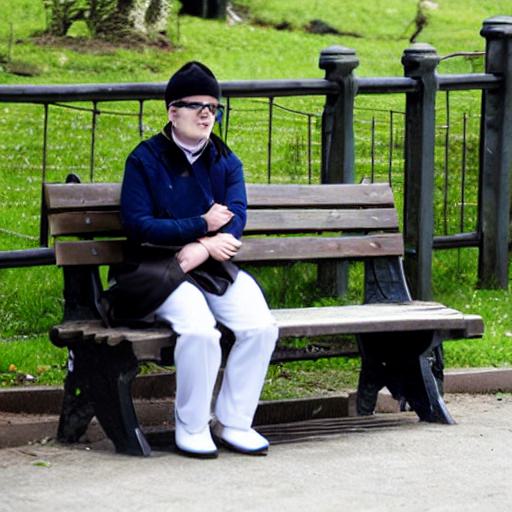}}
        & \raisebox{-0.5\height}{\includegraphics[width=.105\linewidth]{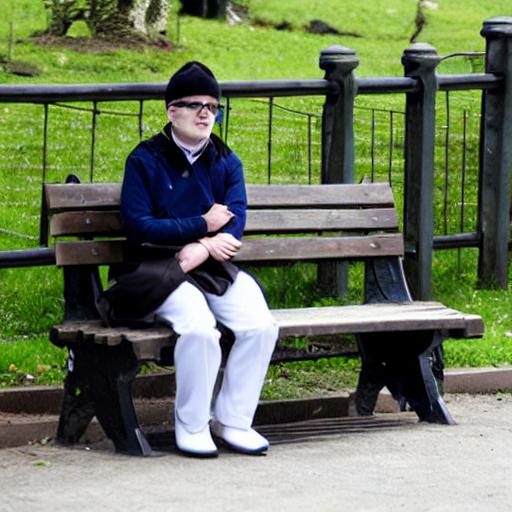}}
        & \raisebox{-0.5\height}{\includegraphics[width=.105\linewidth]{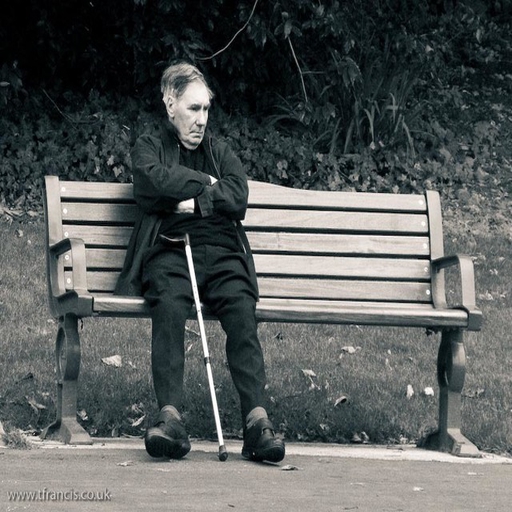}}\\
    \end{tabular}
    \captionsetup{type=figure}
    \caption{Ablation of scheduled sampling rate $\omega$. It adjust the degree of attentiveness to interaction condition. Zoom in for detail.}
    \label{fig:ablation_sampling_qualitative}
\end{table*}
\begin{figure}
    \centering
    \includegraphics[width=1\linewidth]{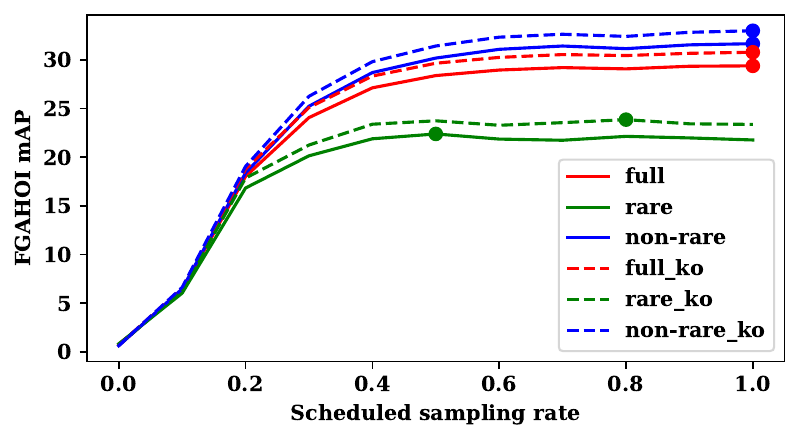}
    \caption{HOI detection score for various $\omega$ measured using FGAHOI with Swin-Tiny.}
    \label{fig:ablate_sampling_map}
\end{figure}
\begin{figure}
    \centering
    \includegraphics[width=1\linewidth]{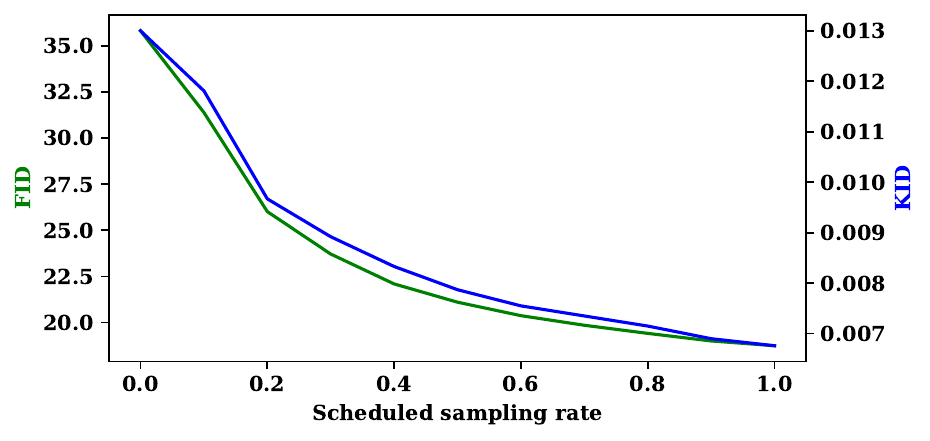}
    \caption{Quality scores for various $\omega$.}
    \label{fig:ablate_sampling_fid}
\end{figure}
\subsection{Scheduled Sampling} The scheduled sampling rate $\omega$ is a hyper-parameter in Interaction Transformer (\cref{eq:gated_attention}), which could greatly impact the generation as it control the degree of adherence to the interaction conditions. Thus, we ablate this hyper-parameter in interval of 0.1 from 0.0 to 1.0. \cref{fig:ablate_sampling_map} and \cref{fig:ablate_sampling_fid} show the mAP and FID score for different values of scheduled sampling rate $\omega$ while \cref{fig:ablation_sampling_qualitative} shows qualitative samples for different values of scheduled sampling rate $\omega$.

From \cref{fig:ablate_sampling_map,fig:ablate_sampling_fid}, we find that the interaction controllability improves as $\omega$ increases and converges around $\omega=0.6$ and $\omega=1.0$ produces best results in term of HOI detection score for every subset, while FID and KID decreases gradually as $\omega$ increases and $\omega=1.0$ produces least FID and KID distance when compared to original HICO-DET dataset. We recommend $\omega=0.8$ in most of the cases, as it stride a balance between text caption and interaction condition adherence.
In \cref{fig:ablation_sampling_qualitative}, the interaction correspondence increases gradually as $\omega$ increases, which is more obvious especially in range $\omega=0.1$ to $\omega=0.3$. When $\omega=0.0$ is used, the model reduces back to the Stable Diffusion model where the Interaction Transformer is ignored.

\subsection{Model Transferability}\label{sec:transferability}
\begin{table*}[!htbp]
    \centering
    \setlength{\tabcolsep}{1pt} 
    \renewcommand{\arraystretch}{1} 
    \begin{tabular}{cccccccc}
        \footnotesize Input & \footnotesize CuteYukiMix & \footnotesize{RCNZCartoon3D} & \footnotesize ToonYou & \footnotesize Lyriel & \footnotesize DarkSushiMix & \footnotesize RealisticVision & \footnotesize ChilloutMix \\
        \raisebox{-0.5\height}{\frame{\includegraphics[width=.118\linewidth]{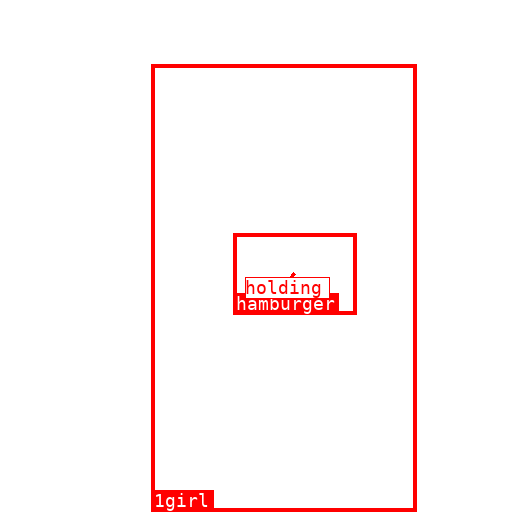}}}
        & \raisebox{-0.5\height}{\includegraphics[width=.118\linewidth]{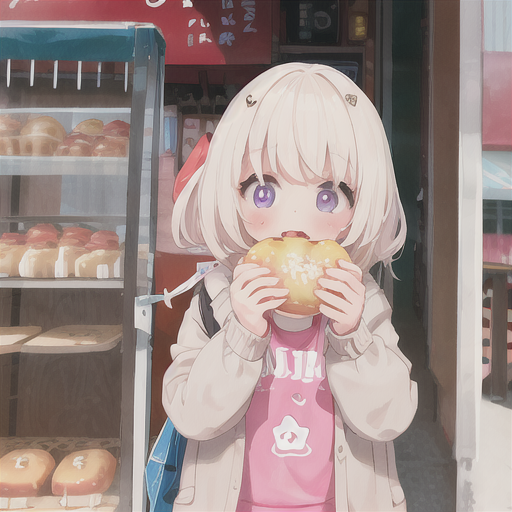}}
        & \raisebox{-0.5\height}{\includegraphics[width=.118\linewidth]{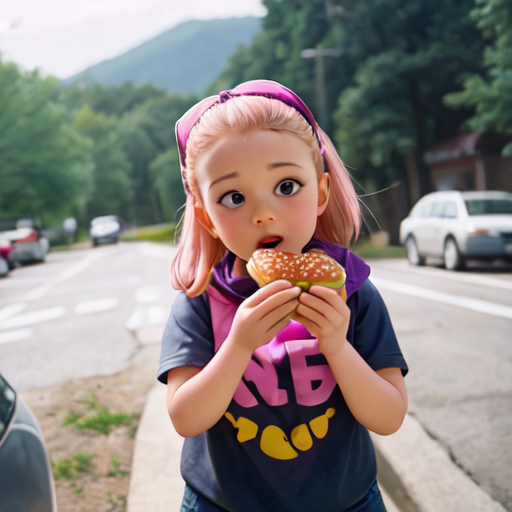}}
        & \raisebox{-0.5\height}{\includegraphics[width=.118\linewidth]{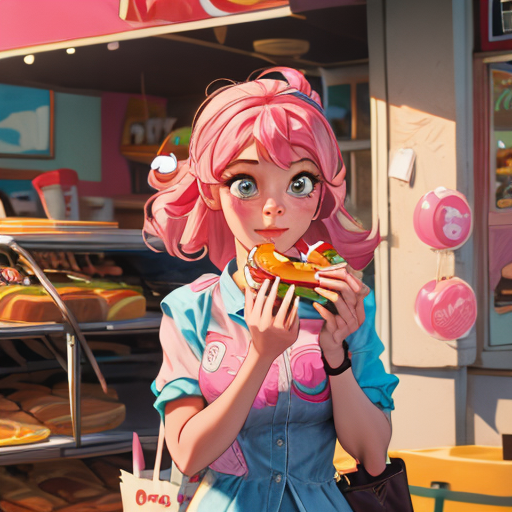}}
        & \raisebox{-0.5\height}{\includegraphics[width=.118\linewidth]{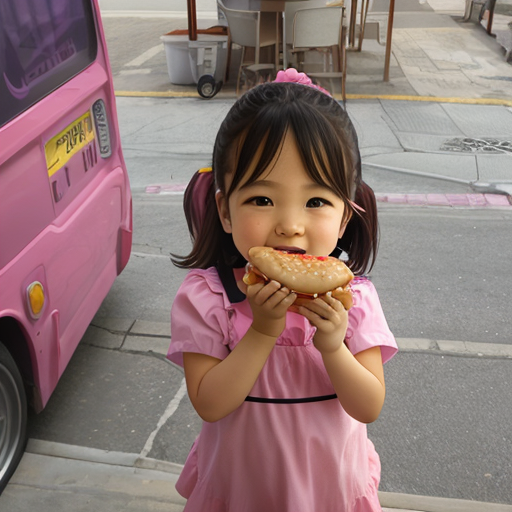}}
        & \raisebox{-0.5\height}{\includegraphics[width=.118\linewidth]{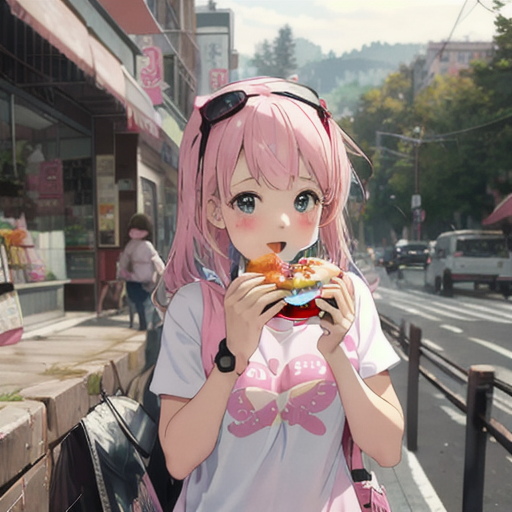}}
        & \raisebox{-0.5\height}{\includegraphics[width=.118\linewidth]{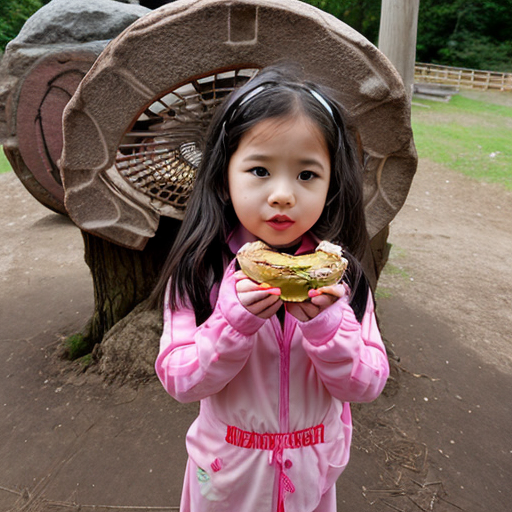}}
        & \raisebox{-0.5\height}{\includegraphics[width=.118\linewidth]{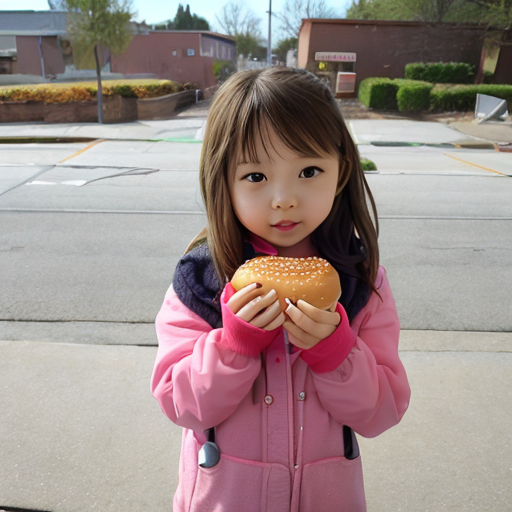}}\\ 
        & \multicolumn{7}{l}{\footnotesize best quality, pink, 1girl holding a hamburger}\\
        
        \raisebox{-0.5\height}{\frame{\includegraphics[width=.118\linewidth]{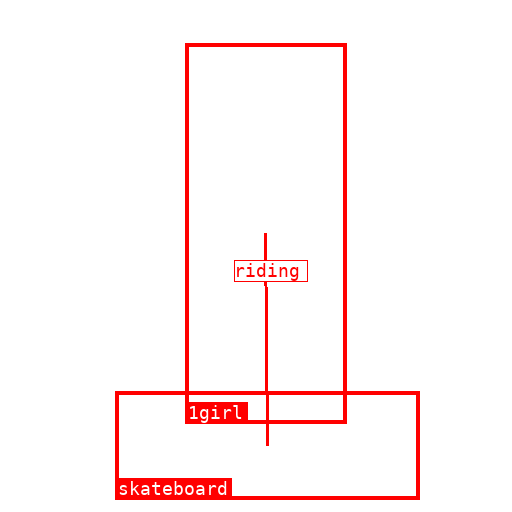}}}
        & \raisebox{-0.5\height}{\includegraphics[width=.118\linewidth]{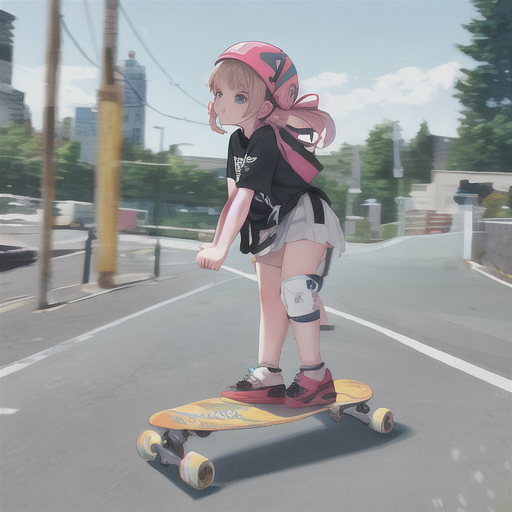}}
        & \raisebox{-0.5\height}{\includegraphics[width=.118\linewidth]{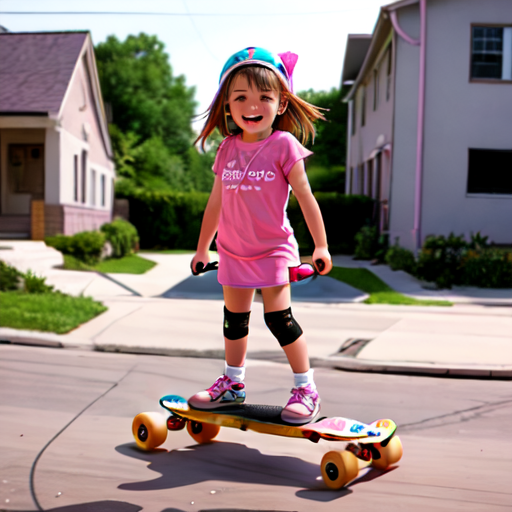}}
        & \raisebox{-0.5\height}{\includegraphics[width=.118\linewidth]{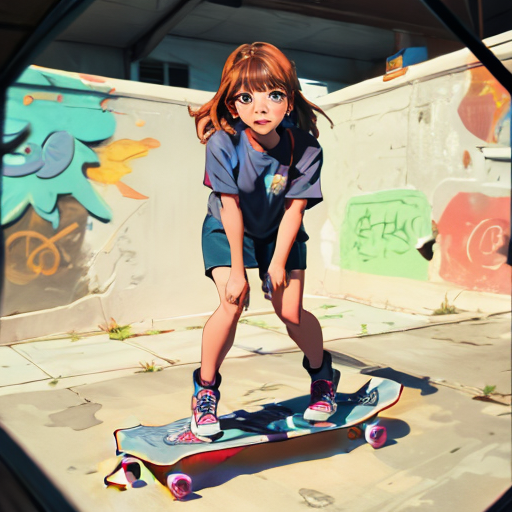}}
        & \raisebox{-0.5\height}{\includegraphics[width=.118\linewidth]{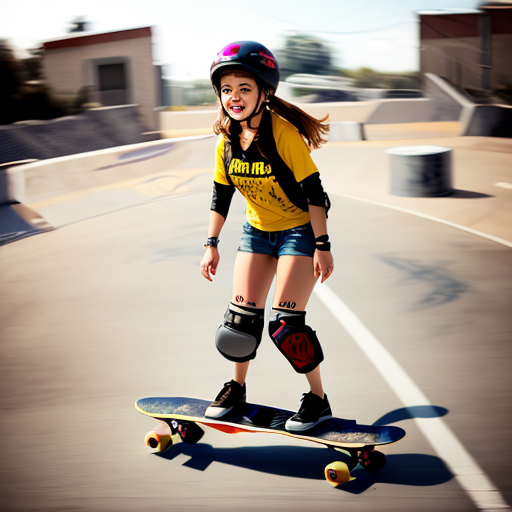}}
        & \raisebox{-0.5\height}{\includegraphics[width=.118\linewidth]{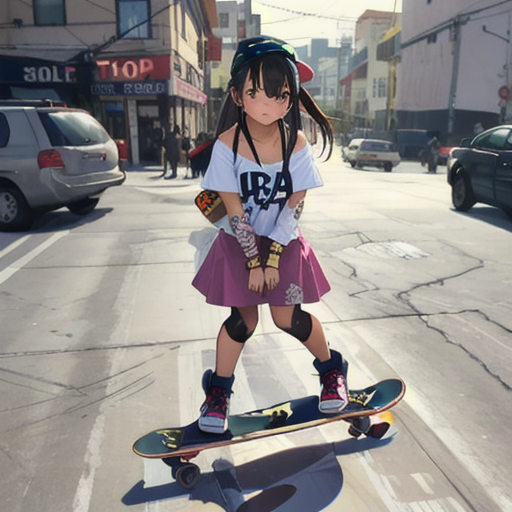}}
        & \raisebox{-0.5\height}{\includegraphics[width=.118\linewidth]{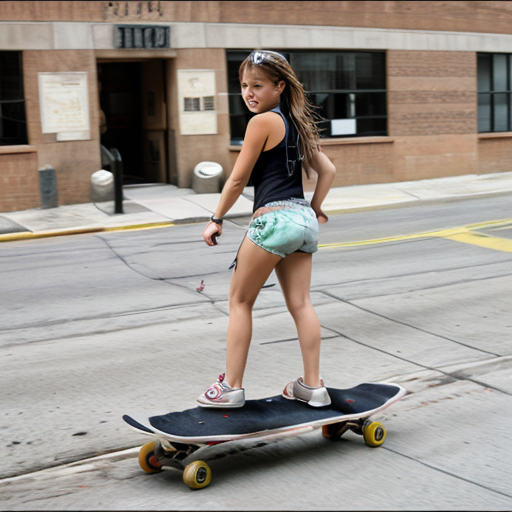}}
        & \raisebox{-0.5\height}{\includegraphics[width=.118\linewidth]{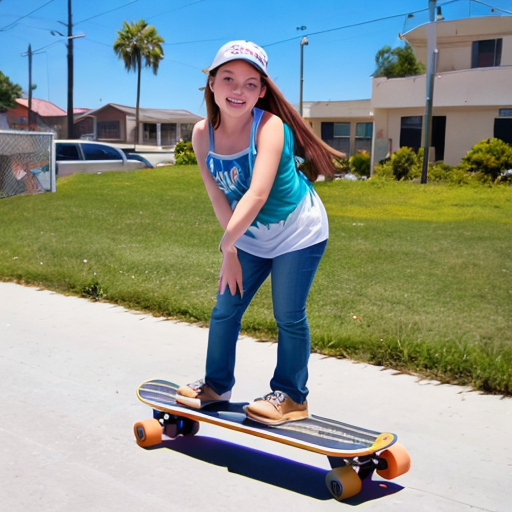}}\\ 
        
        \raisebox{-0.5\height}{\includegraphics[width=.118\linewidth]{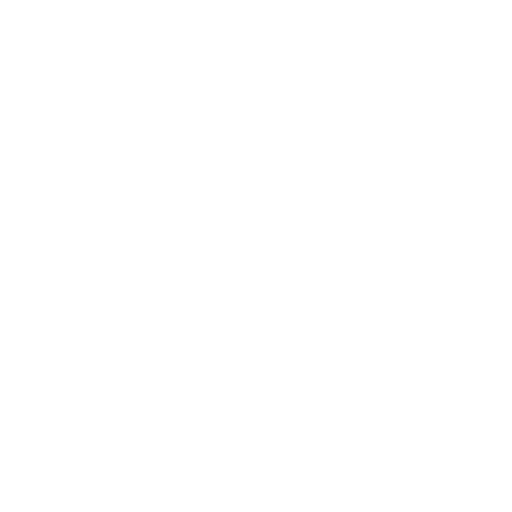}}
        & \raisebox{-0.5\height}{\includegraphics[width=.118\linewidth]{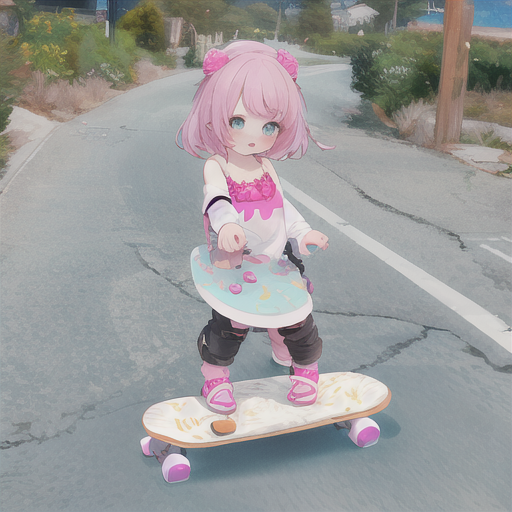}}
        & \raisebox{-0.5\height}{\includegraphics[width=.118\linewidth]{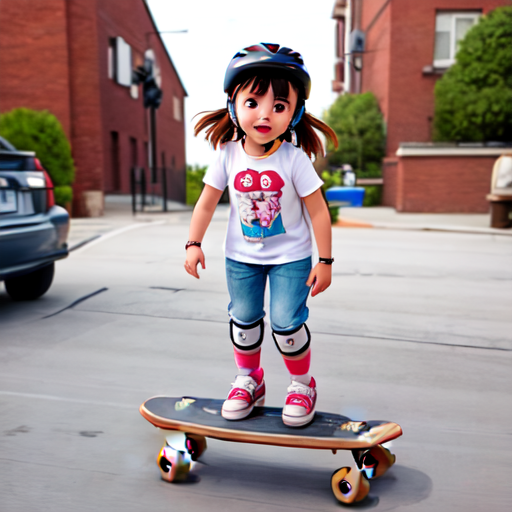}}
        & \raisebox{-0.5\height}{\includegraphics[width=.118\linewidth]{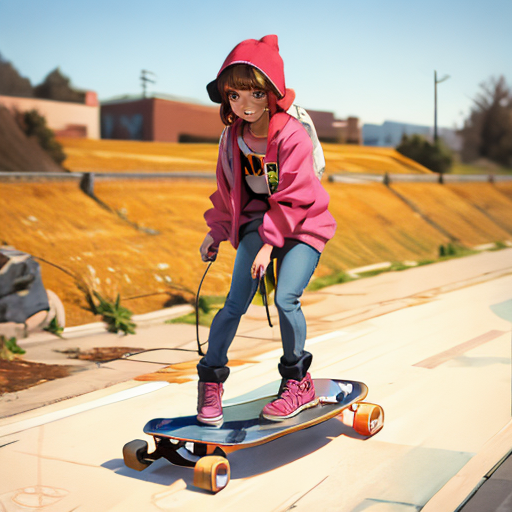}}
        & \raisebox{-0.5\height}{\includegraphics[width=.118\linewidth]{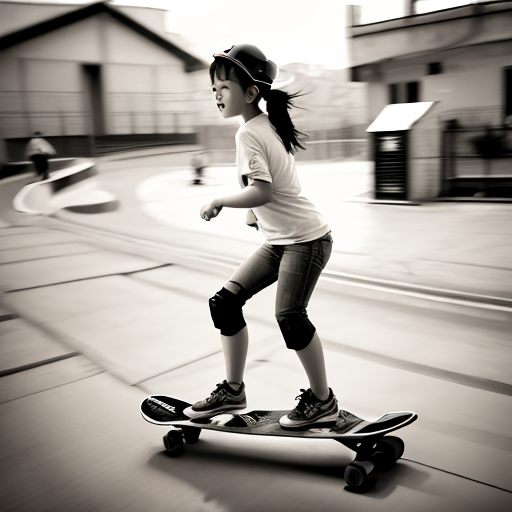}}
        & \raisebox{-0.5\height}{\includegraphics[width=.118\linewidth]{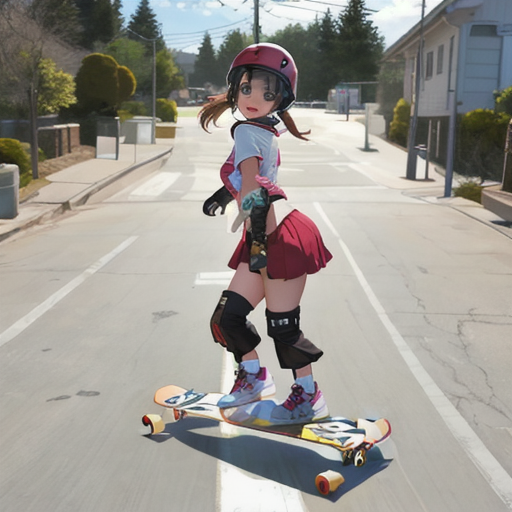}}
        & \raisebox{-0.5\height}{\includegraphics[width=.118\linewidth]{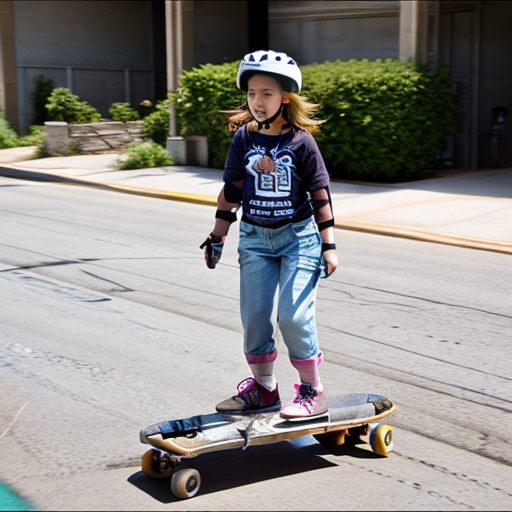}}
        & \raisebox{-0.5\height}{\includegraphics[width=.118\linewidth]{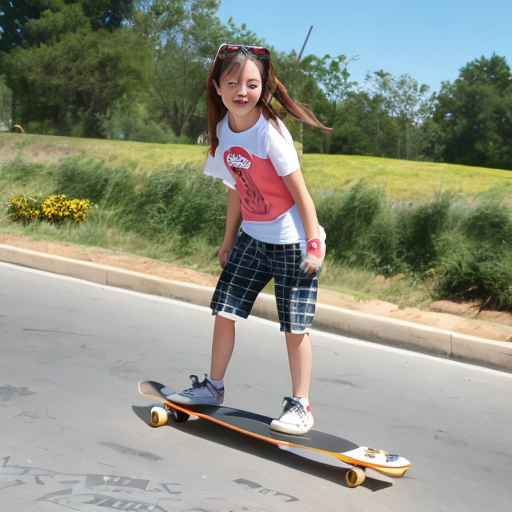}}\\ 
        & \multicolumn{7}{l}{\footnotesize best quality, 1girl is riding a skateboard}\\

        \raisebox{-0.5\height}{\frame{\includegraphics[width=.118\linewidth]{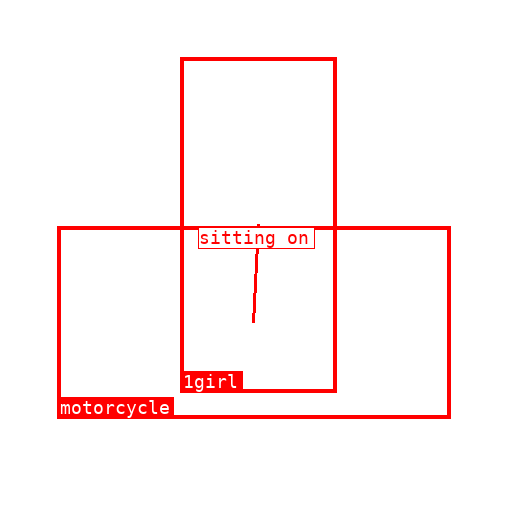}}}
        & \raisebox{-0.5\height}{\includegraphics[width=.118\linewidth]{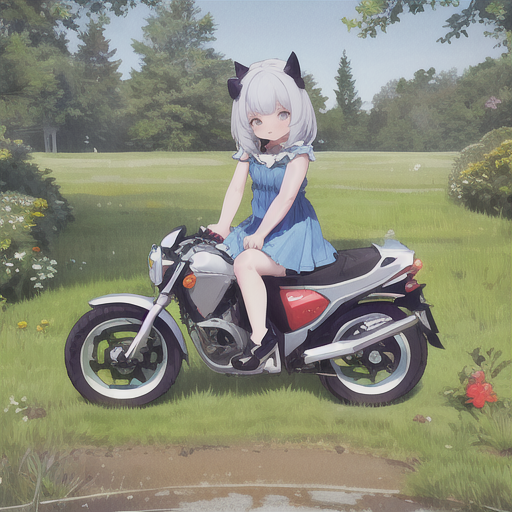}}
        & \raisebox{-0.5\height}{\includegraphics[width=.118\linewidth]{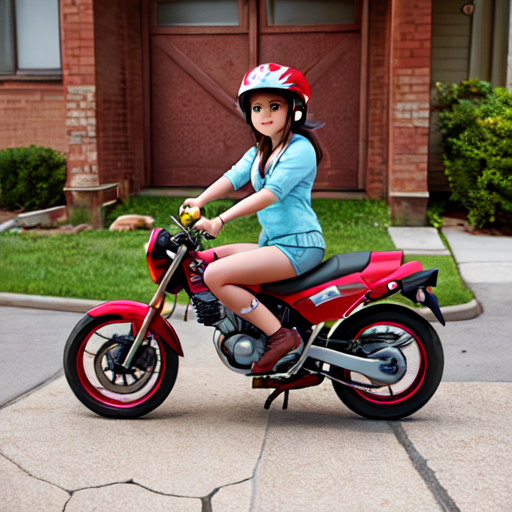}}
        & \raisebox{-0.5\height}{\includegraphics[width=.118\linewidth]{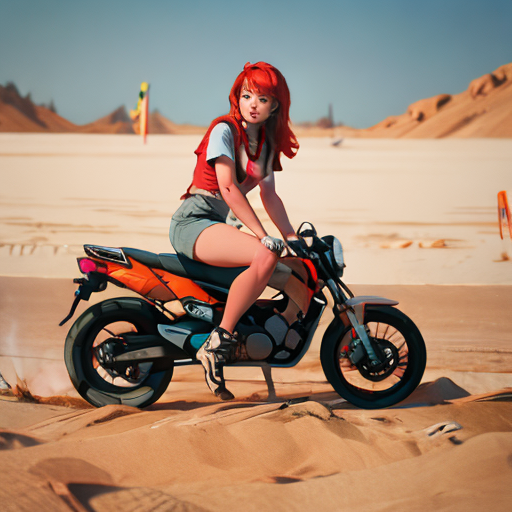}}
        & \raisebox{-0.5\height}{\includegraphics[width=.118\linewidth]{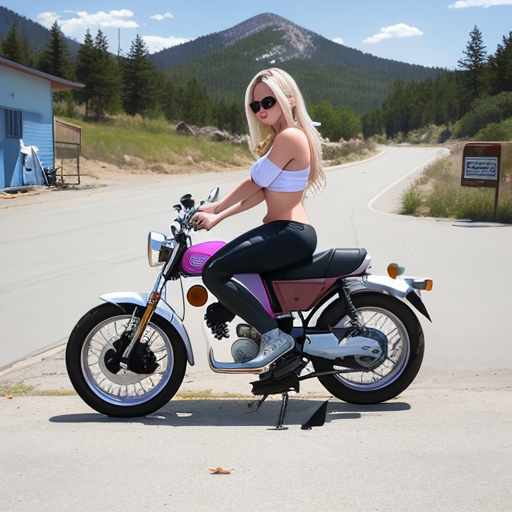}}
        & \raisebox{-0.5\height}{\includegraphics[width=.118\linewidth]{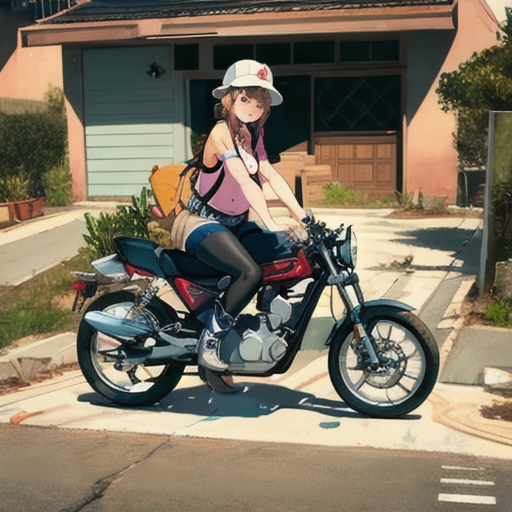}}
        & \raisebox{-0.5\height}{\includegraphics[width=.118\linewidth]{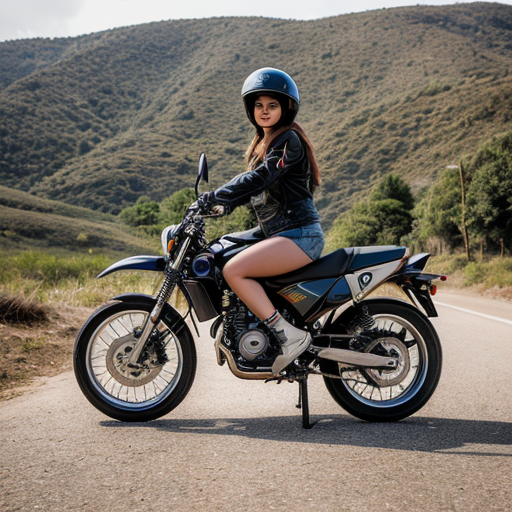}}
        & \raisebox{-0.5\height}{\includegraphics[width=.118\linewidth]{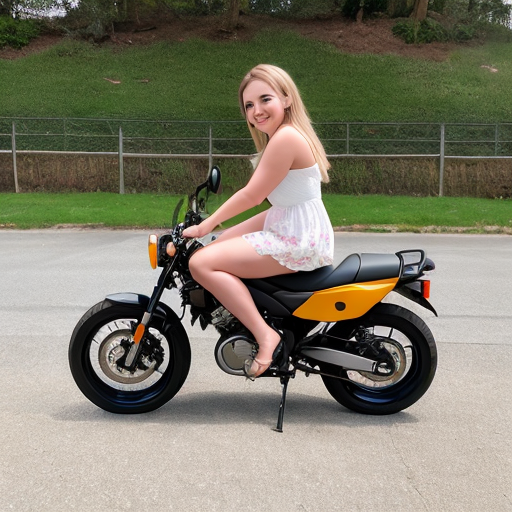}}\\ 
        & \multicolumn{7}{l}{\footnotesize best quality, 1girl is sitting on a motorcycle}\\

        \raisebox{-0.5\height}{\frame{\includegraphics[width=.118\linewidth]{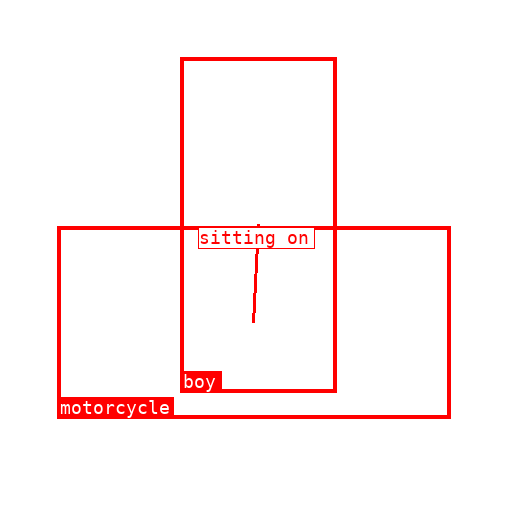}}}
        & \raisebox{-0.5\height}{\includegraphics[width=.118\linewidth]{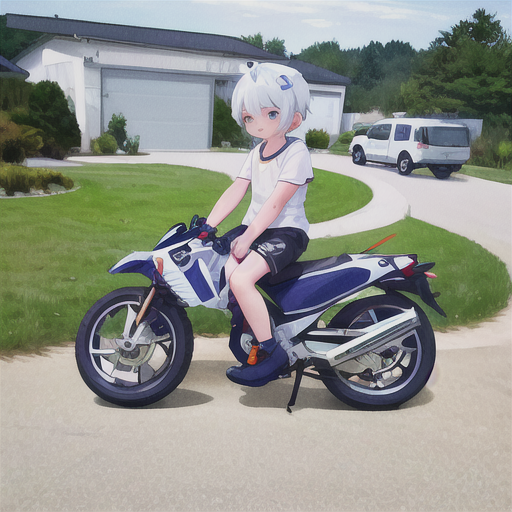}}
        & \raisebox{-0.5\height}{\includegraphics[width=.118\linewidth]{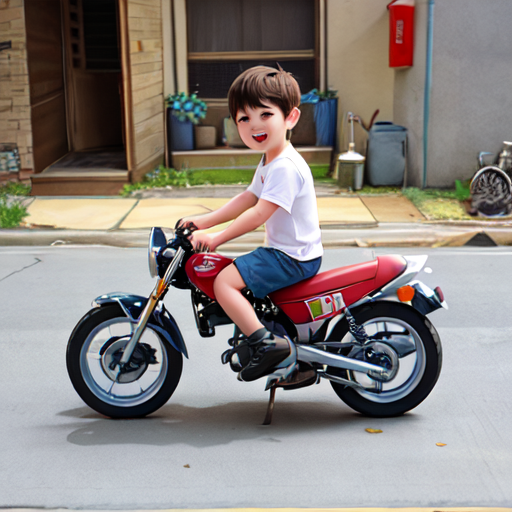}}
        & \raisebox{-0.5\height}{\includegraphics[width=.118\linewidth]{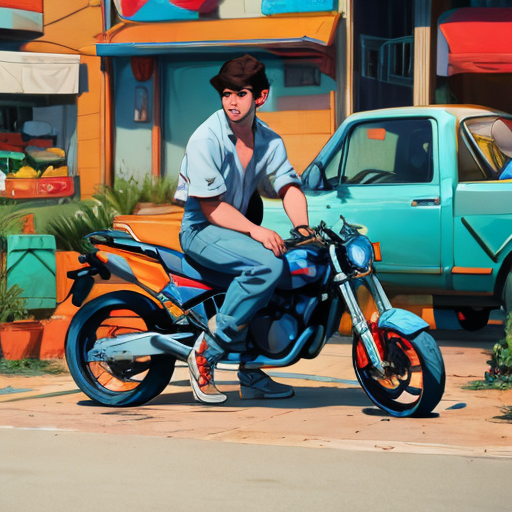}}
        & \raisebox{-0.5\height}{\includegraphics[width=.118\linewidth]{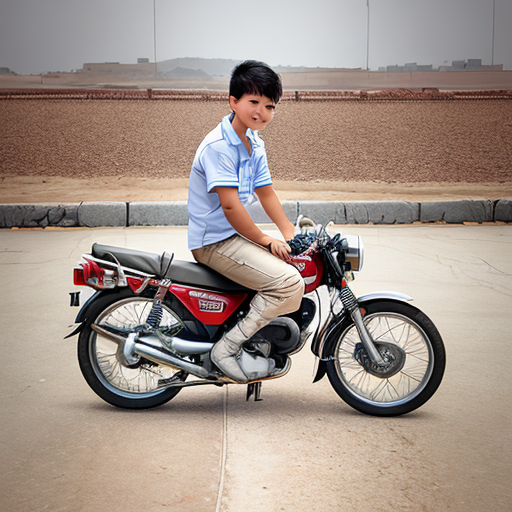}}
        & \raisebox{-0.5\height}{\includegraphics[width=.118\linewidth]{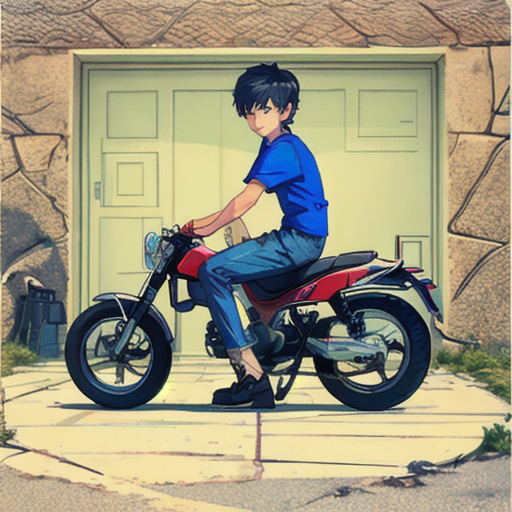}}
        & \raisebox{-0.5\height}{\includegraphics[width=.118\linewidth]{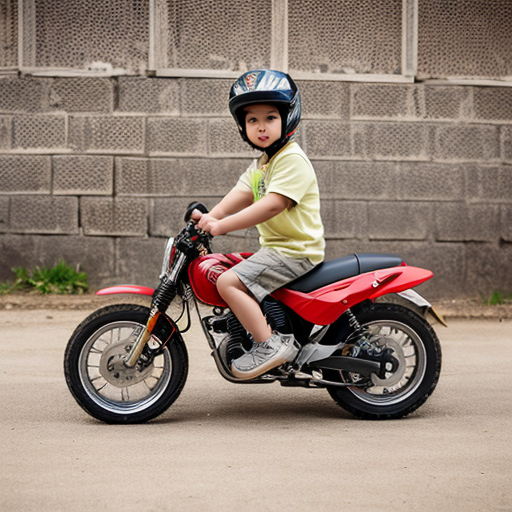}}
        & \raisebox{-0.5\height}{\includegraphics[width=.118\linewidth]{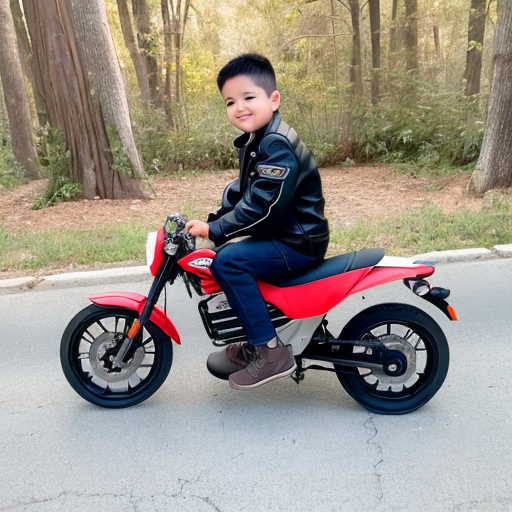}}\\ 
        & \multicolumn{7}{l}{\footnotesize best quality, a boy is sitting on a motorcycle}\\

        \raisebox{-0.5\height}{\frame{\includegraphics[width=.118\linewidth]{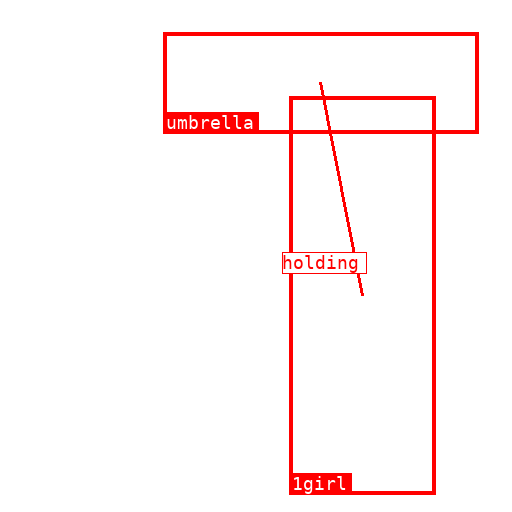}}}
        & \raisebox{-0.5\height}{\includegraphics[width=.118\linewidth]{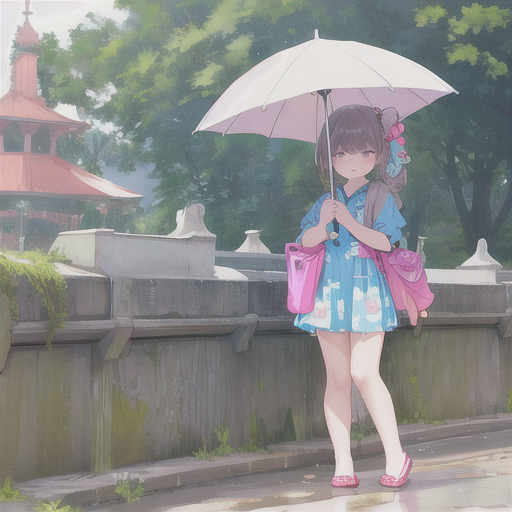}}
        & \raisebox{-0.5\height}{\includegraphics[width=.118\linewidth]{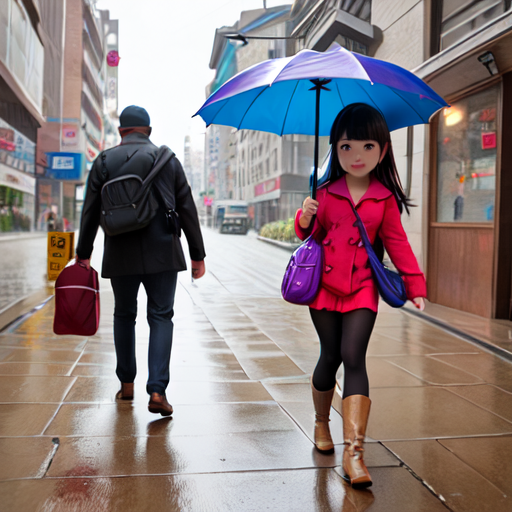}}
        & \raisebox{-0.5\height}{\includegraphics[width=.118\linewidth]{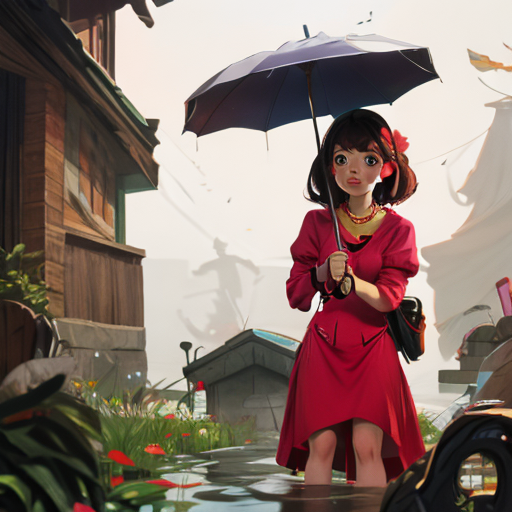}}
        & \raisebox{-0.5\height}{\includegraphics[width=.118\linewidth]{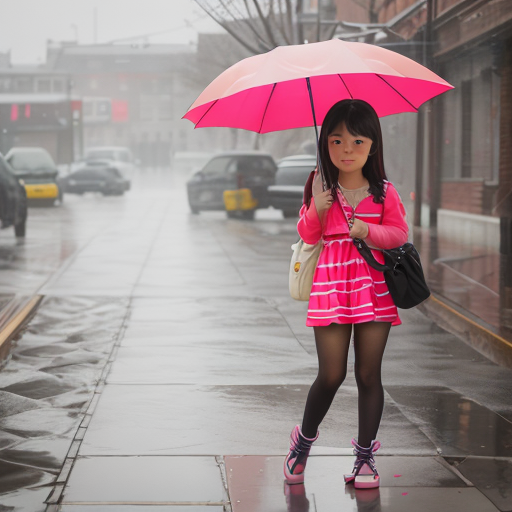}}
        & \raisebox{-0.5\height}{\includegraphics[width=.118\linewidth]{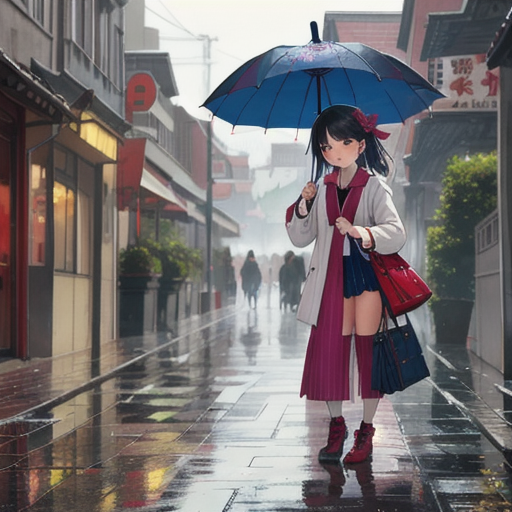}}
        & \raisebox{-0.5\height}{\includegraphics[width=.118\linewidth]{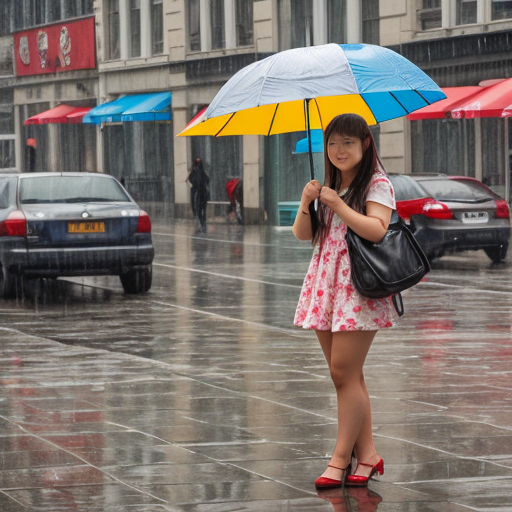}}
        & \raisebox{-0.5\height}{\includegraphics[width=.118\linewidth]{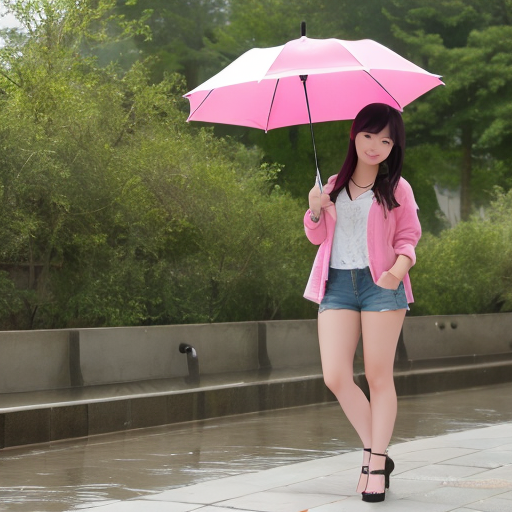}}\\ 
        & \multicolumn{7}{l}{\footnotesize best quality, 1girl is holding an umbrella}\\

        \raisebox{-0.5\height}{\frame{\includegraphics[width=.118\linewidth]{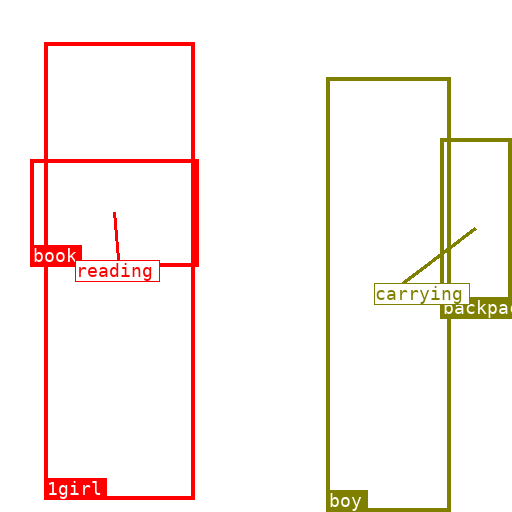}}}
        & \raisebox{-0.5\height}{\includegraphics[width=.118\linewidth]{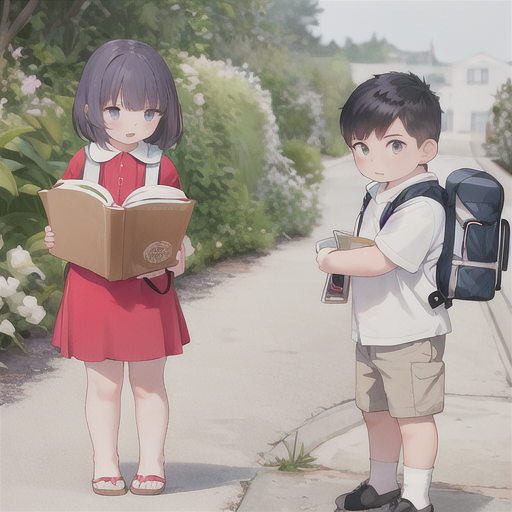}}
        & \raisebox{-0.5\height}{\includegraphics[width=.118\linewidth]{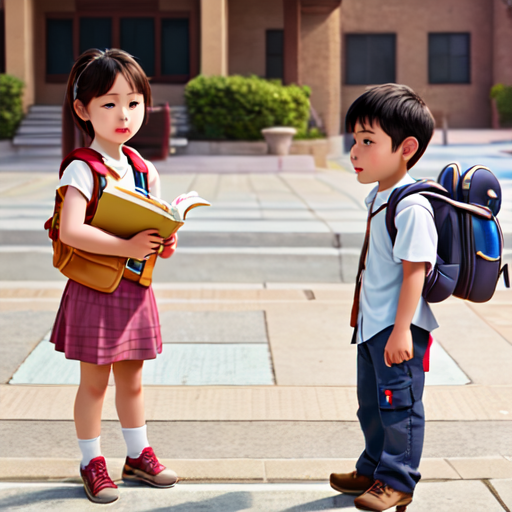}}
        & \raisebox{-0.5\height}{\includegraphics[width=.118\linewidth]{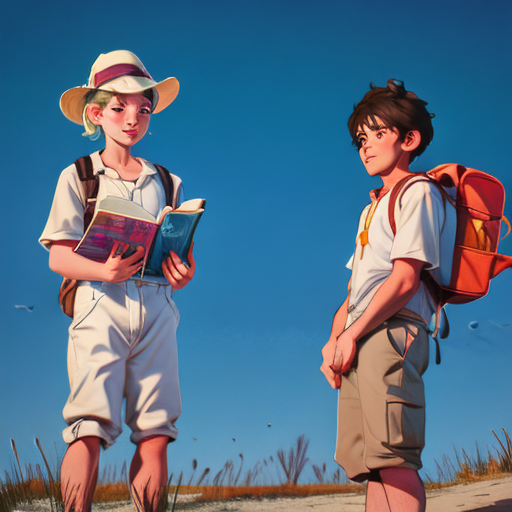}}
        & \raisebox{-0.5\height}{\includegraphics[width=.118\linewidth]{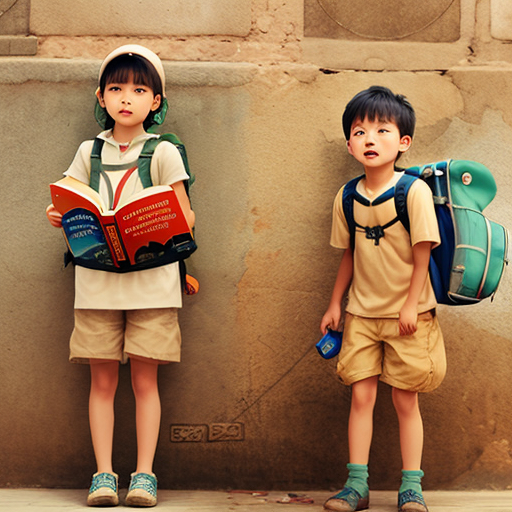}}
        & \raisebox{-0.5\height}{\includegraphics[width=.118\linewidth]{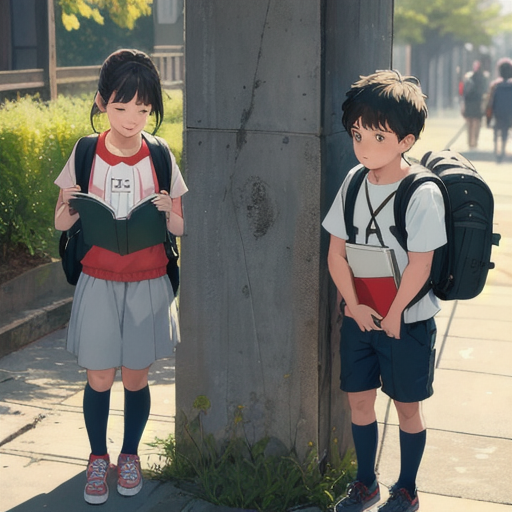}}
        & \raisebox{-0.5\height}{\includegraphics[width=.118\linewidth]{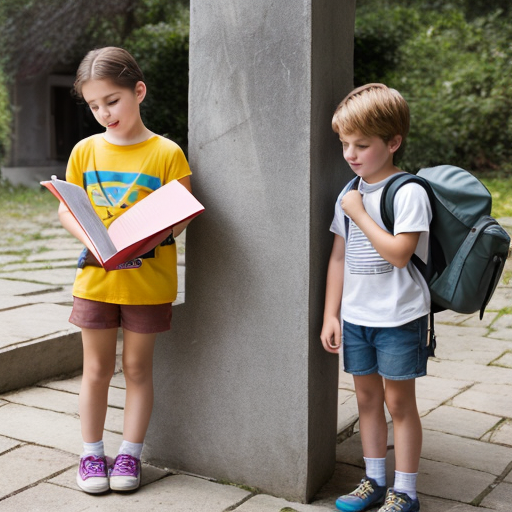}}
        & \raisebox{-0.5\height}{\includegraphics[width=.118\linewidth]{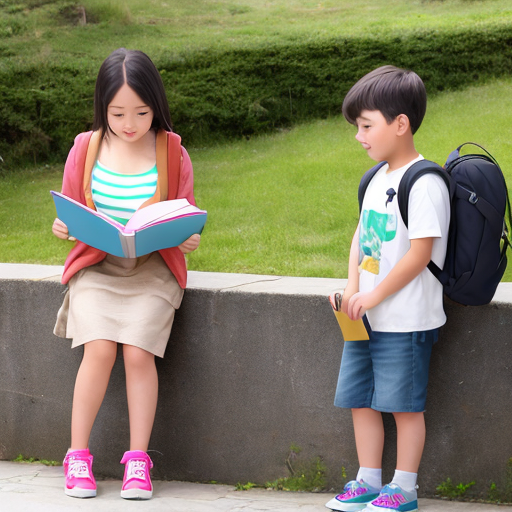}}\\ 
        & \multicolumn{7}{l}{\footnotesize best quality, 1girl is reading a book, a boy is carrying a backpack}\\

        \raisebox{-0.5\height}{\frame{\includegraphics[width=.118\linewidth]{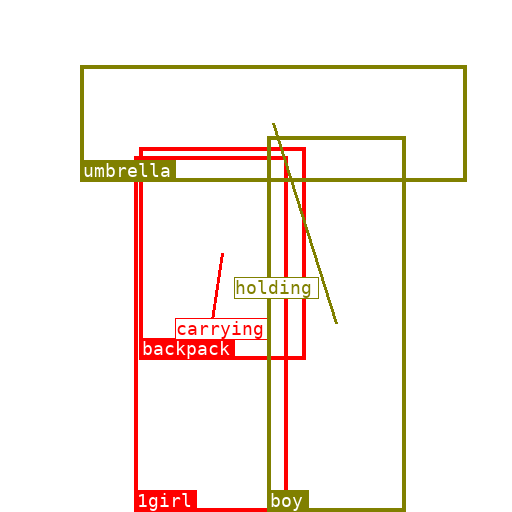}}}
        & \raisebox{-0.5\height}{\includegraphics[width=.118\linewidth]{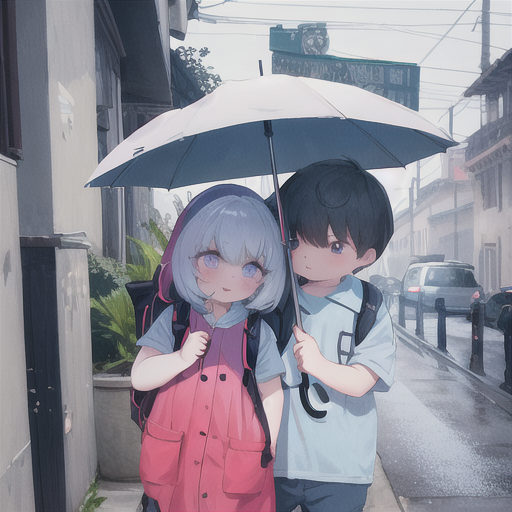}}
        & \raisebox{-0.5\height}{\includegraphics[width=.118\linewidth]{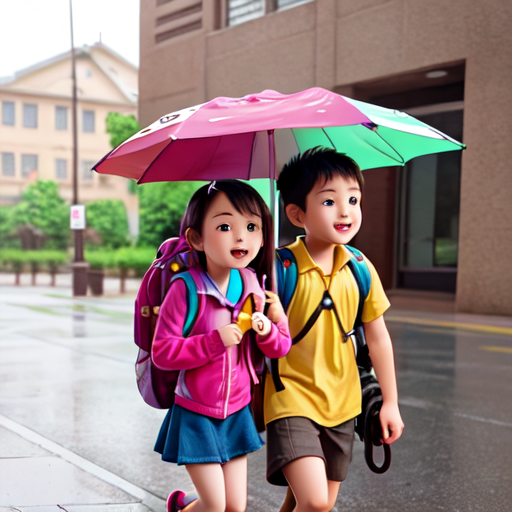}}
        & \raisebox{-0.5\height}{\includegraphics[width=.118\linewidth]{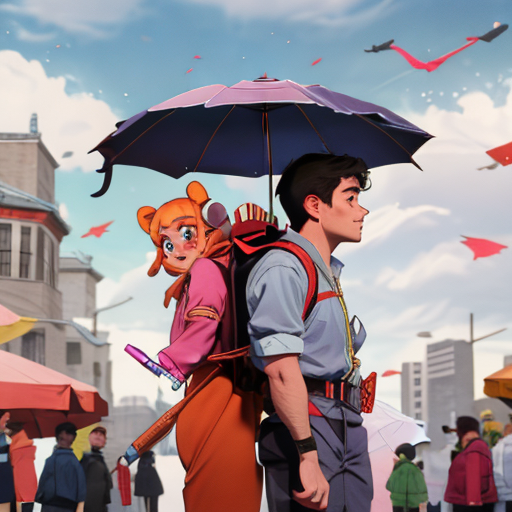}}
        & \raisebox{-0.5\height}{\includegraphics[width=.118\linewidth]{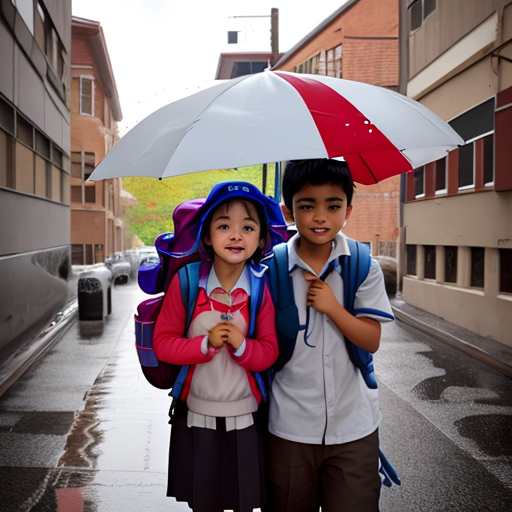}}
        & \raisebox{-0.5\height}{\includegraphics[width=.118\linewidth]{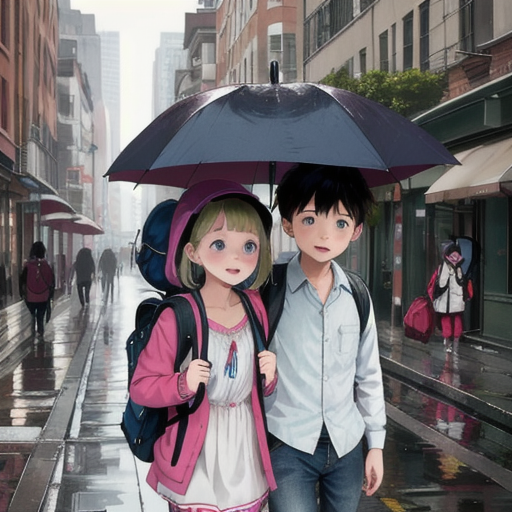}}
        & \raisebox{-0.5\height}{\includegraphics[width=.118\linewidth]{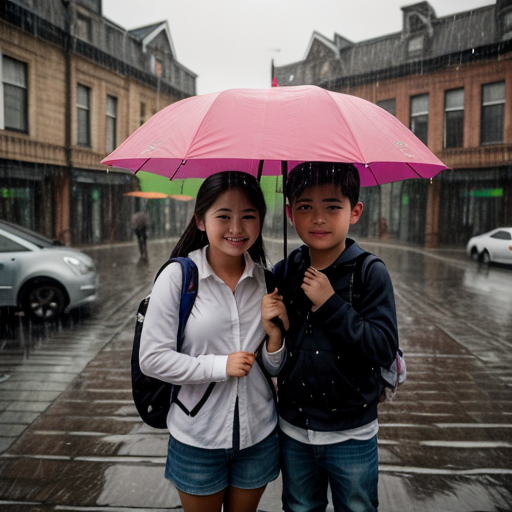}}
        & \raisebox{-0.5\height}{\includegraphics[width=.118\linewidth]{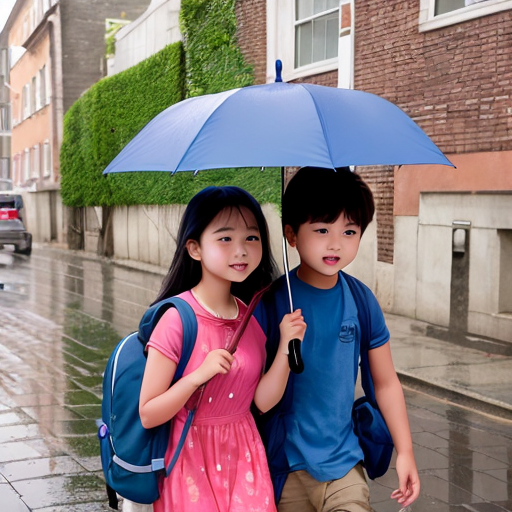}}\\ 
        & \multicolumn{7}{l}{\footnotesize best quality, 1girl is carrying a backpack, a boy is holding an umbrella}\\

    \end{tabular}
    \captionsetup{type=figure}
    \caption{Visualization of InteractDiffusion on various personalized StableDiffusion models. Zoom in for detail.}
    \label{fig:personalized}
\end{table*}
\begin{table*}[t]
\centering
\begin{tabular}{lcccccc}
\hline
\multicolumn{1}{l|}{}                               & \multicolumn{3}{c|}{Tiny}                                                                                                                                                                                       & \multicolumn{3}{c}{Large}                                                                                                                                                                                       \\ \cline{2-7} 
\multicolumn{1}{l|}{\multirow{-2}{*}{Method}}       & Full                                                                & Unseen                                                              & \multicolumn{1}{c|}{Seen}                                           & Full                                                                & Unseen                                                              & Seen                                                                \\ \hline
\rowcolor[HTML]{EFEFEF} 
{\color[HTML]{333333} \small{\textbf{Zero-shot}}}   & \multicolumn{1}{l}{\cellcolor[HTML]{EFEFEF}{\color[HTML]{333333} }} & \multicolumn{1}{l}{\cellcolor[HTML]{EFEFEF}{\color[HTML]{333333} }} & \multicolumn{1}{l}{\cellcolor[HTML]{EFEFEF}{\color[HTML]{333333} }} & \multicolumn{1}{l}{\cellcolor[HTML]{EFEFEF}{\color[HTML]{333333} }} & \multicolumn{1}{l}{\cellcolor[HTML]{EFEFEF}{\color[HTML]{333333} }} & \multicolumn{1}{l}{\cellcolor[HTML]{EFEFEF}{\color[HTML]{333333} }} \\
\multicolumn{1}{l|}{InteractDiffusion (ZS)}  & 28.47(-0.65)                                                        & 20.75(-3.10)                                                        & \multicolumn{1}{c|}{30.41(-0.03)}                                   & 30.31(-0.73)                                                        & 23.06(-2.30)                                                        & 32.12(-0.34)                                                        \\ \hline
\rowcolor[HTML]{EFEFEF} 
{\color[HTML]{333333} \small{\textbf{Fully Seen}}}   & \multicolumn{1}{l}{\cellcolor[HTML]{EFEFEF}{\color[HTML]{333333} }} & \multicolumn{1}{l}{\cellcolor[HTML]{EFEFEF}{\color[HTML]{333333} }} & \multicolumn{1}{l}{\cellcolor[HTML]{EFEFEF}{\color[HTML]{333333} }} & \multicolumn{1}{l}{\cellcolor[HTML]{EFEFEF}{\color[HTML]{333333} }} & \multicolumn{1}{l}{\cellcolor[HTML]{EFEFEF}{\color[HTML]{333333} }} & \multicolumn{1}{l}{\cellcolor[HTML]{EFEFEF}{\color[HTML]{333333} }} \\
\multicolumn{1}{l|}{GLIGEN*} & 25.23                                                               & 17.77                                                               & \multicolumn{1}{c|}{27.10}                                          & 26.45                                                               & 19.23                                                               & 28.25                                                               \\
\multicolumn{1}{l|}{InteractDiffusion} & 29.12                                                               & 23.85                                                               & \multicolumn{1}{c|}{30.44}                                          & 31.04                                                               & 25.36                                                               & 32.46                                                               \\
\rowcolor[HTML]{EFEFEF} 
{\color[HTML]{333333} \small{\textbf{Reference}}}   & \multicolumn{1}{l}{\cellcolor[HTML]{EFEFEF}{\color[HTML]{333333} }} & \multicolumn{1}{l}{\cellcolor[HTML]{EFEFEF}{\color[HTML]{333333} }} & \multicolumn{1}{l}{\cellcolor[HTML]{EFEFEF}{\color[HTML]{333333} }} & \multicolumn{1}{l}{\cellcolor[HTML]{EFEFEF}{\color[HTML]{333333} }} & \multicolumn{1}{l}{\cellcolor[HTML]{EFEFEF}{\color[HTML]{333333} }} & \multicolumn{1}{l}{\cellcolor[HTML]{EFEFEF}{\color[HTML]{333333} }} \\
\multicolumn{1}{l|}{HICO-DET}                       & 29.81                                                               & 22.69                                                               & \multicolumn{1}{c|}{32.59}                                          & 37.11                                                               & 32.59                                                               & 38.24                                                               \\ \hline
\end{tabular}
\caption{Zero-shot performance of InteractDiffusion compared to default fully-seen setting. Comparison were made in relatively to Fully-Seen setting.}
\label{tab:zeroshot}
\end{table*}

\begin{table*}[t]
    \centering
    \setlength{\tabcolsep}{0.5pt} 
    \renewcommand{\arraystretch}{1} 
    \begin{tabular}{ccccccc}
        \raisebox{-0.15\height}{\frame{\includegraphics[width=.138\linewidth]{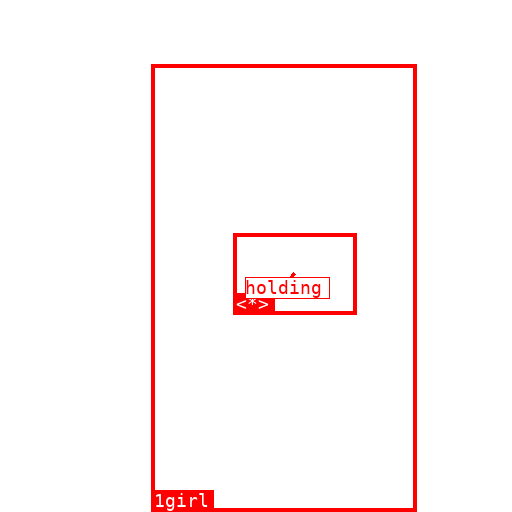}}}
        & \raisebox{-0.15\height}{\includegraphics[width=.138\linewidth]{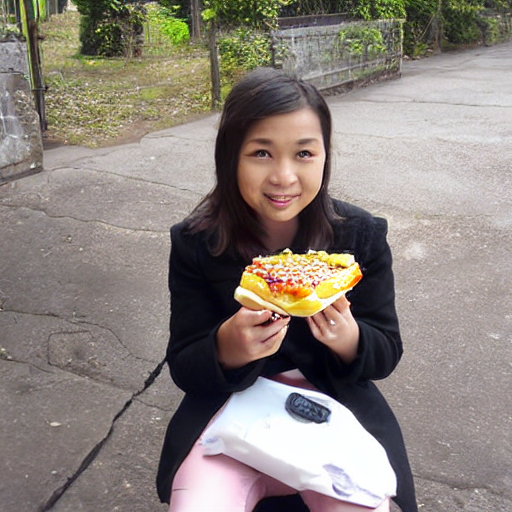}}
        & \raisebox{-0.15\height}{\includegraphics[width=.138\linewidth]{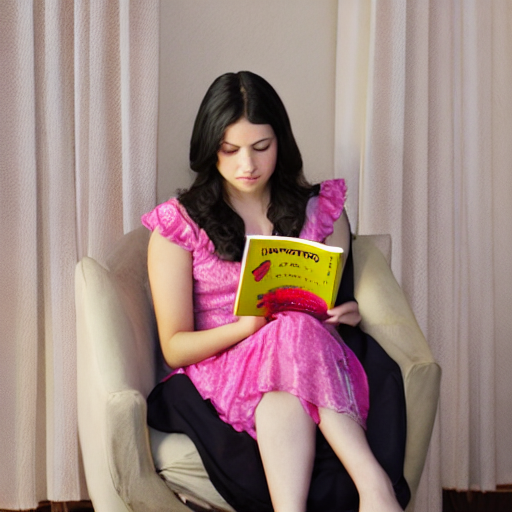}}
        & \raisebox{-0.15\height}{\includegraphics[width=.138\linewidth]{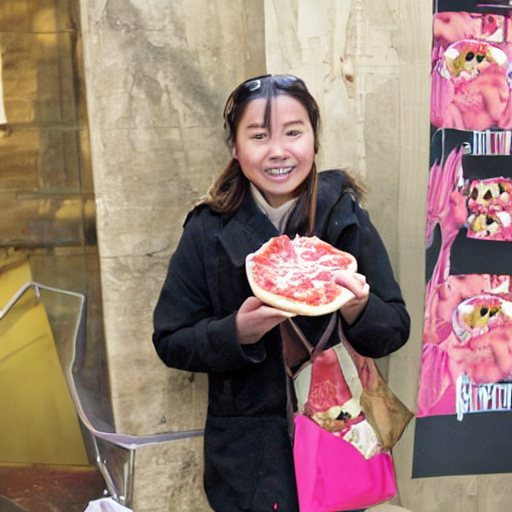}}
        & \raisebox{-0.15\height}{\includegraphics[width=.138\linewidth]{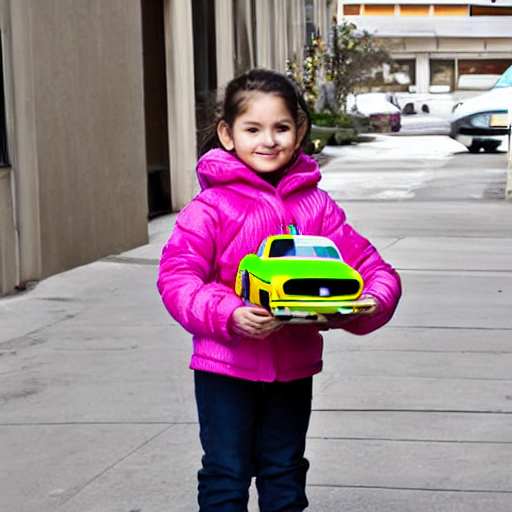}}
        & \raisebox{-0.15\height}{\includegraphics[width=.138\linewidth]{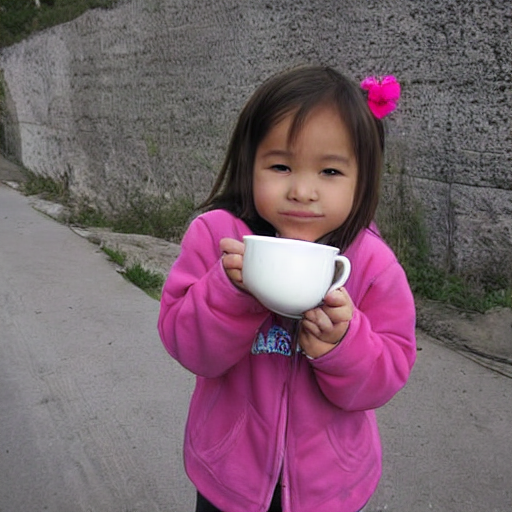}}
        & \raisebox{-0.15\height}{\includegraphics[width=.138\linewidth]{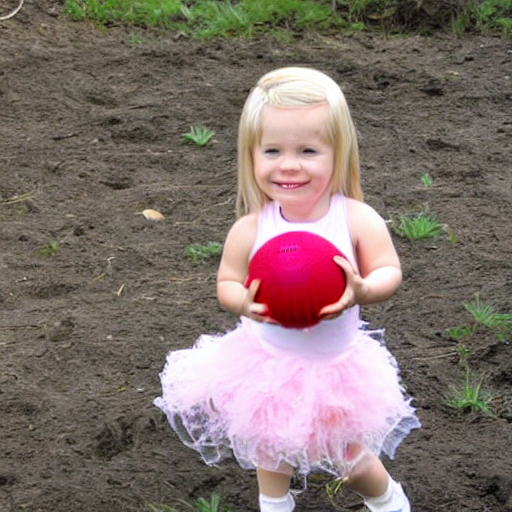}}\\
        \footnotesize a girl is holding a \textlangle*\textrangle
        & hamburger
        & book
        & pizza
        & toy car
        & mug
        & ball\\
    \end{tabular}
    \captionsetup{type=figure}
    \caption{Visualization of InteractDiffusion and others demonstrating the generation of {\it different objects} for the same action.}
    \label{fig:diff_object}
    \vspace{-10pt}
\end{table*}

\begin{table*}[ht!]
    \centering
    \setlength{\tabcolsep}{0.5pt} 
    \renewcommand{\arraystretch}{1} 
    \begin{tabular}{ccccccc}
        \raisebox{-0.35\height}{\frame{\includegraphics[width=.138\linewidth]{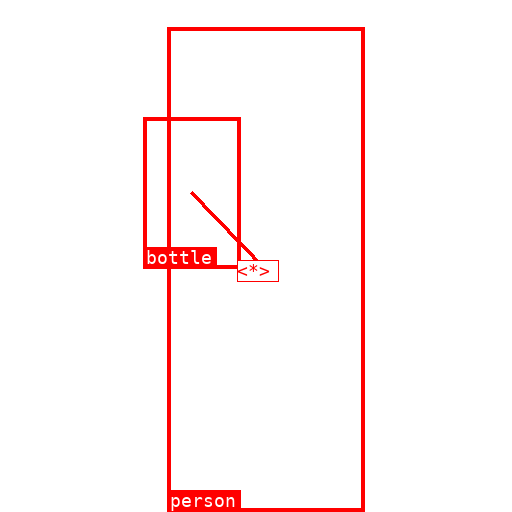}}}
        & \raisebox{-0.35\height}{\includegraphics[width=.138\linewidth]{assets/images/diff_action/bottle_drinking_with_1.png}} 
        & \raisebox{-0.35\height}{\includegraphics[width=.138\linewidth]{assets/images/diff_action/bottle_holding.png}}
        & \raisebox{-0.35\height}{\includegraphics[width=.138\linewidth]{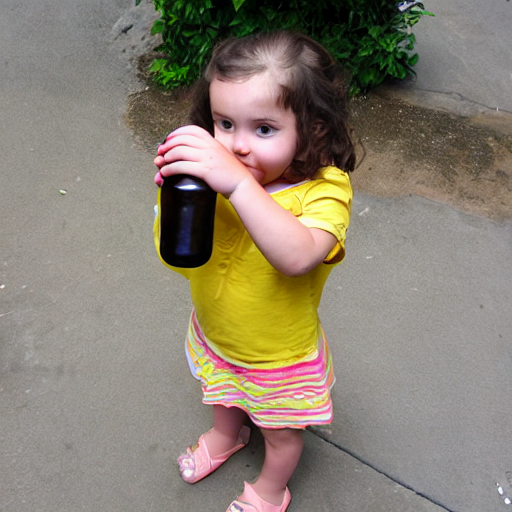}}
        & \raisebox{-0.35\height}{\includegraphics[width=.138\linewidth]{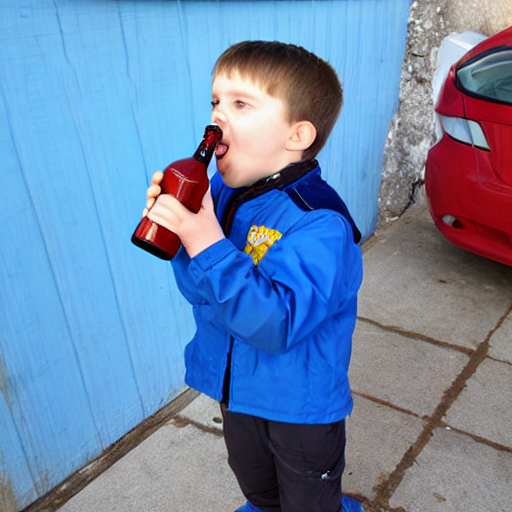}}
        & \raisebox{-0.35\height}{\includegraphics[width=.138\linewidth]{assets/images/diff_action/bottle_pouring_2.png}} 
        & \raisebox{-0.35\height}{\includegraphics[width=.138\linewidth]{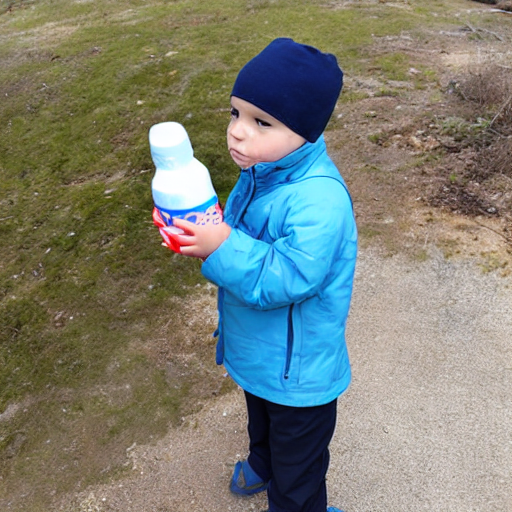}}\\
        \footnotesize a person is \textlangle*\textrangle a bottle
        & drinking
        & holding
        & opening
        & licking
        & pouring
        & inspecting\\
    \end{tabular}
    \captionsetup{type=figure}
    \caption{Visualization of InteractDiffusion demonstrating the generation of {\it different actions} for the same object. Zoom in for detail.}
    \label{fig:diff_action}
    \vspace{-10pt}
\end{table*}
In the rapidly evolving field of text-to-image synthesis, personalized Stable Diffusion models have gained popularity for their capacity to generate images with distinct styles and traits. 
The interaction module's integration allowed for fine-grained interaction control over the generative process without necessitating extensive retraining. 
In our experiments, we conducted evaluations to assess the impact of the Interaction Module on several personalized Stable Diffusion models, including CuteYukiMix\footnote{\url{https://civitai.com/models/28169/cuteyukimixadorable-style}}, RCNZCartoon3D\footnote{\url{https://civitai.com/models/66347/rcnz-cartoon-3d}}, ToonYou\footnote{\url{https://civitai.com/models/30240/toonyou}}, Lyriel\footnote{\url{https://civitai.com/models/22922/lyriel}}, DarkSushiMix\footnote{\url{https://civitai.com/models/24779/dark-sushi-mix-mix}}, RealisticVision\footnote{\url{https://civitai.com/models/4201/realistic-vision-v51}}, and ChilloutMix\footnote{\url{https://civitai.com/models/6424/chilloutmix}}. 
We observed that our transferable interaction module successfully maintains the unique stylistic attributes of personalized models while offering improved interaction controllability. 
We demonstrates visualization of InteractDiffusion on various personalized Stable Diffusion models on \cref{fig:personalized}, further affirming the module's potential to introduce interaction control without hindering the distinct qualities of these models.

\subsection{Zero-shot experiments}\label{sec:generalizability}
Following the setting in zero-shot HOI detection work \cite{thid2022}, we choose 120 HOI classes from total 600 classes in HICO-DET as unseen subset which does not involve in training, while the remaining 480 classes are in seen subset, which will be used in training. We use the same split as in \cite{thid2022}. We train the InteractDiffusion for similar number of iterations as the default setting to ensure fairness.

\cref{tab:zeroshot} shows the zero-shot performance of InteractDiffusion. In seen subset, no significant performance drop is observed, while for unseen setting, we observe mAP drop of only 3.10 and 2.30 for FGAHOI with Swin-Tiny and Swin-Large backbones, respectively. This shows that our InteractDiffusion only suffer a minor drop in its zero-shot performance, demonstrate its capability in generate unseen interaction combinations.

\section{More Qualitative Results}\label{sec:more_qualitative}
In \cref{fig:diff_object}, we visualize how our InteractDiffusion renders different objects with the same action; while \cref{fig:diff_action} shows how our InteractDiffusion renders different actions with the same object. This shows that our model can generate various combinations of interactions that maintain the coherence and naturalness of interactions between people and objects.

\section{Limitations}
Despite significant improvements in various metrics, the generated interaction still show some difference from realistic, especially in finer detail. This could be discovered on the mAP of larger detector (\ie FGAHOI(Swin-Large)), which pays attention to the finer detail in detecting HOI.
Besides, we discovered that existing large pretrained models(CLIP\cite{clip2021},StableDiffusion\cite{stablediffusion2021}) are object-focused in pre-training stage, thus lack of understanding of interaction, which hinders the performance of InteractDiffusion in controlling the interaction. We expect that a more diversely trained large model that includes the both object and interaction could boost the interaction controllability of InteractDiffusion.

\end{document}